\documentclass[10pt,twocolumn,letterpaper]{article}

\usepackage{cvpr}      

\usepackage{graphicx}
\usepackage{amsmath}
\usepackage{amssymb}
\usepackage{booktabs}

\usepackage{comment}
\usepackage{subcaption}
\usepackage{bm}
\usepackage{dsfont}
\newcommand{\rulesep}{\color{black} \unskip\ \vrule\ }
\usepackage{enumitem}
\usepackage{pythonhighlight}

\usepackage[pagebackref,breaklinks,colorlinks]{hyperref}

\usepackage[capitalize]{cleveref}
\crefname{section}{Sec.}{Secs.}
\Crefname{section}{Section}{Sections}
\Crefname{table}{Table}{Tables}
\crefname{table}{Tab.}{Tabs.}

\def\cvprPaperID{****} 
\def\confName{CVPR}
\def\confYear{2023}

\begin{document}
	
	\title{Learning 3D Scene Priors with 2D Supervision}
	
	\author{Yinyu Nie\textsuperscript{1} \ \ \ \ Angela Dai\textsuperscript{1} \ \ \ \ Xiaoguang Han\textsuperscript{2} \ \ \ \ Matthias Nießner\textsuperscript{1}\\
		\textsuperscript{1}Technical University of Munich\ \ \ \ \textsuperscript{2}The Chinese University of Hong Kong (Shenzhen)
}
\maketitle

\begin{abstract}
	Holistic 3D scene understanding entails estimation of both layout configuration and object geometry in a 3D environment.
	Recent works have shown advances in 3D scene estimation from various input modalities (e.g., images, 3D scans), by leveraging 3D supervision (e.g., 3D bounding boxes or CAD models), for which collection at scale is expensive and often intractable.
	To address this shortcoming, we propose a new method to learn 3D scene priors of layout and shape without requiring any 3D ground truth.
	Instead, we rely on 2D supervision from multi-view  RGB images.
	Our method represents a 3D scene as a latent vector, from which we can progressively decode to a sequence of objects characterized by their class categories, 3D bounding boxes, and meshes.
	With our trained autoregressive decoder representing the scene prior, our method facilitates many downstream applications, including scene synthesis, interpolation, and single-view reconstruction. Experiments on 3D-FRONT and ScanNet show that our method outperforms state of the art in single-view reconstruction, and achieves state-of-the-art results in scene synthesis against baselines which require for 3D supervision.
	Project: \url{https://yinyunie.github.io/sceneprior-page/}
\end{abstract}

\section{Introduction}
Understanding the geometric structures in real world scenes takes significant meaning in 3D computer vision.
In recent years, many prominent approaches have been proposed to understand 3D scenes for different applications, e.g., scene synthesis~\cite{wang2018deep,keshavarzi2020scenegen,Paschalidou2021NEURIPS,wang2021sceneformer,yang2021scene,leimer2022atek} and semantic reconstruction~\cite{izadinia2017im2cad,huang2018holistic,kuo2020mask2cad,dahnert2021panoptic,Gkioxari_2019_ICCV,nie2020total3dunderstanding,zhang2021holistic,liu2022towards,Nie_2021_CVPR,hou2020revealnet,gumeli2022roca,gkioxari2022learning}. These methods have shown initial promise in estimating object layouts and shapes in 3D scenes, from RGB or RGB-D input data.

\begin{figure}[!ht]
	\centering
	\includegraphics[width=0.48\textwidth]  
	{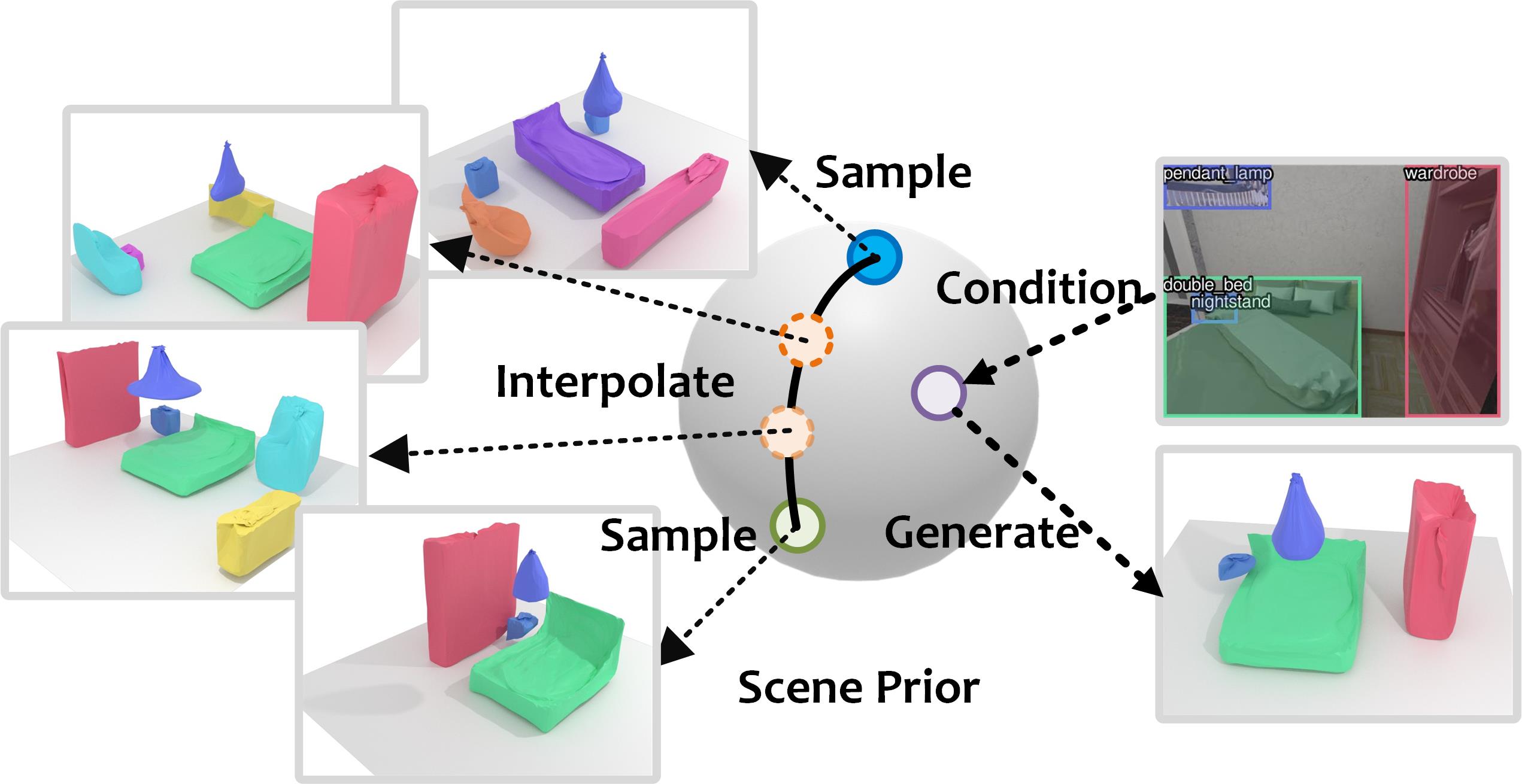}
	\caption{We learn 3D scene priors with 2D supervision. We model a latent hypersphere surface to represent a manifold of 3D scenes, characterizing the semantic and geometric distribution of objects in 3D scenes. This supports many downstream applications, including scene synthesis, interpolation and single-view reconstruction.}
	\label{fig:intro_figure}
	\vspace{-1em}
\end{figure}

However, such methods have focused on leveraging full 3D supervision for layout and shape learning, in task-specific fashion.
Such ground truth 3D scene geometry with semantic decomposition to objects and structures is difficult and expensive to acquire, requiring significant artist modeling efforts \cite{fu20213d} or expert annotations \cite{avetisyan2019scan2cad} of CAD modeling.
In contrast, 2D image and video acquisition is significantly more accessible.
We thus propose to instead learn a general 3D scene prior from only multi-view 2D information.
This approach takes a unified perspective to 3D scene understanding: our 3D scene prior is learned in a generalizable fashion from only multi-view 2D information, enabling application to multiple downstream tasks, including scene synthesis and single-view reconstruction.

In order to understand geometric 3D scene structure from only multi-view RGB images, we must bridge the domain gap between 3D geometry and RGB colors.
To this end, we propose to leverage the 2D semantic domain: we learn our 3D scene prior by ensuring that our learned 3D scene structures and objects, when differentiably rendered, match the 2D semantic distribution of estimated 2D instance segmentations from the RGB images. 
This enables supervision of 3D object and layout structures guided by 2D information.

Our 3D scene prior is learned by optimizing for a mapping from a latent space (parameterized as a hypersphere surface) to a manifold of 3D scenes, as shown in  Fig.~\ref{fig:intro_figure}. 
Each random vector sampled from the latent space is decoded into a sequence of objects autoregressively using our permutation-invariant transformer. Each output object is characterized by their class category, 3D bounding box and mesh. By training this decoder, we encode the scene prior into a latent space, sampling on which enables us to synthesize plausible 3D scenes. With such formulation, our method supports many other downstream tasks on different conditions, e.g., scene synthesis, interpolation and single-view reconstruction.

In summary, we present our contributions as follows:

\begin{itemize}[noitemsep,topsep=0.5em,parsep=0pt,partopsep=0pt]
\item We present a new perspective on general 3D scene prior learning with only 2D supervision. We learn the prior distribution of both semantic object instance layouts and shapes via mapping from a latent space to the 3D scene space, by training from 2D multi-view data.

\item We formulate scene prior learning as an autoregressive sequence decoding problem, and propose a novel permutation-invariant transformer corresponding with a 2D view loss to map a latent vector from a hypersphere surface to a sequence of semantic instances characterized by their class categories, 3D bounding boxes and meshes.

\item Our learned 3D scene prior supports many downstream tasks: single-view scene reconstruction, scene synthesis and interpolation. Experiments on 3D-FRONT and ScanNet show that our method matches state-of-the-art performance in scene synthesis against methods requiring 3D supervision, and outperforms state of the art in single-view reconstruction.
\end{itemize}

\begin{figure*}
	\centering
	\includegraphics[width=\textwidth]  
	{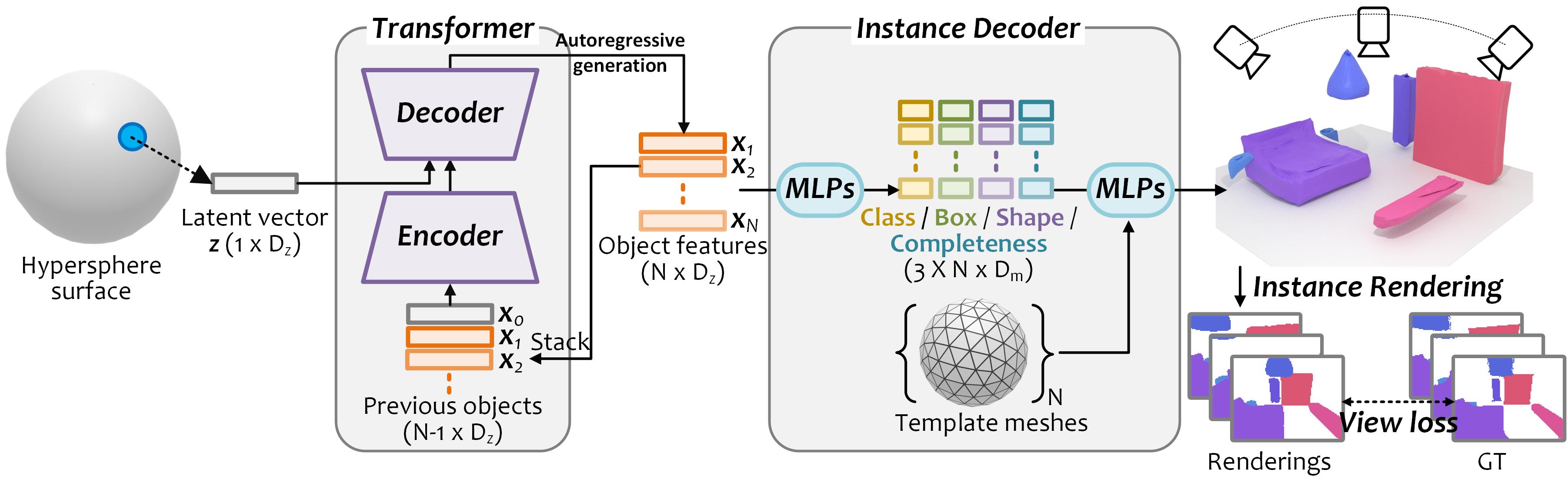}
	\caption{Overview of our approach. Our method represents each scene as a learnable latent vector $\bm{z}$ on a hypersphere. To build mapping from the latent surface to the semantic scene space, we propose a permutation-invariant transformer to generate objects $\{\bm{x}_{k}\}$ autoregressively conditioned on previous objects and $\bm{z}$. For each object feature, we regress its object class label, 3D bounding box, shape feature and completeness score, which respectively represent the object layout, geometry, and the completeness of the output scene. We do not require 3D supervision, and instead leverage differentiable rendering with multi-view 2D instance masks supervision.}
	\label{fig:overview}
	\vspace{-1em}
\end{figure*}

\section{Related Work}
\paragraph{3D Semantic Scene Synthesis} 3D semantic scene synthesis has received extensive exploration in recent decades, particularly following the popularity of 3D indoor scene datasets and 3D deep learning. Given a collection of synthesized or real 3D scenes, many methods have been proposed to learn the distribution of object layouts to synthesize unseen but plausible 3D scene arrangements.

Traditional methods for scene synthesis usually formulate the task as data-driven object placement~\cite{zhang2019survey}, where a scene with object instances is represented as a graph~\cite{zhu2018modeling,chang2014learning,chang2017sceneseer,qi2018human}, or is connected to human activities to build a human-centric representation~\cite{fisher2015activity,fu2017adaptive,ma2016action,qi2018human}. Scene priors are explicitly hard-coded based on design guidelines~\cite{merrell2011interactive,yeh2012synthesizing}, represented by frequency statistics (e.g., objects co-occurrence map)~\cite{fisher2010context,chang2014learning,chang2017sceneseer}, inferred by affordance map from human activities~\cite{jiang2012learning,fu2017adaptive,fisher2015activity}, or learned from arrangement examples in 3D scenes~\cite{fu2017adaptive,fisher2012example}. Given an initial scene~\cite{fu2017adaptive,yeh2012synthesizing,merrell2011interactive}, a 2D sketch~\cite{xu2013sketch2scene} or a text~\cite{chang2014learning,chang2017sceneseer}, several methods synthesize object placements under the guidance of scene priors with different optimization strategies, e.g., manual interactions~\cite{merrell2011interactive,chang2017sceneseer,savva2017scenesuggest}, iterative methods~\cite{fu2017adaptive,li2019grains,xie2013reshuffle,fisher2015activity} or non-linear optimization~\cite{chang2014learning,chang2017sceneseer,fisher2012example,xu2013sketch2scene,yu2011make,yeh2012synthesizing,qi2018human}, to maximize a plausibility score or a joint probability.

With deep learning techniques, many methods instead learn scene priors with neural networks. They keep the problem formulation and use recurrent networks~\cite{wang2018deep,Paschalidou2021NEURIPS,ritchie2019fast,li2019grains,wang2019planit,wang2021sceneformer}, feed-forward models~\cite{zhang2020deep,nie2022pose2room}, GANs~\cite{yang2021indoor}, or VAEs~\cite{purkait2020sg,yang2021scene} to synthesize object poses recursively or in one stage from an initial scene or random noise, and usually require 3D information for training. In contrast, our method directly learns the joint semantic and geometric distribution of instances in scenes with 2D supervision, which enables a general scene priors that supports semantic scene synthesis from a latent random vector.
\vspace{-1em}

\paragraph{Single-view Scene Reconstruction} Single image reconstruction methods have also been designed around various task-specific, data-driven scene priors~\cite{Gkioxari_2019_ICCV,popov2020corenet,dahnert2021panoptic,gkioxari2022learning,gumeli2022roca,kuo2020mask2cad,kuo2021patch2cad,huang2018holistic,izadinia2017im2cad,liu2022towards,nie2020total3dunderstanding,zhang2021holistic,kundu20183d,tulsiani2018factoring}. Izadinia et al.~\cite{izadinia2017im2cad} propose a deep learning-based method to retrieve and align CAD models to an input image. Further methods were developed with a similar top-down approach to reconstruct a scene by detection and retrieval, in which \cite{izadinia2017im2cad,kundu20183d,huang2018holistic} use a render-and-compare strategy. \cite{kuo2020mask2cad,kuo2021patch2cad,gumeli2022roca} improve the offline shape retrieval with an end-to-end 2D-3D shape matching with pose alignment. Other works focus on shape reconstruction instead of retrieval, where object shapes are learned as meshes~\cite{Gkioxari_2019_ICCV,nie2020total3dunderstanding,gkioxari2022learning}, volume grids~\cite{tulsiani2018factoring,popov2020corenet}, or implicit functions~\cite{zhang2021holistic,liu2022towards,liu2022towards} with the supervision of paired 3D object shapes or multi-view images. However, modeling scenes with a top-down manner usually focuses on per-object perception without any global scene reasoning. In contrast, we focus on learning prior distribution of 3D scenes, to generate plausible object layouts conditioned on an input image.

\section{Methodology}
Our work aims to learn the prior distribution of 3D semantic scenes. We formulate this problem by learning a mapping $f$ from a parameterized latent space $\mathcal{Z}$ to the scene space $\mathcal{S}$, and hope each sample $\bm{z}\in\mathcal{Z}$ corresponds to a reasonable 3D scene $ \bm{s}\in\mathcal{S}$ as in
\begin{equation}
\bm{s} = f(\bm{z});\ \bm{z}\in\mathcal{Z},\ \bm{s}\in\mathcal{S}.
\label{eq:formulation}
\end{equation}
In Eq.~\ref{eq:formulation}, there are four key components: the parameterization of latent space $\mathcal{Z}$ and scene space $\mathcal{S}$, the architecture of $f$, and the losses to train $f$. We illustrate our approach in Fig.~\ref{fig:overview}, the latent space $\mathcal{Z}$ is represented as a unit hypersphere surface, each latent vector $\bm{z}$ sampled from which conditions the generation of an object sequence with our permutation-invariant transformer. We formulate a 3D scene $\bm{s}\in\mathcal{S}$ as a set of objects with their class categories, positions and shapes, thus each object from the transformer is decoded into a category label, a 3D bounding box and a mesh with a completeness score indicating whether the generated scene is complete or not. Training our network only needs 2D supervision. We design a 2D view loss between the renderings of generated scenes and the ground-truth instance masks under respective camera views.

\subsection{Latent Space Parameterization}
We parameterize the latent space $\mathcal{Z}$ as a unit hypersphere surface. We choose a spherical surface as the latent representation because it does not have strict boundaries (e.g, uniform distribution), and sampling on it will not produce outliers (e.g., gaussian distribution)~\cite{davydov2022adversarial}, which guarantees the continuity when moving from one latent vector to another.
We formulate the sampling process from $\mathcal{Z}$ as
\begin{equation}
	\bm{z}=\frac{\textstyle\sum_{i=1}^{M} w_{i}\cdot\phi_{i}}{||\textstyle\sum_{i=1}^{M} w_{i}\cdot\phi_{i}||_{2}}; \ \textstyle\sum_{i=1}^{M}w_{i}=1, w_{i}>0
	\label{eq:latent_vec}
\end{equation}
where $\{\phi_{i}|\phi_{i}\in\mathbb{R}^{D_{z}}, ||\phi_{i}||_{2}=1\}$ is a set of $M$ fixed vectors with unit length, which uniformly spreads on the sphere surface. $D_{z}$ is the dimension of the hypersphere. Therefore, by sampling a set of weights $\{w_{i}\}$, we can obtain a latent vector $\bm{z}\in\mathbb{R}^{D_{z}}$ for scene generation. Note that both $\{w_{i}\}$ and our network $f$ are learnable during training.

\subsection{Permutation-invariant Transformer}
We generate a sequence of objects in a scene, conditioned on the latent vector $\bm{z}$, in an autoregressive fashion. New objects are generated conditioned on the previous ones, which can be formulated as to estimate the distribution of
\begin{equation}
	P_{\theta}\left(\bm{x}_{1},...,\bm{x}_{N}|\bm{z}\right)=P_{\theta}\left(\bm{x}_{1}|\bm{z}\right)\textstyle\prod_{k=2}^{N} P_{\theta}\left(\bm{x}_{k}|\bm{x}_{1},..., \bm{x}_{k-1};\bm{z}\right),
\end{equation}
where $\bm{x}_{k}$ denotes the $k$-th object feature. $N$ is the maximal number of objects to be generated. $\theta$ is the trainable network. Every new object $\bm{x}_{k}$ is generated conditioning on $\bm{x}_{1},...,\bm{x}_{k-1}$ and $\bm{z}$. 

We observe that objects in a scene are an unordered set, where each permutation of an object sequence should be equivalent to each other, e.g., \textit{[chair, table, lamp]} should be equivalent to \textit{[lamp, chair, table]} given the three objects. Thus, the generation of the next object $\bm{x}_{k}$ should be independent to the order of previous objects $\{\bm{x}_{1},...,\bm{x_{k-1}}\}$, which can be formulated as

\begin{equation}
\begin{aligned}
	&P_{\theta}\left(\bm{x}_{k}|\bm{x}_{p_{1}},..., \bm{x}_{p_{k-1}};\bm{z}\right)=P_{\theta}\left(\bm{x}_{k}|\bm{x}_{q_{1}},..., \bm{x}_{q_{k-1}};\bm{z}\right) \\
	&1\leq p_{i}, q_{i}\leq k-1\ \ \ \text{and}\ \ \ p_{i}\neq p_{j}, q_{i}\neq q_{j},\ \forall\ i\neq j.
\end{aligned}
\end{equation}

\begin{figure}[!t]
	\centering
	\includegraphics[width=0.5\textwidth]  
	{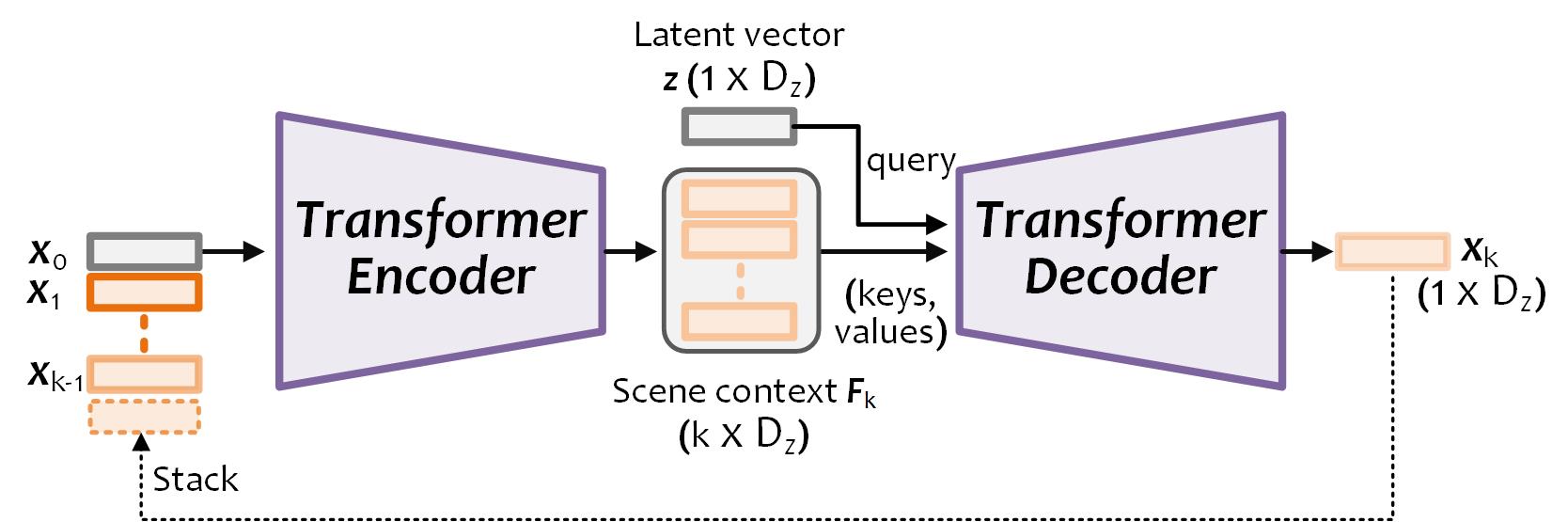}
	\caption{Transformer for autoregressive object generation. From the start token $\bm{x}_{0}$, we design a permutation-invariant transformer to autoregressively generate a new object feature $\bm{x}_{k}$ by querying the encoded scene context $\bm{F}_{k}$ with the latent vector $\bm{z}$.}
	\label{fig:transformer}
	\vspace{-1em}
\end{figure}

This requires our generation network to be permutation-invariant to the input objects. We model this step using a transformer backbone~\cite{katharopoulos_et_al_2020}, which is illustrated in Fig.~\ref{fig:transformer}. Inspired by \cite{paschalidou2021atiss,vaswani2017attention} where removing the positional encoding in a transformer encoder makes the network agnostic to the object order, we consider this design and encode permutation-invariant scene context $\bm{F}_{k}\in\mathbb{R}^{k\times D_{z}}$ from previous objects $\{\bm{x}_{1},...,\bm{x}_{k-1}\}$. Note that $\bm{x}_{0}$ in Fig.~\ref{fig:overview} and~\ref{fig:transformer} is a learnable start token which indicates an empty room.

$\bm{F}_{k}$ contains attention signals between object pairs and the scene context feature. We use $\bm{F}_{k}$ as the keys and values, and the latent vector $\bm{z}$ as the query to infer the next object based on the scene context. We adopt a transformer decoder for object inference, where we also do not use positional encoding. Afterwards, it outputs the next possible object $\bm{x}_{k}\in\mathbb{R}^{D_{z}}$. We forward this step autoregressively to obtain an object feature sequence $\{\bm{x}_{1},...,\bm{x}_{N}\}$.

\subsection{Layout and Shape Decoder}
In this section, we decode object features $\{\bm{x}_{k}\}$ into a set of object categories, 3D bounding boxes and shapes.

\vspace{-1em}
\paragraph{Scene Parameterization}
We focus on indoor scenarios, and represent a 3D scene in a world coordinate system with the floor center located at the origin. Each object inside is characterized by its class label $\bm{l}\in\mathbb{L}$, axis-aligned 3D bounding box with size $\bm{s}\in\mathbb{R}^{3}$ and center $\bm{c}\in\mathbb{R}^{3}$, and 3D mesh $\bm{M}\in\mathbb{M}$ in the canonical system, where $\mathbb{L}$ denotes all class categories and $\mathbb{M}$ is the shape space.

\vspace{-1em}
\paragraph{Layout Decoder}
We adopt a layout decoder to estimate the class category $\bm{l}_{k}$, box center $\bm{c}_{k}$, size $\bm{s}_{k}$ and completeness score $\bm{p}_{k}$ from each object feature $\bm{x}_{k}$, which is implemented by two MLPs respectively:
\begin{equation}
	\{\bm{l}_{k},\bm{c}_{k},\bm{s}_{k},\bm{p}_{k}\}=\text{MLP}_{2}(\text{MLP}_{1}(\bm{x}_{k})),\ k=1,...,N.
	\label{eq:layout_decoder}
\end{equation}
The completeness $\bm{p}_{k}$ is a binary score  of whether $\bm{x}_{k}$ is the last object or not, indicating whether we should continue object generation.

\vspace{-1em}
\paragraph{Shape Decoder}
We estimate object shapes by deforming a template sphere from $\bm{x}_{k}$. Inspired by \cite{groueix2018papier}, we use MLPs to learn the offsets from vertices $\bm{V}\in\mathbb{R}^{|\bm{V}|\times3}$ on the template sphere to the target surface by
\begin{equation}
	\Delta \bm{v}_{k} = \text{MLP}_{3}(\text{Concat}[\text{MLP}_{1}(\bm{x}_{k}), \bm{v}]), \bm{v}\in\bm{V},
	\label{eq:shape_decoder}
\end{equation}
where $\bm{V}$ denotes the set of vertices on the template; MLP\textsubscript{3} is an point-wise MLP, and we share MLP\textsubscript{1} with Eq.~\ref{eq:layout_decoder}. Afterwards, the vertices of the $k$-th object mesh $\bm{M}_{k}$ is predicted as $\bm{S}_{k}=\{\bm{v}+\Delta\bm{v}_{k}\}$. Note that $\bm{M}_{k}$ is in a canonical system, which is transformed to the world system with the bounding boxes in Eq.~\ref{eq:layout_decoder} by $\bm{o}_{k}=\bm{s}_{k}\cdot\bm{M}_{i}+\bm{c}_{k}$. Finally, we obtain $\{\bm{o}_{k}\}$ as the set of output objects in a scene.

\begin{figure}[!t]
	\centering
	\includegraphics[width=0.5\textwidth]  
	{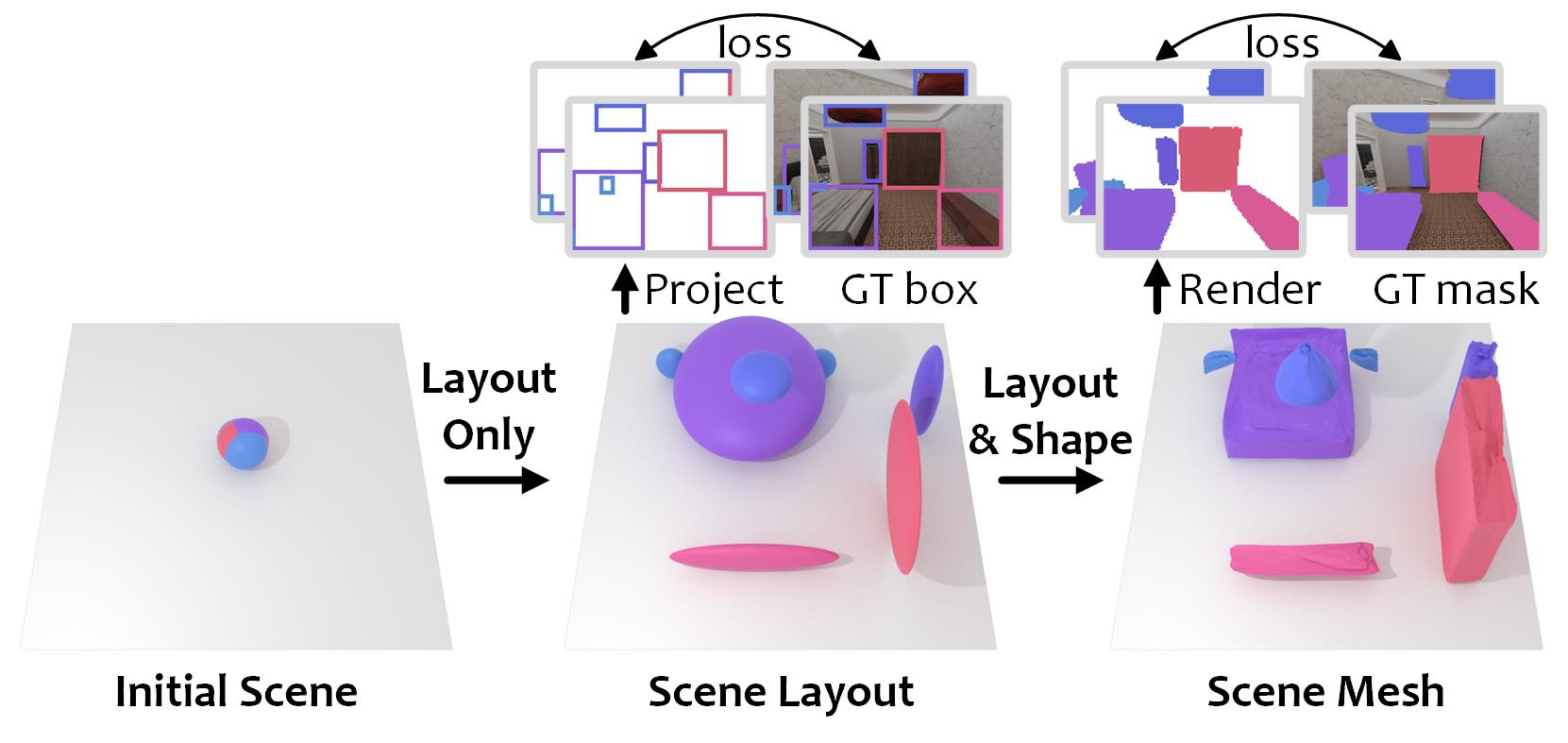}
	\caption{Optimization strategy for layout and shape training. We train layout learning and mesh deformation end-to-end with two stages to guarantee that objects are deformed after alignment.}
	\label{fig:optimization}
	\vspace{-1em}
\end{figure}

\subsection{View Loss}
\label{sec:view_loss}
Instead of using 3D data, our method leverages multi-view posed instance masks with tracks to supervise layout and shape generation. We consider $T$ views in a 3D scene with $n$ tracked objects $\{\bm{o}^{gt}_{j}\}$, $j$=1,...,$n$. Each object $\bm{o}^{gt}_{j}$ is observable in $T_{j}$ views, where $T_{j}\subseteq T$. We denote $\bm{m}^{p}_{j}$ as the mask of object $\bm{o}^{gt}_{j}$ observed in the view $p\in T_{j}$. For each mask $\bm{m}^{p}_{j}$, there are a class label $\bm{l}_{j}^{gt}$ and camera parameters $\text{cam}^{p}$.

We train our model with a set of view losses between the ground-truth masks and the renderings of generated objects. We illustrate our training strategy in Fig.~\ref{fig:optimization}, which is trained end-to-end with two stages: \textbf{1) layout pretraining} and \textbf{2) layout+shape joint learning}. We pretrain the layout first because 1) shape deformation from 2D requires an ideal alignment between its renderings and ground-truth masks; 2) differentiable rendering is not required in the first stage thus the efficiency is largely improved.

\vspace{-1em}
\paragraph{Layout Loss}
Our network is initialized with all objects $\{\bm{o}_{k}\}$ at the floor center. In the first stage, we learn the object layout without shape decoding (i.e., $\Delta\bm{v}_{k}=0$ in Eq.~\ref{eq:shape_decoder}). Only object categories, centers, sizes and completeness scores are optimized ($\{\bm{l}_{k},\bm{c}_{k},\bm{s}_{k},\bm{p}_{k}\}$). Therefore, we define the layout loss $\mathcal{L}_{layout}$ as the combination of 1) classification loss $\mathcal{L}_{l}$; 2) 2D bounding box loss $\mathcal{L}_{box}$; 3) completeness loss $\mathcal{L}_{p}$, and 4) frustum loss $\mathcal{L}_{f}$.

For an object prediction $\bm{o}_{k}$, we project the vertices on its mesh to all views in $T$ and calculate the 2D bounding boxes $\bm{B}_{k}=\{\bm{b}^{1},..., \bm{b}^{T}\}_{k}$. For each ground-truth object $\bm{o}^{gt}_{j}$, we also have its 2D bounding boxes $\bm{B}^{gt}_{j}=\{\bm{b}^{1},...,\bm{b}^{T_{j}}\}^{gt}_{j}$ in $T_{j}$ views. We let $\bm{l}_{k}$ as the predicted class label of $\bm{o}_{k}$, and $\bm{l}^{gt}_{j}$ as the ground-truth label of $\bm{o}^{gt}_{j}$ respectively. For each prediction pair $(\bm{B}_{k},\bm{l}_{k})$, we adopt the Hungarian algorithm~\cite{stewart2016end,carion2020end} to find its optimal bipartite matching $(\bm{B}^{gt}_{\sigma(k)}, \bm{l}^{gt}_{\sigma(k)})$ from the ground truth. Therefore we can obtain the matches from each prediction $\bm{o}_{k}$ to a ground-truth $\bm{o}^{gt}_{\sigma(k)}$, which is the basis of our loss functions. For details of Hungarian matching, we refer to the supplemental.

Based on Hungarian matching, our layout loss can be formulated as
\begin{equation}
\mathcal{L}_{L}=\lambda_{l}\mathcal{L}_{l}+\lambda_{box}\mathcal{L}_{box}+\lambda_{p}\mathcal{L}_{p}+\lambda_{f}\mathcal{L}_{f},
\end{equation}
where $\mathcal{L}_{l}$ is the cross-entropy loss between $\bm{l}_{k}$ and $\bm{l}^{gt}_{\sigma(k)}$ for object classification; $\mathcal{L}_{box}$ is the average $L_{1}$ distance between 2D bounding boxes in $\bm{B}_{k}$ and $\bm{B}^{gt}_{\sigma(k)}$ on each view of $T_{\sigma(k)}$; $\mathcal{L}_{p}$ is the binary cross entropy loss to judge whether $\bm{o}_{k}$ is the last object or not. For objects whose initial position is not located in a view frustum of their matched ground-truth, we design a frustum loss $\mathcal{L}_{f}$ to enforce their 3D center $\bm{c}$ moving into the target frustum. Further details can be found in the supplemental.

\vspace{-1em}
\paragraph{Shape Loss}
After the layout loss converges, we switch on the shape decoder in Eq.~\ref{eq:shape_decoder} and use 2D instance masks $\{\bm{m}^{p}_{j}\}$ for mesh deformation. In addition to 2D bounding boxes, we also output the silhouette renderings of each $\bm{o}_{k}$ on all views $T$, with a differentiable renderer from ~\cite{ravi2020accelerating}. The shape loss of $\bm{o}_{k}$ is defined as a multi-view mask loss as
\begin{equation}
	\mathcal{L}_{S}=\textstyle\sum_{p\in T_{\sigma(k)}}\text{BCE}(\bm{r}^{p}_{k},\bm{m}^{p}_{\sigma(k)})\mathds{1}[\text{IoU}(\bm{r}^{p}_{k},\bm{m}^{p}_{\sigma(k)})>0.5],
\end{equation}
where $\bm{r}^{p}_{k}$ is the rendering of $\bm{o}_{k}$ in view $p\in T_{\sigma(k)}$; $\bm{m}^{p}_{\sigma(k)}$ is the corresponding mask of $\bm{o}^{gt}_{\sigma(k)}$. For pairs whose IoU larger than 0.5, we use binary cross entropy to calculate the mask loss, which is inspired by \cite{gkioxari2022learning} that we only use those well-aligned masks for supervision.

Finally, we train the our network end-to-end with $\mathcal{L}=\mathcal{L}_{L} + \lambda_{S}\mathcal{L}_{S}$.
All losses above are averaged across all objects. $\{\lambda\}$ are weights that balance the losses.

\begin{figure*}[!ht]
	\centering
	\begin{subfigure}[t]{0.23\textwidth}
		\includegraphics[width=0.49\textwidth]
		{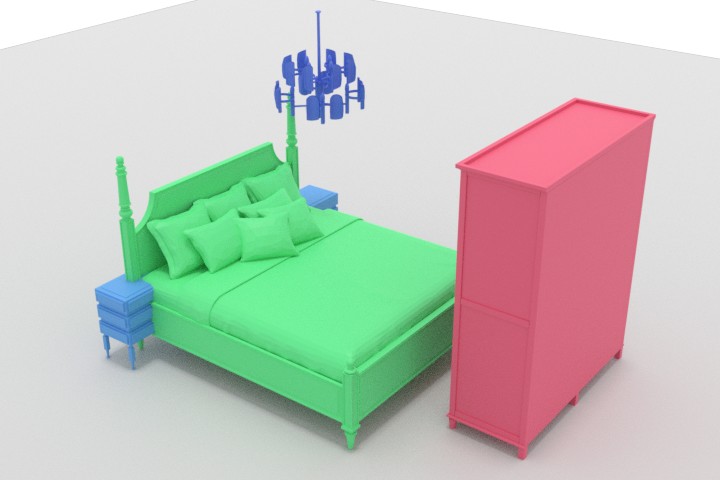}%
		\hfill
		\includegraphics[width=0.49\textwidth]
		{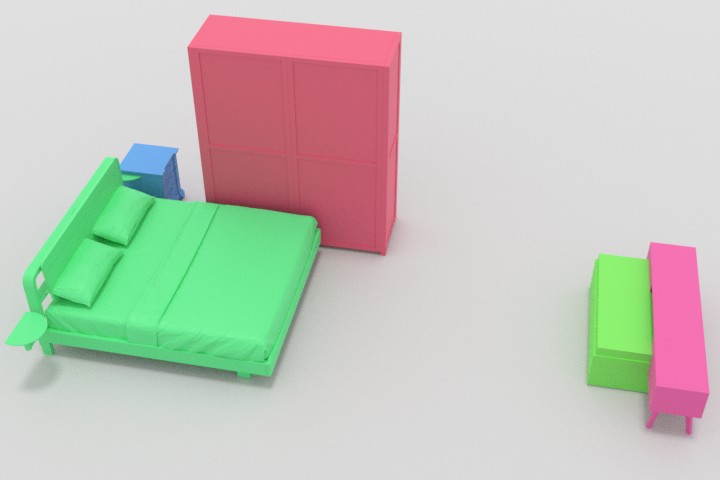}
		\includegraphics[width=0.49\textwidth]
		{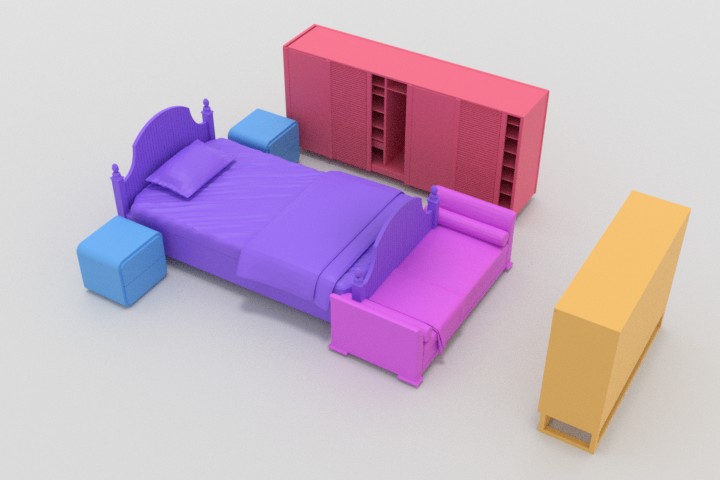}%
		\hfill
		\includegraphics[width=0.49\textwidth]
		{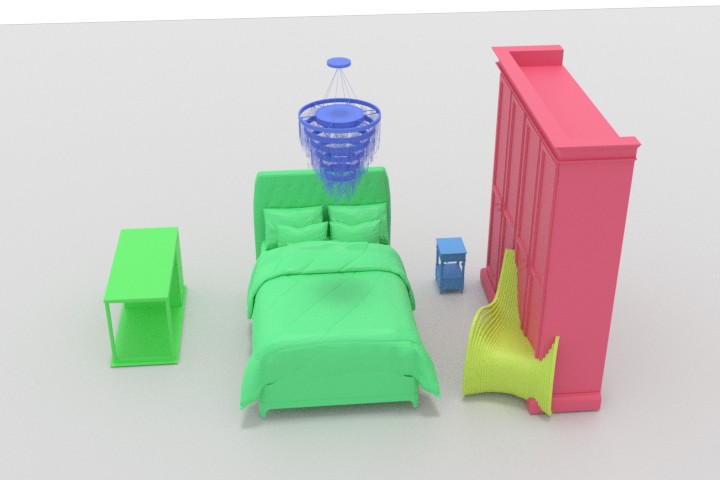}
		\includegraphics[width=0.49\textwidth]
		{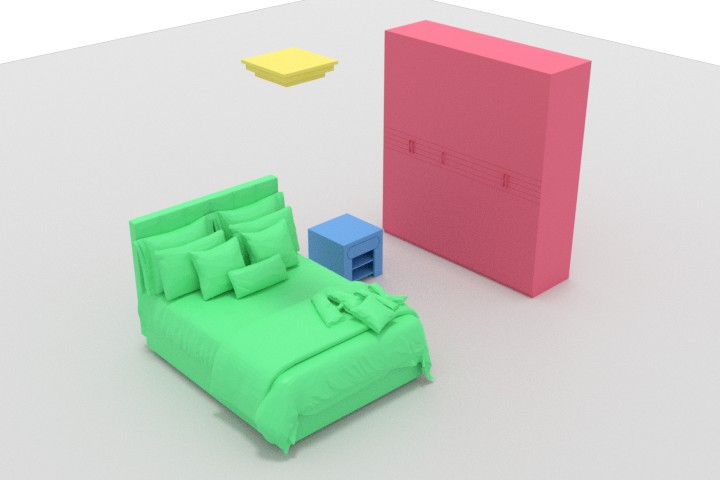}%
		\hfill
		\includegraphics[width=0.49\textwidth]
		{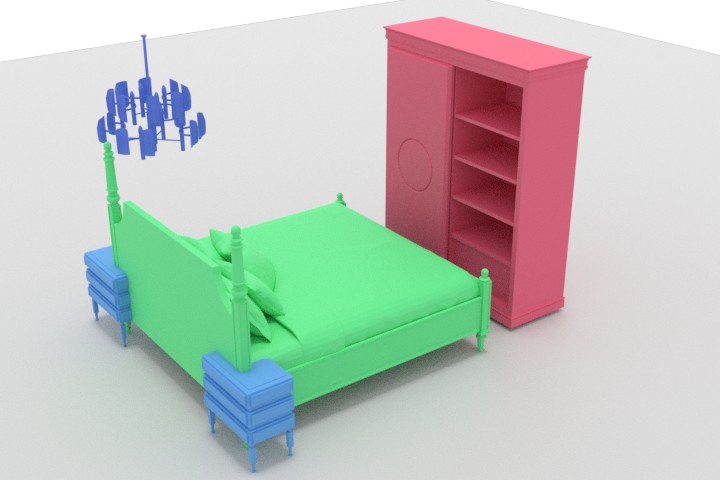}
		\includegraphics[width=0.49\textwidth]
		{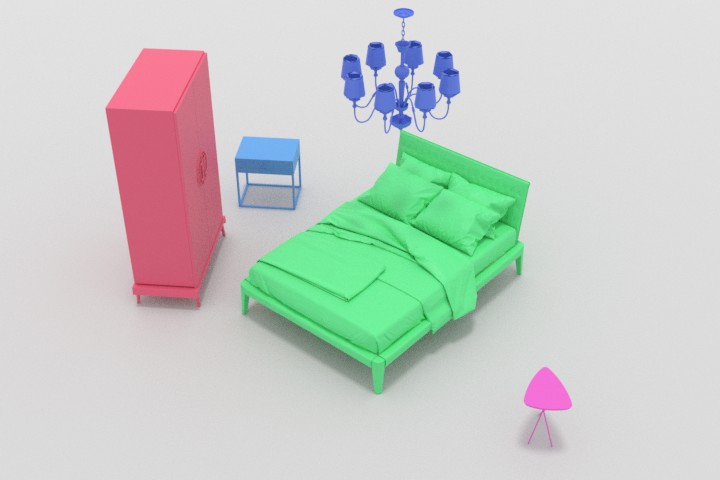}%
		\hfill
		\includegraphics[width=0.49\textwidth]
		{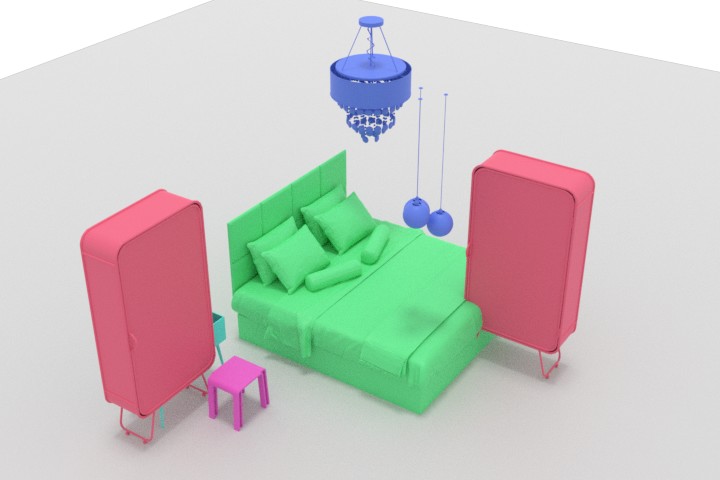}
		\includegraphics[width=0.49\textwidth]
		{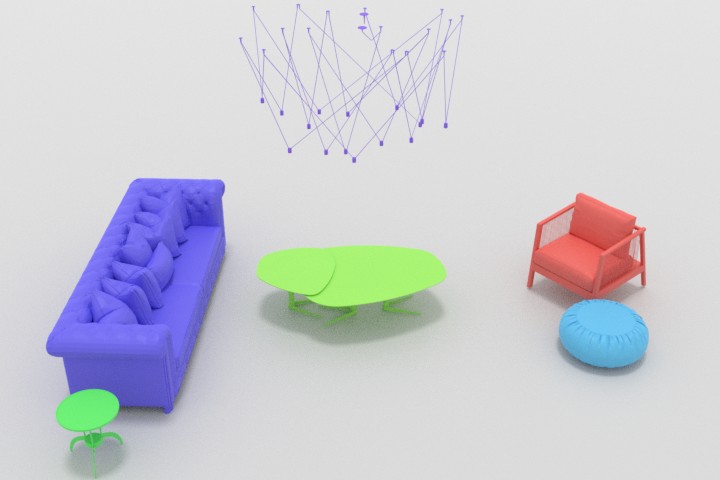}%
		\hfill
		\includegraphics[width=0.49\textwidth]
		{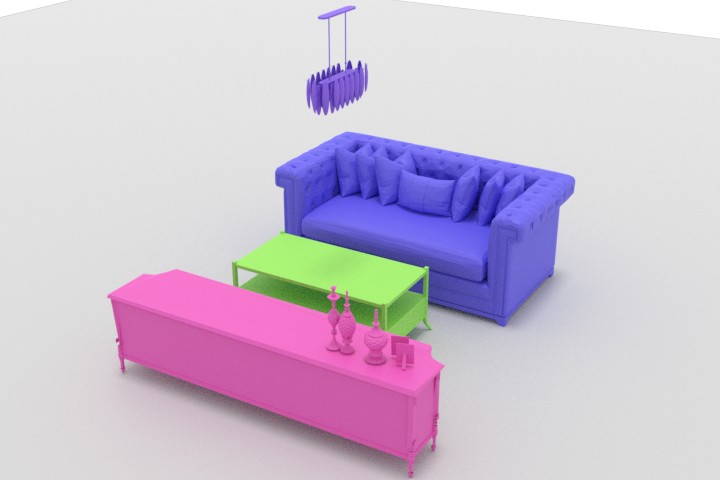}
		\includegraphics[width=0.49\textwidth]
		{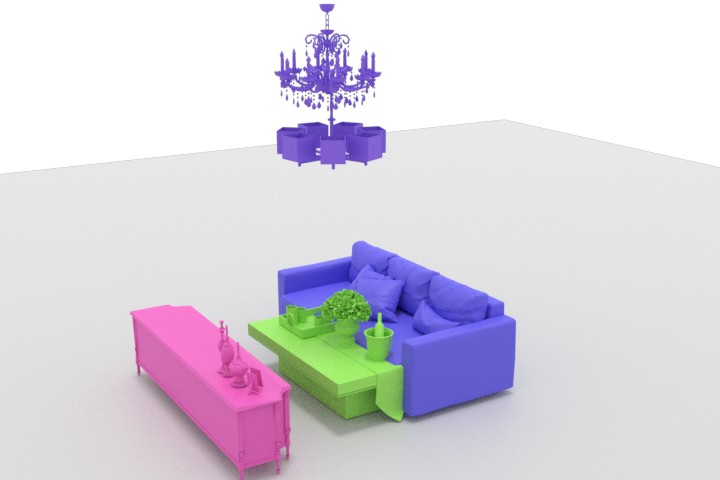}%
		\hfill
		\includegraphics[width=0.49\textwidth]
		{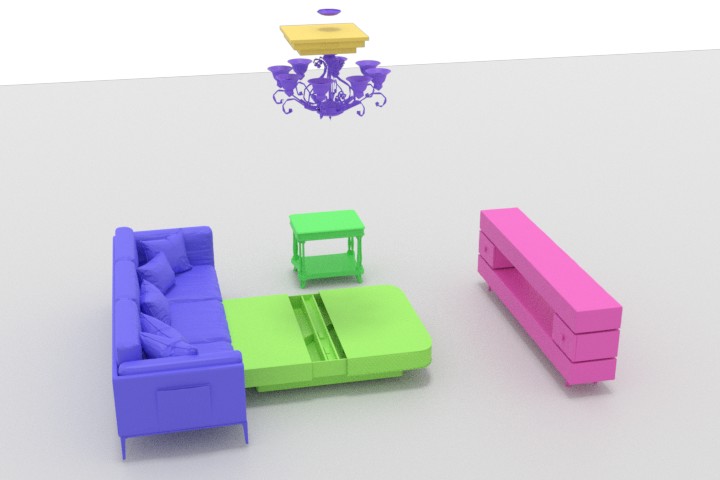}
		\includegraphics[width=0.49\textwidth]
		{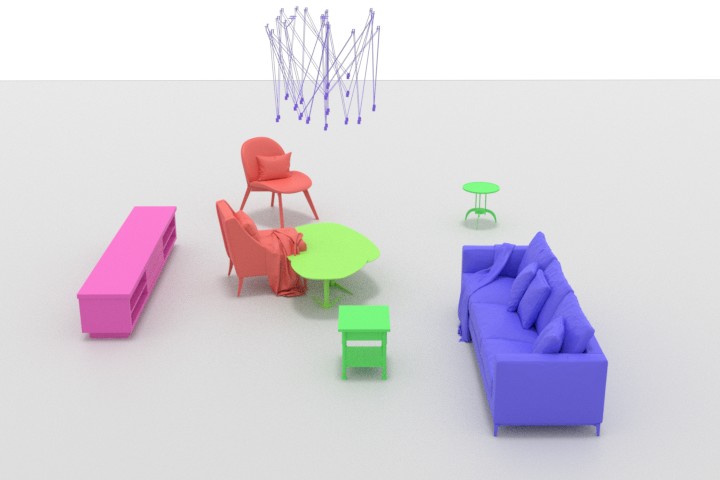}%
		\hfill
		\includegraphics[width=0.49\textwidth]
		{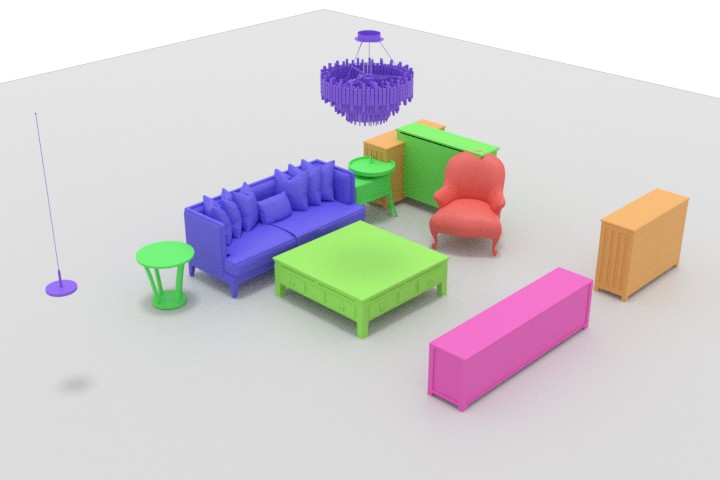}
		\caption{ATISS~\cite{paschalidou2021atiss}}
	\end{subfigure}
	\rulesep
	\begin{subfigure}[t]{0.23\textwidth}
		\includegraphics[width=0.49\textwidth]
		{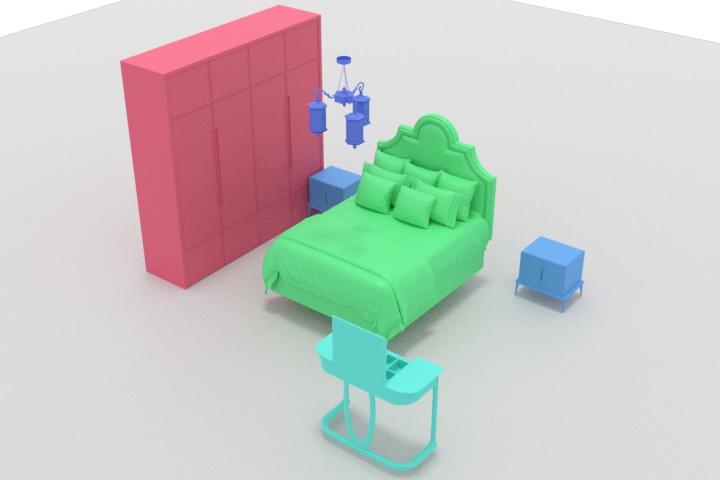}%
		\hfill
		\includegraphics[width=0.49\textwidth]
		{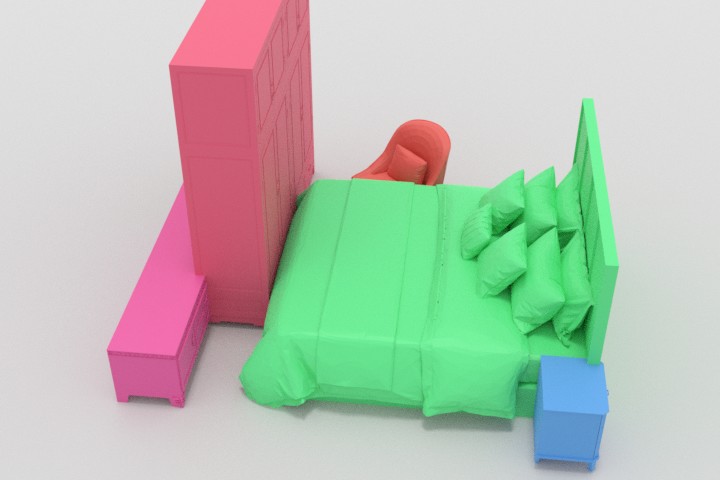}
		\includegraphics[width=0.49\textwidth]
		{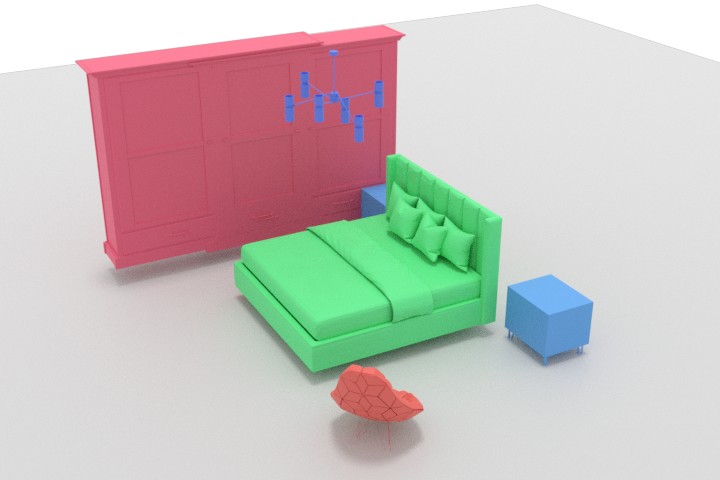}%
		\hfill
		\includegraphics[width=0.49\textwidth]
		{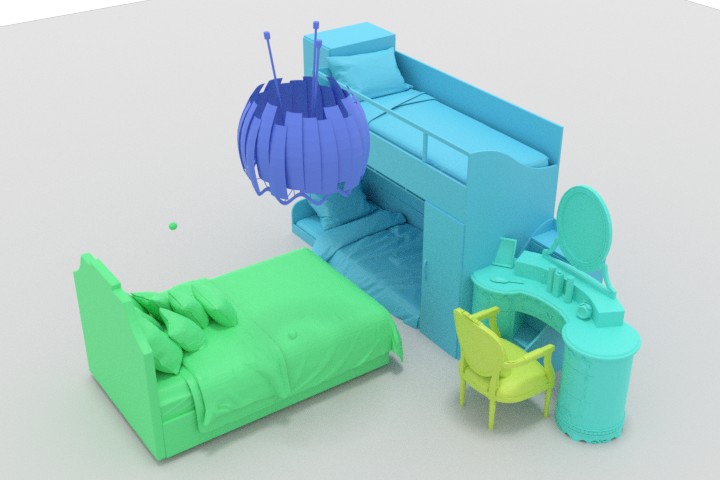}
		\includegraphics[width=0.49\textwidth]
		{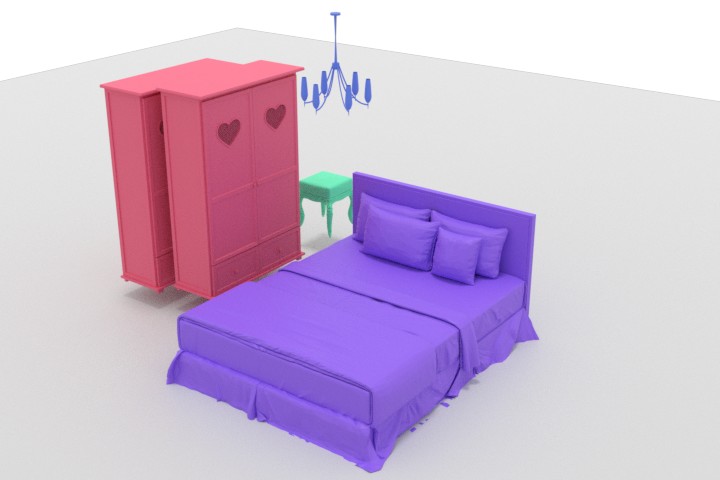}%
		\hfill
		\includegraphics[width=0.49\textwidth]
		{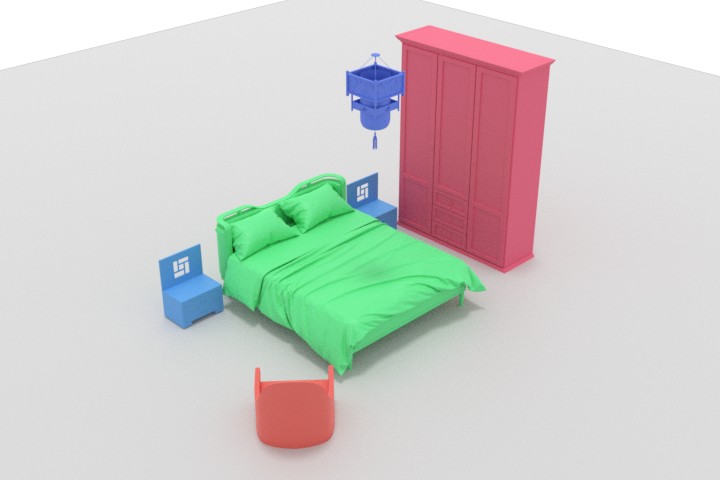}
		\includegraphics[width=0.49\textwidth]
		{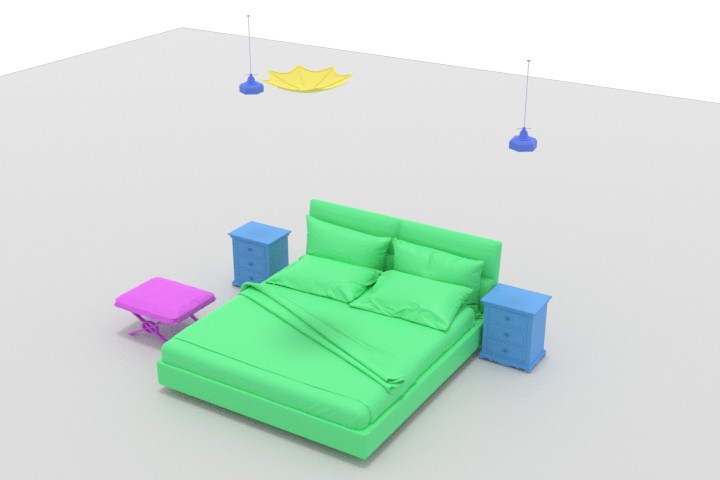}%
		\hfill
		\includegraphics[width=0.49\textwidth]
		{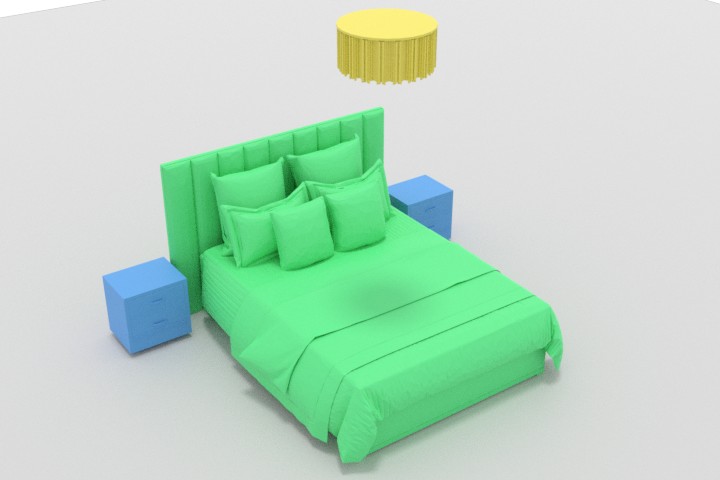}
		\includegraphics[width=0.49\textwidth]
		{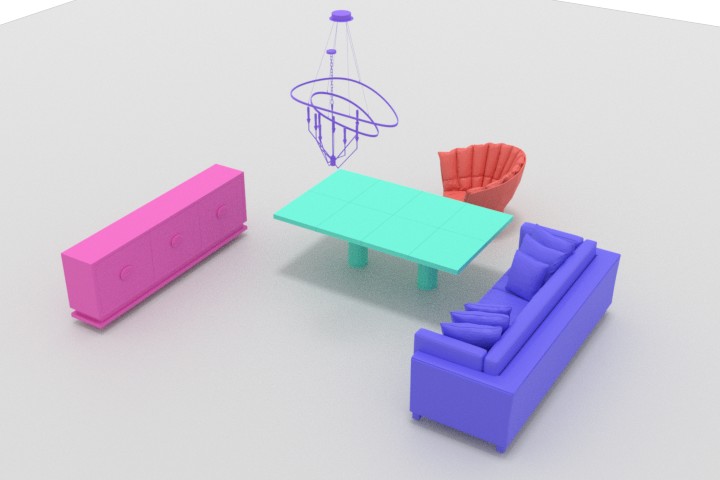}%
		\hfill
		\includegraphics[width=0.49\textwidth]
		{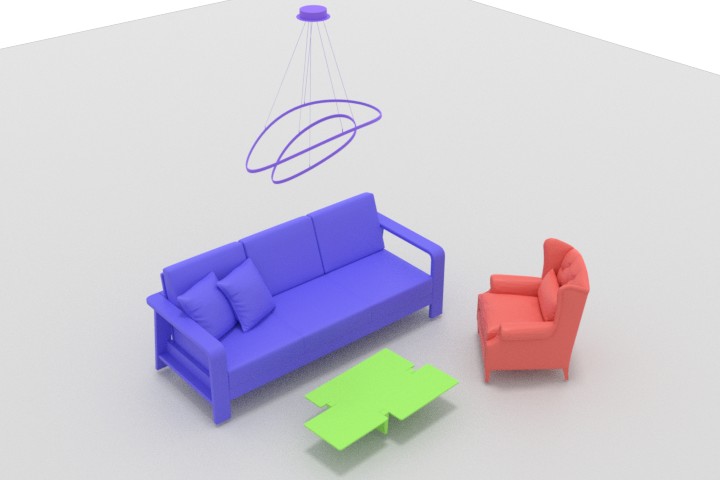}
		\includegraphics[width=0.49\textwidth]
		{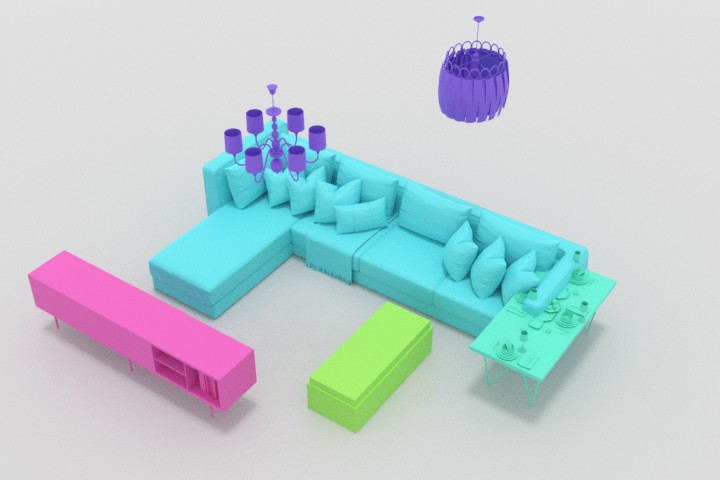}%
		\hfill
		\includegraphics[width=0.49\textwidth]
		{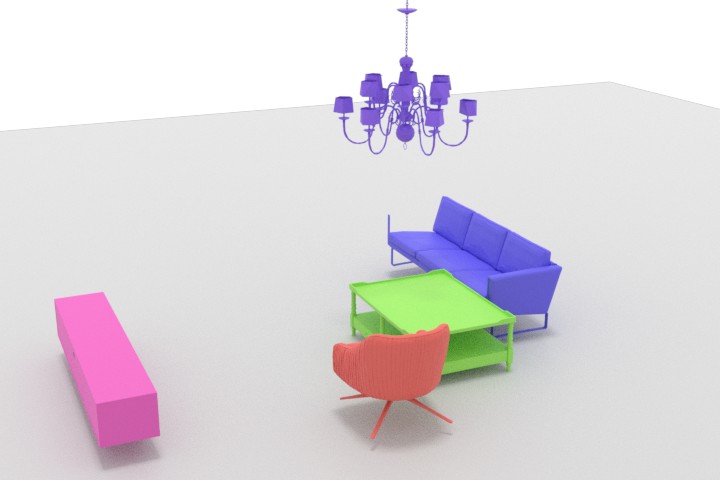}
		\includegraphics[width=0.49\textwidth]
		{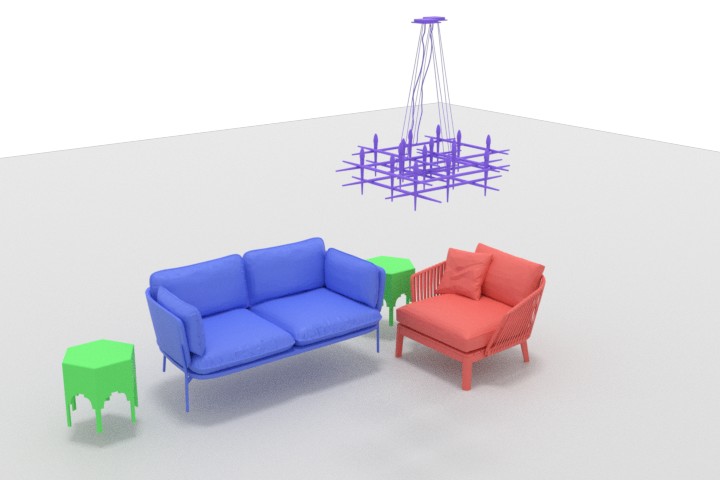}%
		\hfill
		\includegraphics[width=0.49\textwidth]
		{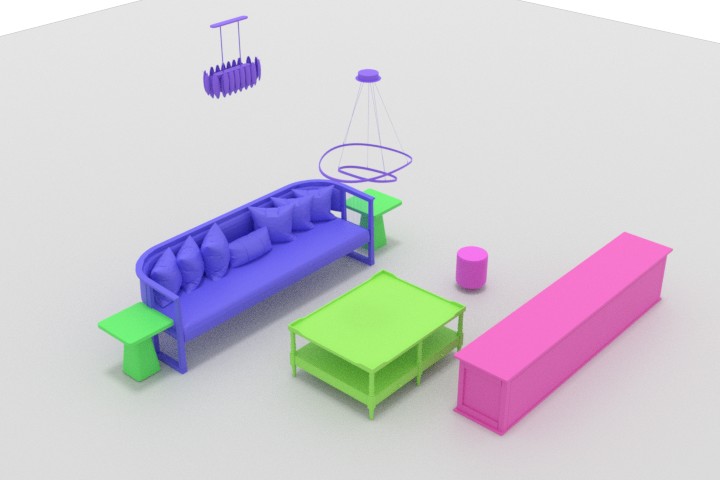}
		\caption{Sync2Gen~\cite{yang2021scene}}
	\end{subfigure}
	\rulesep
	\begin{subfigure}[t]{0.23\textwidth}
		\includegraphics[width=0.49\textwidth]
		{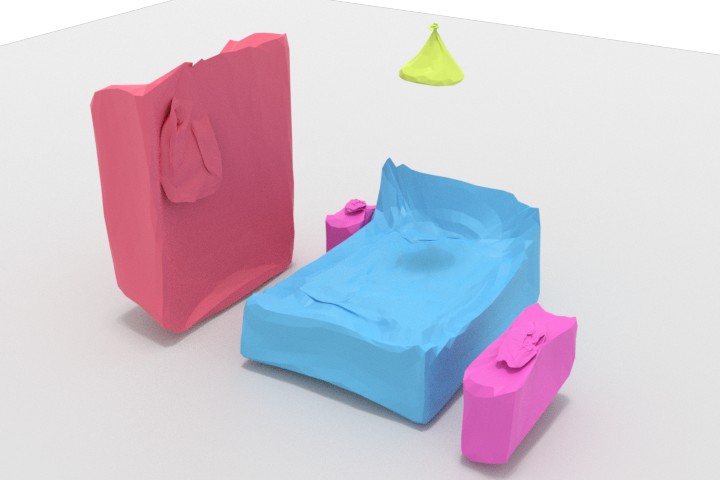}%
		\hfill
		\includegraphics[width=0.49\textwidth]
		{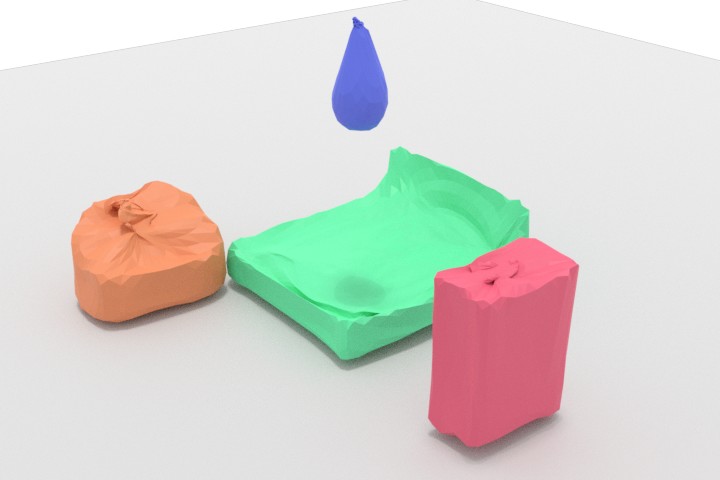}
		\includegraphics[width=0.49\textwidth]
		{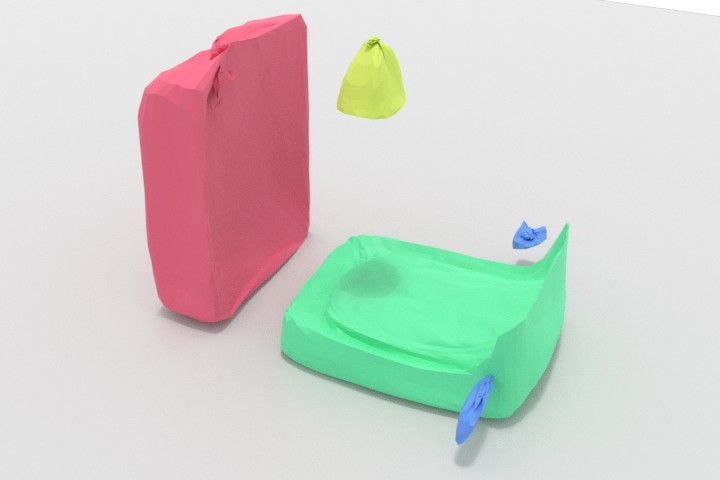}%
		\hfill
		\includegraphics[width=0.49\textwidth]
		{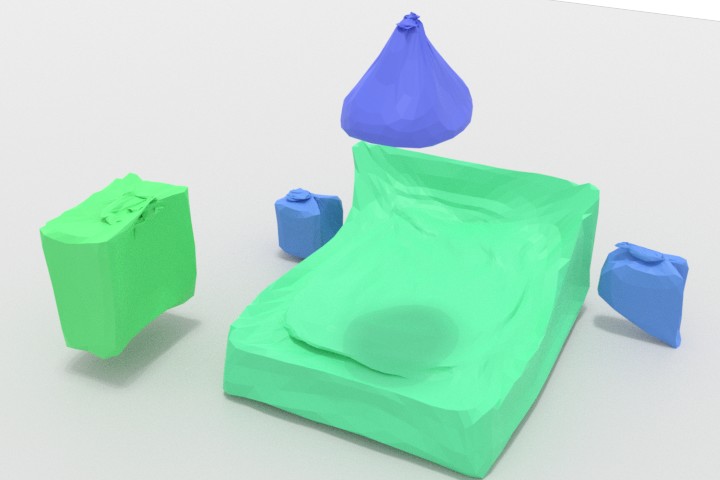}
		\includegraphics[width=0.49\textwidth]
		{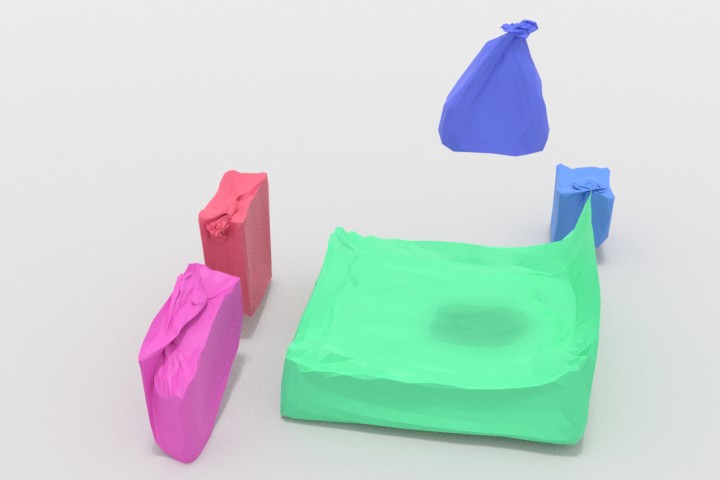}%
		\hfill
		\includegraphics[width=0.49\textwidth]
		{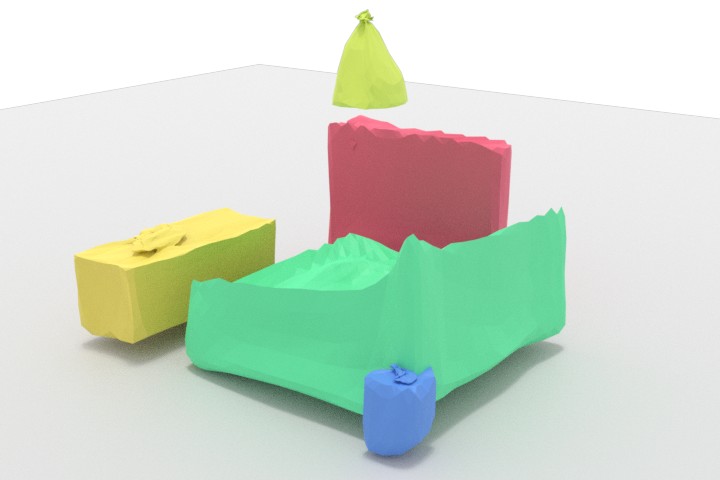}
		\includegraphics[width=0.49\textwidth]
		{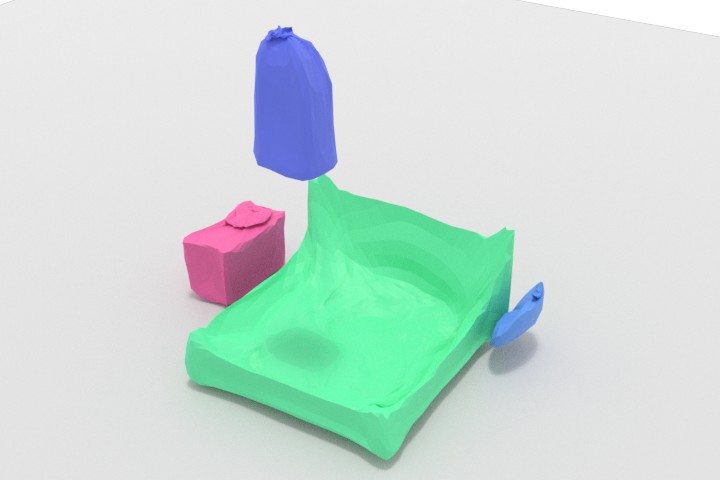}%
		\hfill
		\includegraphics[width=0.49\textwidth]
		{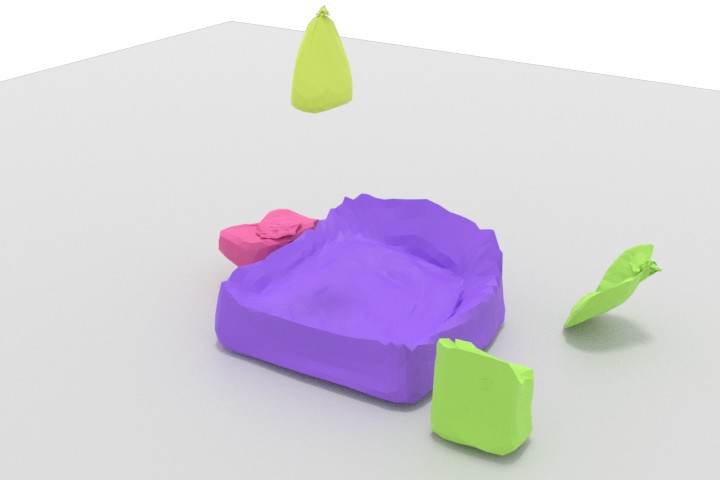}
		\includegraphics[width=0.49\textwidth]
		{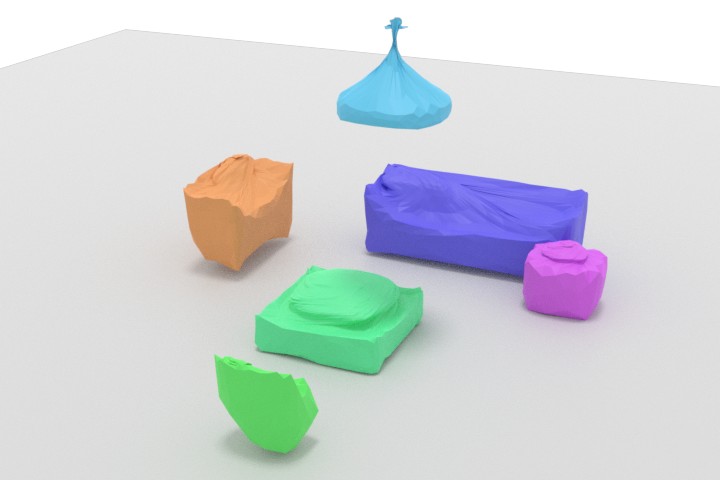}%
		\hfill
		\includegraphics[width=0.49\textwidth]
		{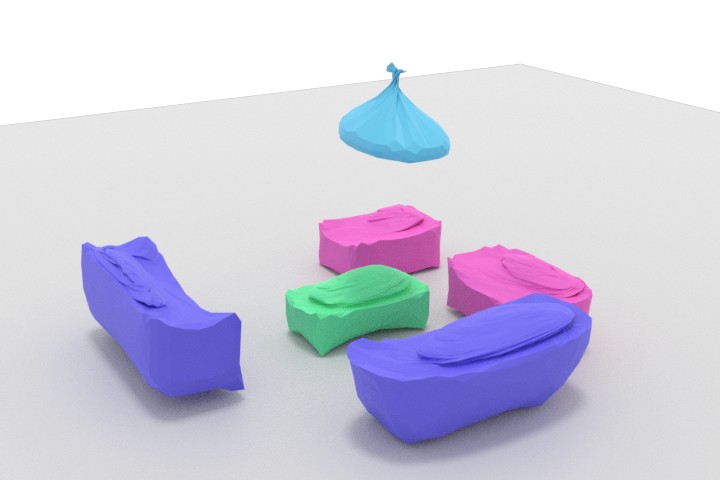}
		\includegraphics[width=0.49\textwidth]
		{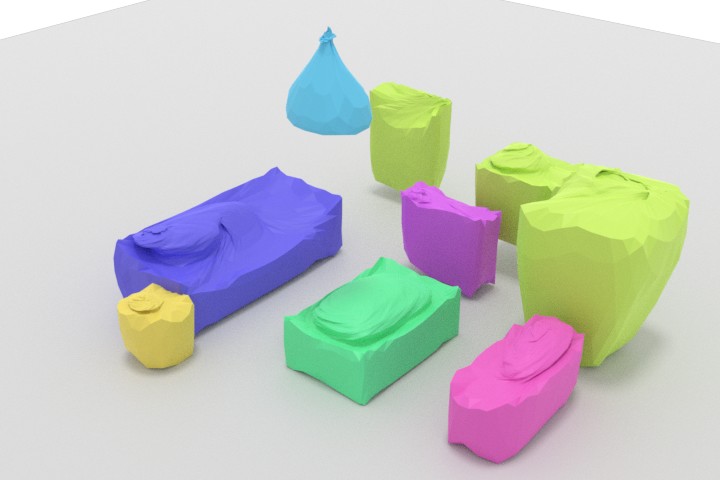}%
		\hfill
		\includegraphics[width=0.49\textwidth]
		{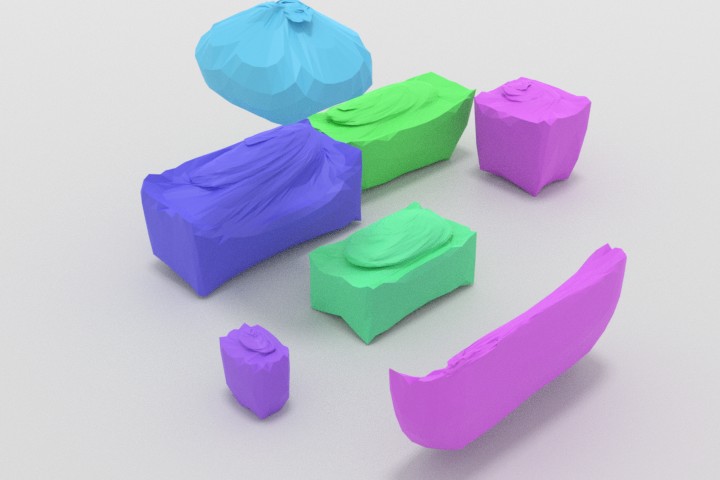}
		\includegraphics[width=0.49\textwidth]
		{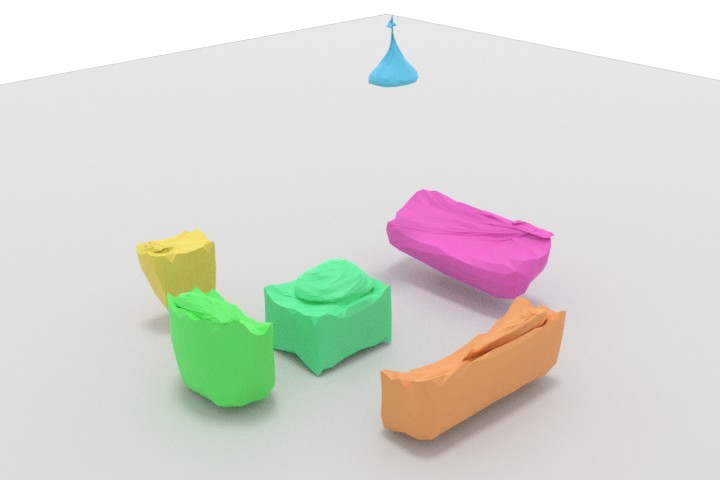}%
		\hfill
		\includegraphics[width=0.49\textwidth]
		{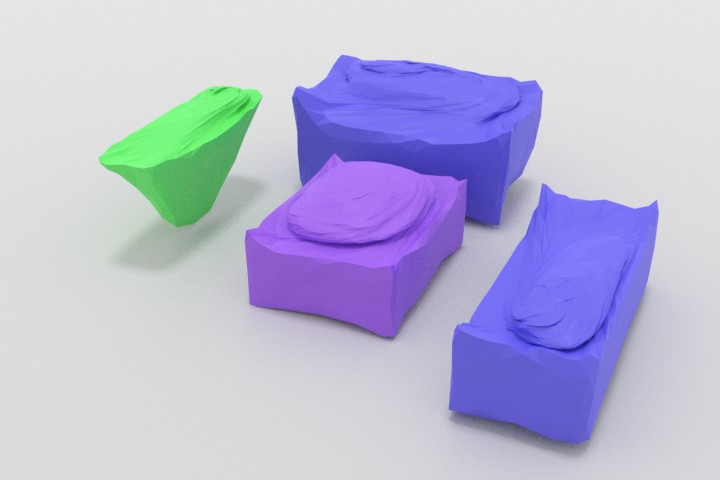}
		\caption{Ours (w/o retrieval)}
	\end{subfigure}
	\rulesep
	\begin{subfigure}[t]{0.23\textwidth}
		\includegraphics[width=0.49\textwidth]
		{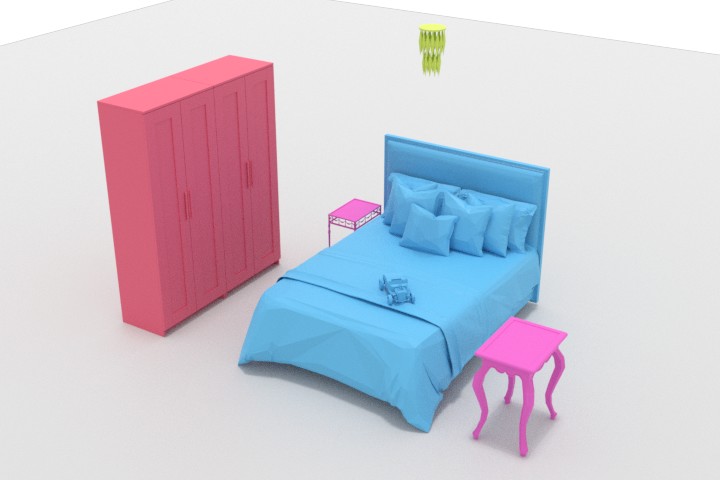}%
		\hfill
		\includegraphics[width=0.49\textwidth]
		{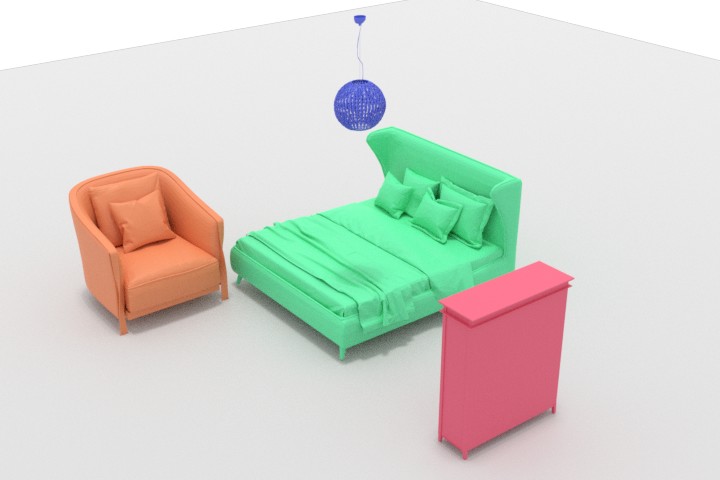}
		\includegraphics[width=0.49\textwidth]
		{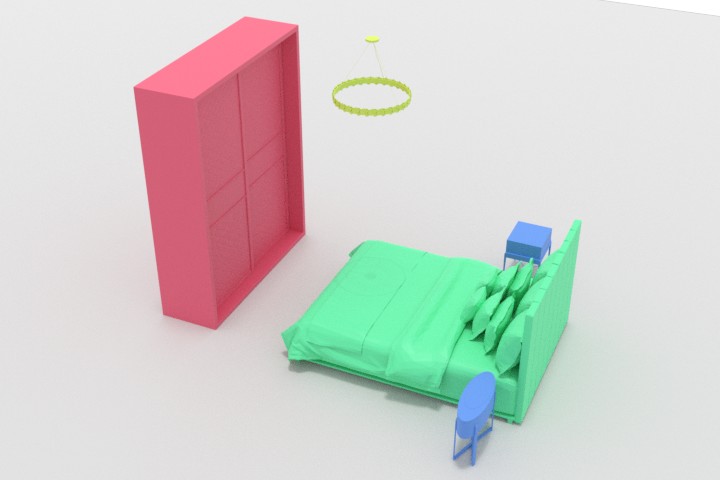}%
		\hfill
		\includegraphics[width=0.49\textwidth]
		{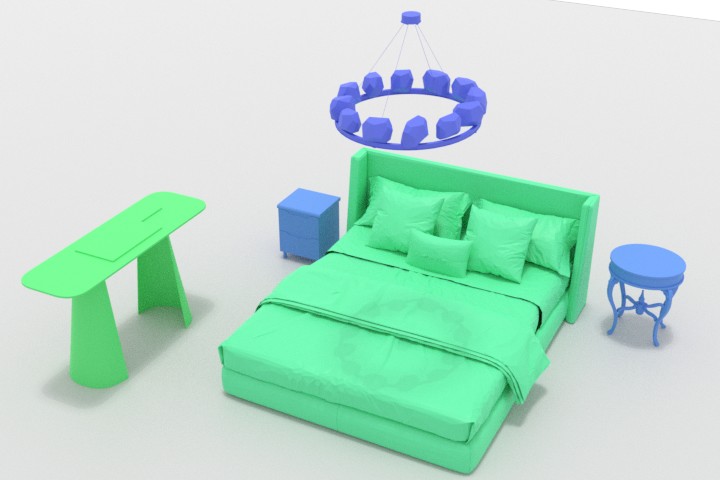}
		\includegraphics[width=0.49\textwidth]
		{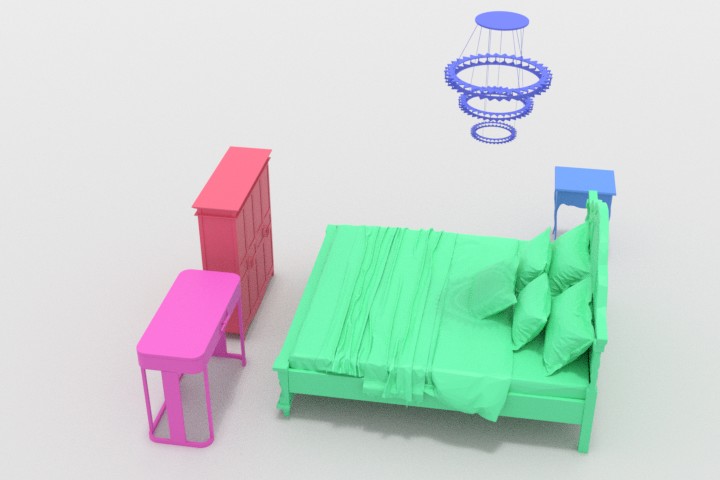}%
		\hfill
		\includegraphics[width=0.49\textwidth]
		{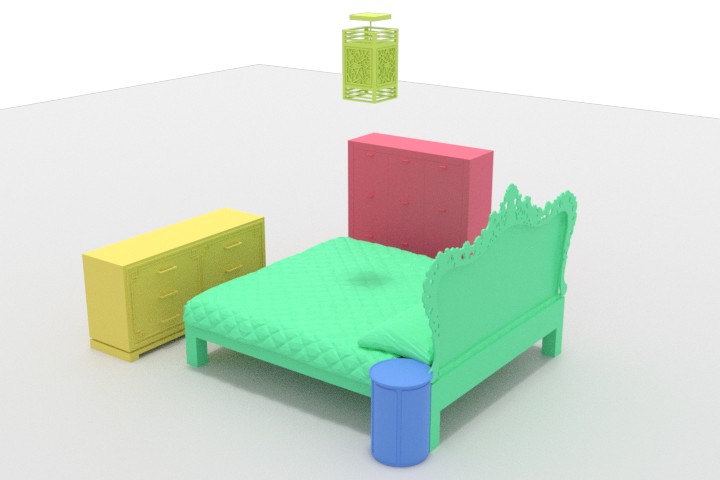}
		\includegraphics[width=0.49\textwidth]
		{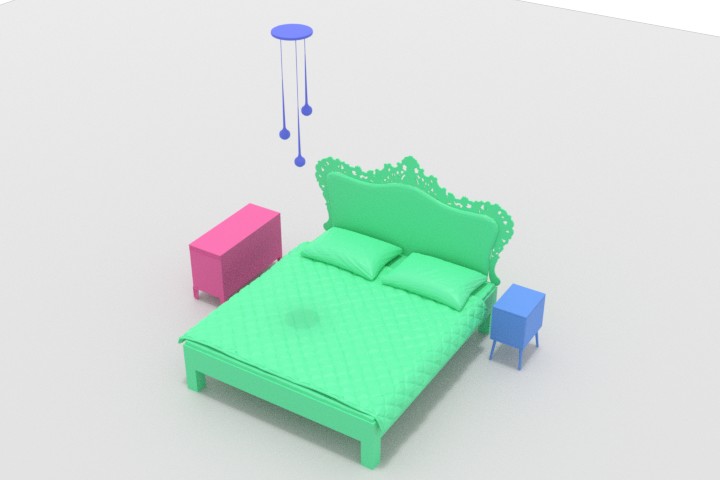}%
		\hfill
		\includegraphics[width=0.49\textwidth]
		{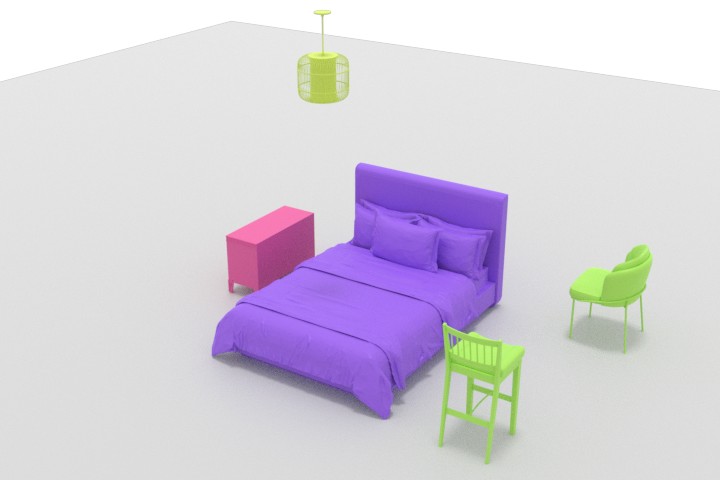}
		\includegraphics[width=0.49\textwidth]
		{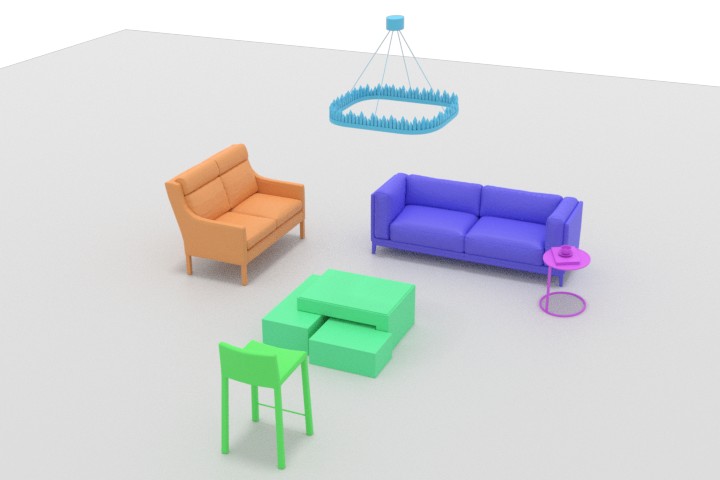}%
		\hfill
		\includegraphics[width=0.49\textwidth]
		{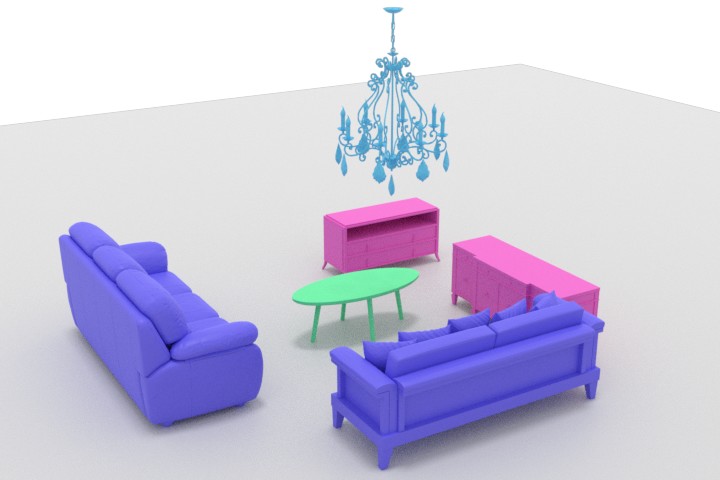}
		\includegraphics[width=0.49\textwidth]
		{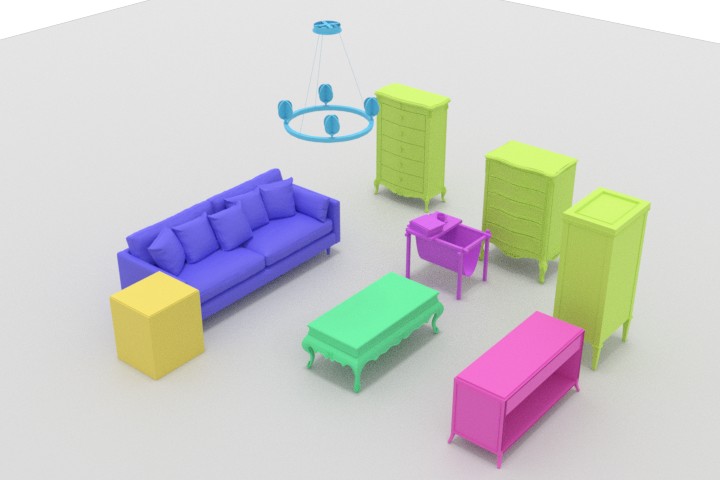}%
		\hfill
		\includegraphics[width=0.49\textwidth]
		{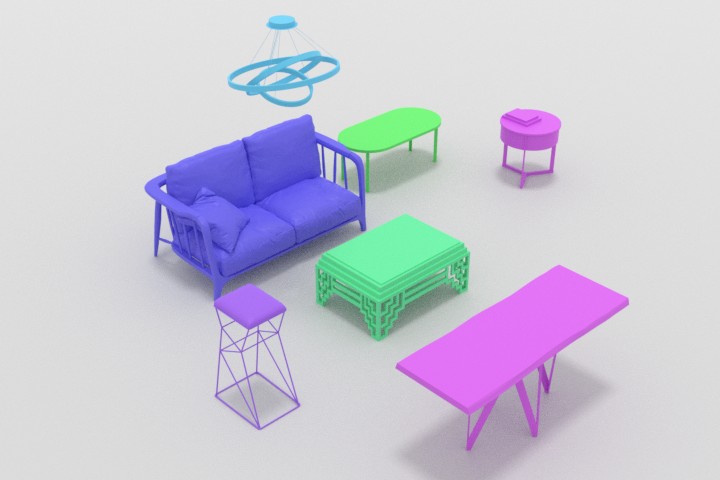}
		\includegraphics[width=0.49\textwidth]
		{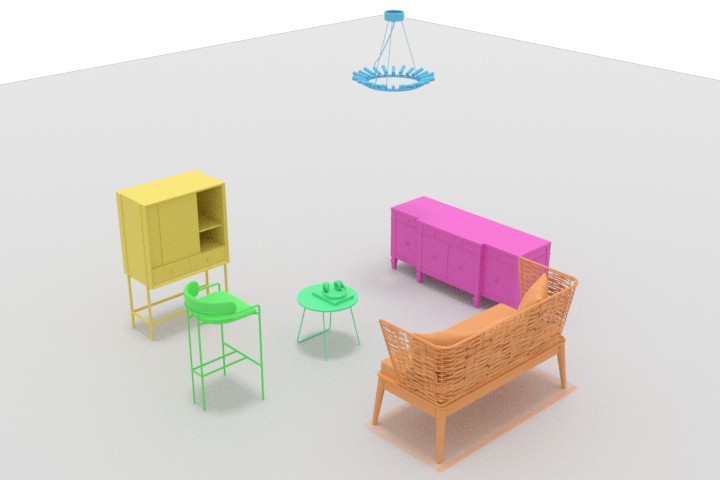}%
		\hfill
		\includegraphics[width=0.49\textwidth]
		{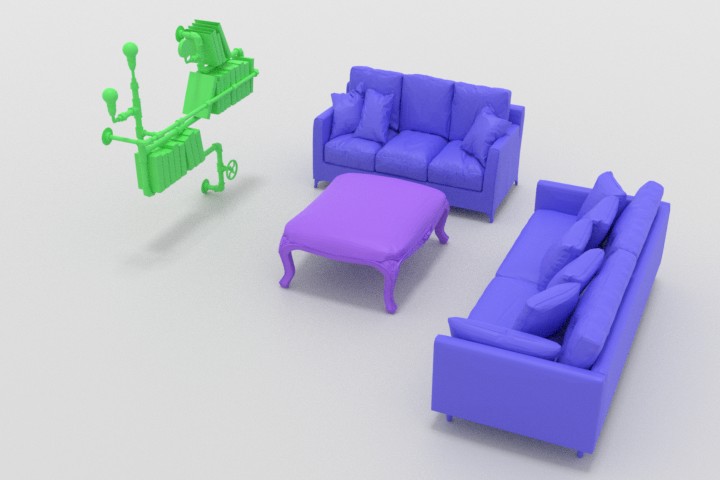}
		\caption{Ours}
	\end{subfigure}
	\caption{Qualitative comparisons with the state-of-the-art on scene synthesis. Note that both ATISS~\cite{paschalidou2021atiss} and Sync2Gen~\cite{yang2021scene} use 3D data for training while our method learns with only 2D supervision.}
	\label{fig:qualitative_scene_gen}
	\vspace{-1em}
\end{figure*}

\section{Experiment Setup}
\paragraph{Datasets and Tasks} We train and test our method on two downstream tasks: 3D scene synthesis and single-view scene reconstruction.

For scene synthesis, we use the 3D-FRONT~\cite{fu20213d} dataset for training and evaluation as in~\cite{Paschalidou2021NEURIPS,yang2021scene}. 3D-FRONT is a synthetic indoor scene dataset with 18,968 rooms populated with 3D furniture objects, wherein we choose bedrooms and living rooms for our training and testing. We use \cite{denninger2019blenderproc} to sample virtual camera poses in each room and render the 2D instance masks for objects under different views, which produces 102,268 views in 3,922 bedrooms ($\approx$5.53 objects in each room), and 80,997 views in 1,496 living rooms ($\approx$9.82 objects in each room). For each room type, the train/test split ratio is 9:1 over rooms. We provide additional rendering details in the supplemental.

For single-view scene reconstruction, we use the ScanNet dataset~\cite{dai2017scannet} for benchmarking. ScanNet is a real indoor scene dataset with 1,513 rooms in 21 room types. It provides 3D scans with semantic instance labels, as well as 2D scanning data where each room is captured into an RGB video sequence, and each frame contains a calibrated camera pose and 2D instance masks with tracking IDs. We uniformly sample up to 100 frames on each video and obtains 108,059 valid frames from all scenes. We use the official train/val split for our training and evaluation.

\vspace{-1em}
\paragraph{Implementation} For each task, we train our network on 4 NVIDIA 2080 Ti GPUs with batch size at 16. Each sample in a batch contains a scene with 20 randomly sampled views. We augment each scene in 3D-FRONT and ScanNet with axis-aligned rotations by 90, 180, 270 degrees. As in Sec.~\ref{sec:view_loss}, we train our network end-to-end with two stages. We first train the layout learning in first 800 epochs, followed by the shape learning with 500 epochs. We use Adam as the optimizer with initial learning rate 1e-4, which is decayed by 0.1$\times$ after 300 epochs, in both of the two stages. The losses are weighted by $\lambda_{l}$=$\lambda_{p}$=$\lambda_{f}$=1, $\lambda_{box}$=5, $\lambda_{S}$=2 to balance loss values. For layer and differentiable rendering specifications, we refer to the supplemental.

For scene synthesis, we randomly sample latent vectors by Eq.~\ref{eq:latent_vec} to synthesize diverse scenes. For single-view reconstruction, we freeze the generation network and take the single-view instance masks as the ground-truth to optimize the latent vector with our loss function. After it converges, we generate the 3D scene with the learned latent vector. We provide additional details about this optimization step in the supplemental.

\vspace{-1em}
\paragraph{Evaluation Metrics} We evaluate our method on the tasks of scene synthesis and single-view scene reconstruction using 3D-FRONT and ScanNet respectively. For scene synthesis, we generate 1000 scenes for each method and use FID score (FID), scene classification accuracy (SCA) and category KL divergence (KL) from \cite{paschalidou2021atiss} to measure the generation plausibility against the ground-truth in the test set. We refer readers to \cite{paschalidou2021atiss} for their definitions. For FID and SCA calculation, we render the generated and the ground-truth scenes into 512\textsuperscript{2} images with a top-down orthographic projection, where each object is textured with a unique category-specific color. We unify this setting among all baselines for a fair comparison.

For single-view reconstruction, we use the metrics of 3D box IoU and chamfer distance between generated objects and the ground-truth. 3D box IoU measures the layout accuracy by comparing the 3D object bounding box between each ground-truth object and the prediction. Chamfer distance measures the accuracy of both layout and shape prediction. It calculates the point-wise $L_{2}$ distance between generated object surface points and the ground-truth.

\section{Results and Analysis}
We evaluate our method qualitatively and quantitatively on the tasks of scene synthesis and single-view reconstruction in comparisons with state-of-the-art baselines, as well as scene interpolation and an ablation analysis on different network configurations.
\subsection{Scene Synthesis}
\label{sec:exp_scene_gen}
\paragraph{Baselines}
Since there are no previous works that synthesize 3D semantic instance scenes by learning from only 2D supervision, for a fair comparison, we use several state-of-the-art baselines while train them under 2D supervision. Additionally, we also compare with the state-of-the-art methods which use 3D training data, to explore the boundary of scene synthesis by learning with only 2D supervision. We consider the following baselines: \textbf{1)} ATISS~\cite{Paschalidou2021NEURIPS}, a transformer encoder-based method that learns the distribution of 3D object bounding box properties. By progressively sampling from each property distribution, it generates a sequence of object boxes as the output. \textbf{2)} Sync2Gen~\cite{yang2021scene}, a VAE-based method that maps a scene arrangement (parameterized with a sequence of 3D object attributes) to itself. After training, a 3D scene can be generated by sampling from the bottleneck latent distribution. \textbf{3)} ATISS-2D. We use the transformer in ATISS~\cite{Paschalidou2021NEURIPS} as the backbone and adapt it to be trainable under 2D supervision with our view loss. \textbf{4)} GAN. We customize the generative adversary network for semantic scene synthesis from \cite{yang2021indoor} to make it trainable with only 2D supervision. \textbf{5)} LSTM. We replace our transformer with an LSTM network, while keeping other modules unchanged. Detailed architecture specifications are given in the supplemental. We train/test all methods on the same split for fair comparisons.
\vspace{-1em}
\paragraph{Qualitative Comparisons}
Fig.~\ref{fig:qualitative_scene_gen} visualizes the qualitative comparisons with ATISS~\cite{Paschalidou2021NEURIPS} and Sync2Gen~\cite{yang2021scene}, which are the state-of-the-art scene synthesis methods using 3D object data for training. While our method learns with only 2D supervision. Both the two methods add a shape retrieval post-processing to obtain object CAD models for evaluation. For fair comparison, we adopt shape retrieval by using Chamfer distance to measure the similarity between our reconstructed shape and CAD models in 3D-FRONT. We provide the details in the supplemental. From Fig.~\ref{fig:qualitative_scene_gen}, we observe that ATISS is vulnerable to the penetration issues between object bounding boxes (row 2,4,5,6). Sync2Gen addresses this by involving a Bayesian optimization post-processing to refine object placement based on a 3D object arrangement prior, but it is still vulnerable to penetration issues (row 1,2,6). Our method learns object layout and shape from only 2D data, where we learn each object conditioned on the existence of previous generated objects under the multi-view constraint, which helps to avoid object intersections and produces plausible scene arrangements.

\vspace{-1em}
\paragraph{Quantitative Comparisons}
Tab.~\ref{tab:quant_scene_gen} presents the quantitative comparisons between methods under different supervision sources. We observe that our method achieves better FID, SCA and KL scores against these methods using the same 2D supervision, which indicates that we produce better generation plausibility and more realistic object category distribution. To explore the boundary of our method, we also compare with  prior arts which use 3D supervision. The results show that our method still presents better visual plausibility (FID and SCA scores) than ATISS, and shows comparable performance against Sync2Gen.

\begin{table}[!t]
	\centering
	\resizebox{1\columnwidth}{!}{
		\begin{tabular}{|l|c|c c c|c c c|}
			\hline
			&  & \multicolumn{3}{c|}{Bedroom} & \multicolumn{3}{c|}{Living room}\\
			& Supervision & FID ($\downarrow$) & SCA & KL ($\downarrow$) & FID ($\downarrow$) & SCA & KL ($\downarrow$)  \\
			\hline
			\hline
			ATISS~\cite{Paschalidou2021NEURIPS} & 3D & 26.82 & 0.97 & \underline{0.01} & 50.02 & 0.97 & 0.02  \\
			Sync2Gen*~\cite{yang2021scene} & 3D & 25.24 & \underline{0.80} & 0.03 & \underline{35.34} & 0.88 & \underline{0.02} \\
			Sync2Gen~\cite{yang2021scene} & 3D & \underline{23.86} & 0.83 & 0.02 & 35.60 & \underline{0.86} & 0.08\\
			\hline
			\hline
			ATISS-2D~\cite{Paschalidou2021NEURIPS} & 2D & 38.13 & 0.96 & 0.10 & 65.07 & 0.98 & 0.36\\
			GAN~\cite{yang2021indoor} & 2D & 24.42 & 0.97 & 0.05
			& 44.76 & 0.99 & 0.08 \\
			LSTM & 2D & 67.97 & 1.00 & 0.30
			& 86.77 & 1.00 & 0.26 \\
			Ours & 2D & \textbf{21.59} & \textbf{0.85} & \textbf{0.03} & \textbf{40.47} & \textbf{0.96} & \textbf{0.02} \\
			\hline
	\end{tabular}}
	\caption{Quantitative comparisons on scene synthesis. Note that SCA score closer to 0.5 is better. Sync2Gen* is the version of ~\cite{yang2021scene} without Bayesian optimization. ATISS-2D is customized from ATISS~\cite{Paschalidou2021NEURIPS} while using our 2D view loss. GAN is customized from \cite{yang2021indoor} with using only 2D supervision.}
	\label{tab:quant_scene_gen}
	\vspace{-1em}
\end{table}

\begin{figure*}[!ht]
	\centering
	\begin{subfigure}[t]{0.105\textwidth}
		\includegraphics[width=\textwidth]
		{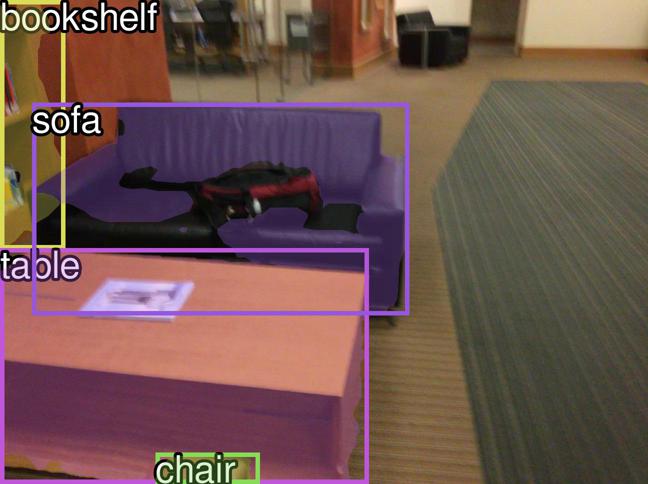}
		\includegraphics[width=\textwidth]
		{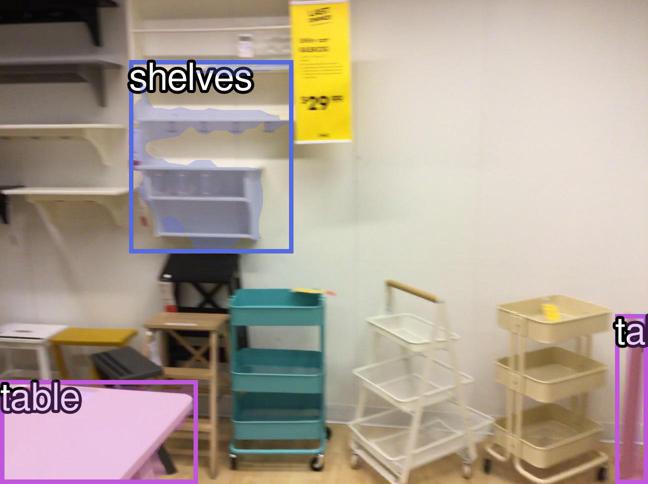}
		\includegraphics[width=\textwidth]
		{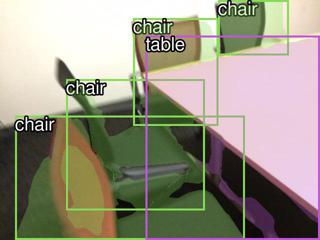}
		\includegraphics[width=\textwidth]
		{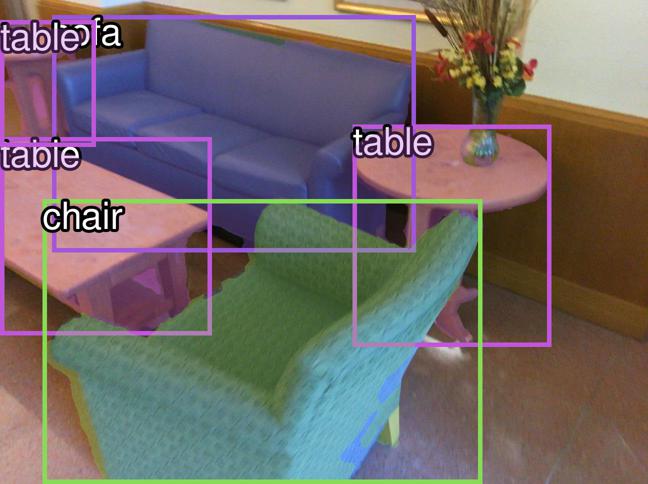}
		\includegraphics[width=\textwidth]
		{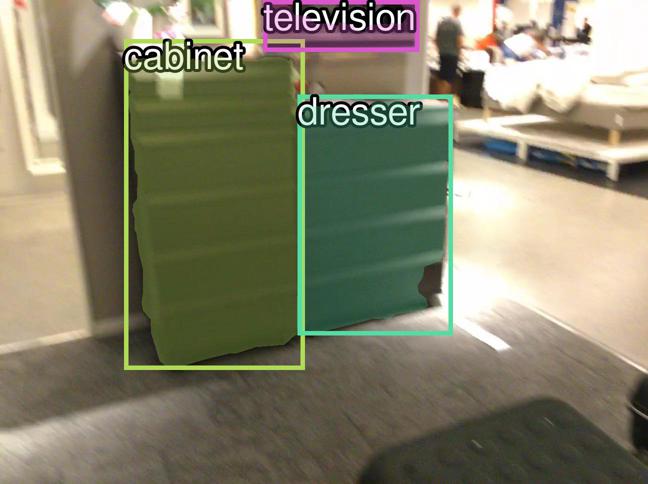}
		\includegraphics[width=\textwidth]
		{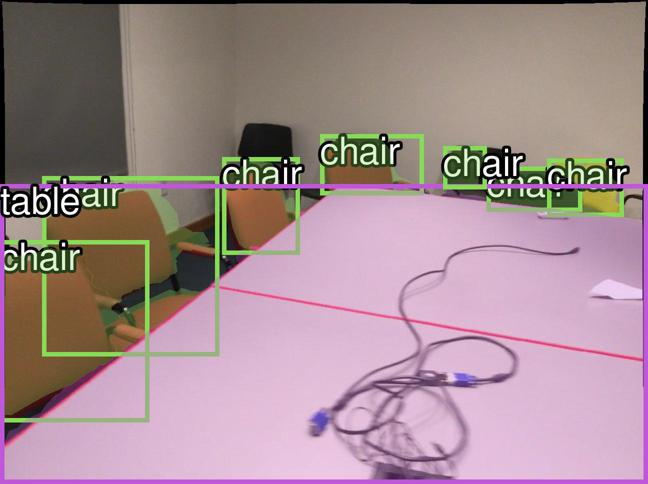}
		\includegraphics[width=\textwidth]
		{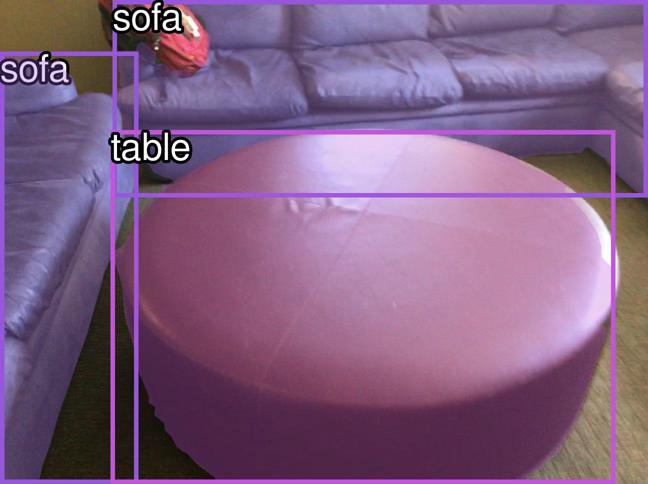}
		\caption{Input}
	\end{subfigure}
	\rulesep
	\begin{subfigure}[t]{0.213\textwidth}
		\includegraphics[width=0.49\textwidth]
		{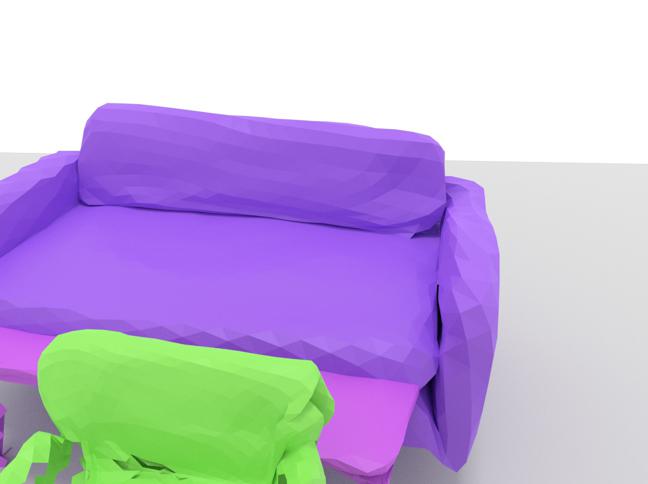}%
		\hfill
		\includegraphics[width=0.49\textwidth]
		{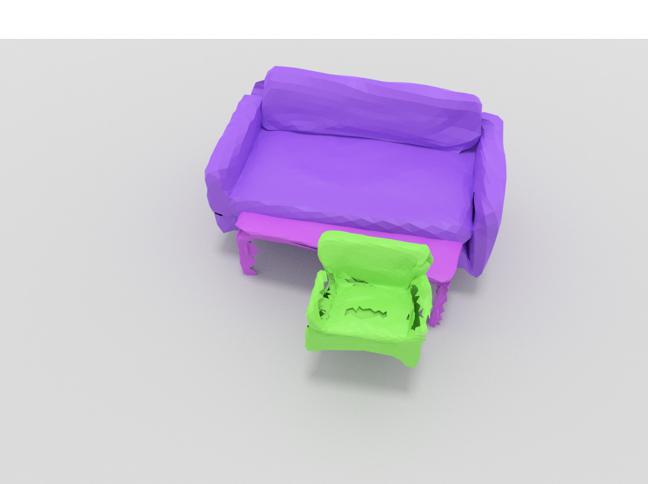}
		\includegraphics[width=0.49\textwidth]
		{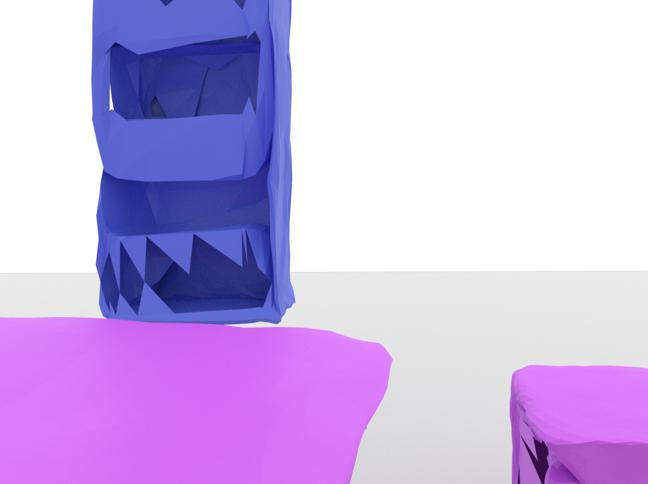}%
		\hfill
		\includegraphics[width=0.49\textwidth]
		{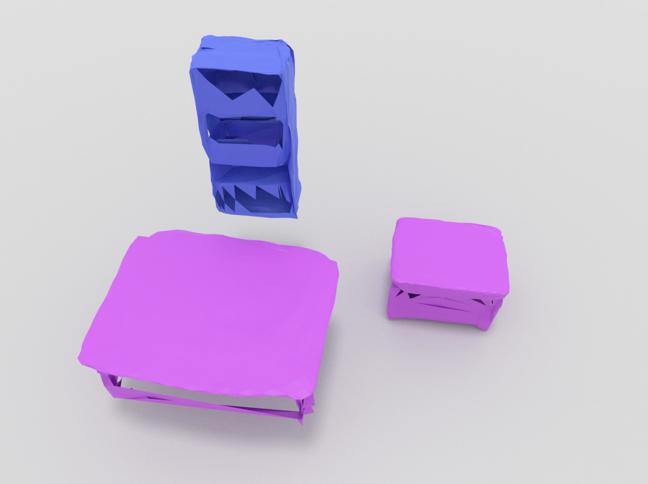}
		\includegraphics[width=0.49\textwidth]
		{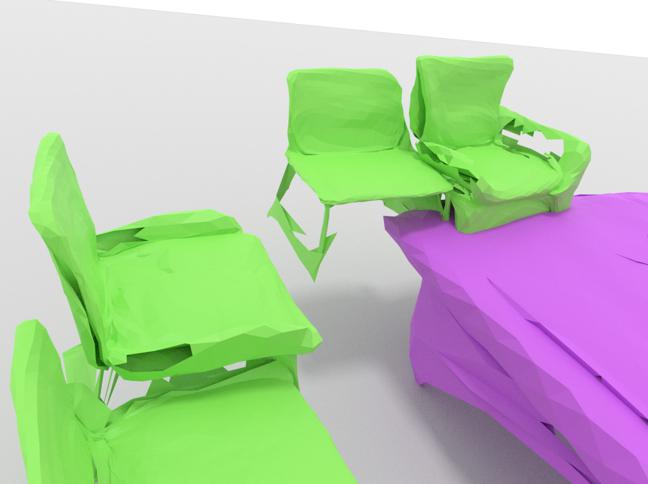}%
		\hfill
		\includegraphics[width=0.49\textwidth]
		{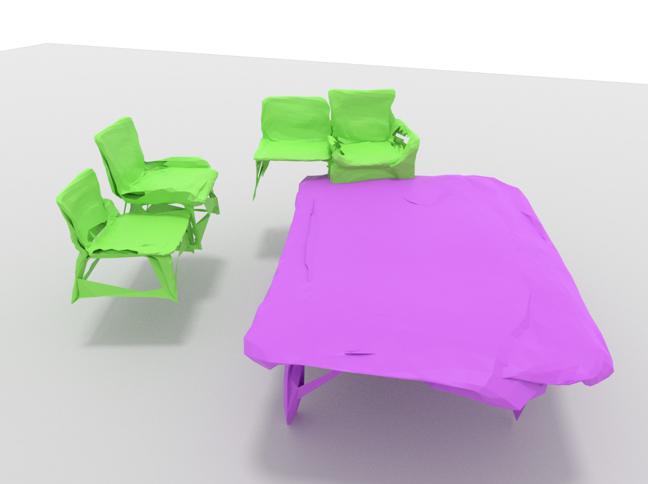}
		\includegraphics[width=0.49\textwidth]
		{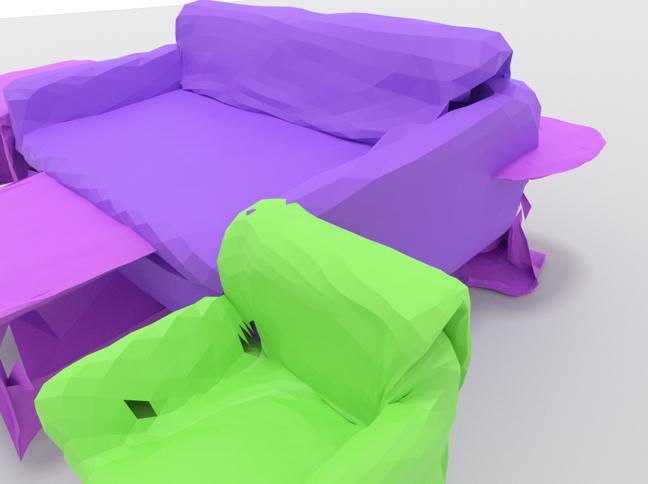}%
		\hfill
		\includegraphics[width=0.49\textwidth]
		{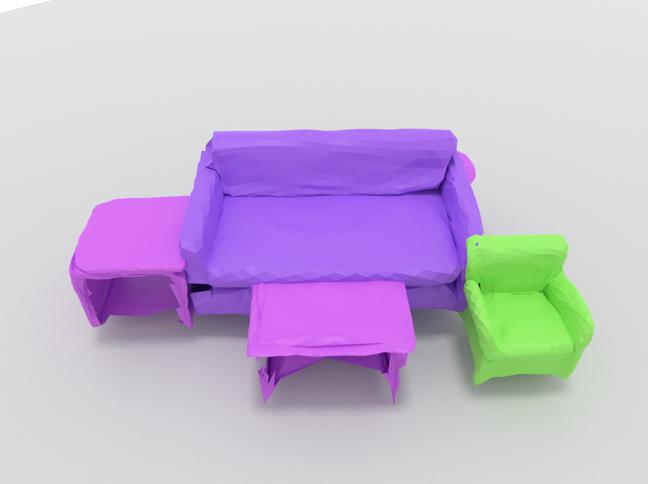}
		\includegraphics[width=0.49\textwidth]
		{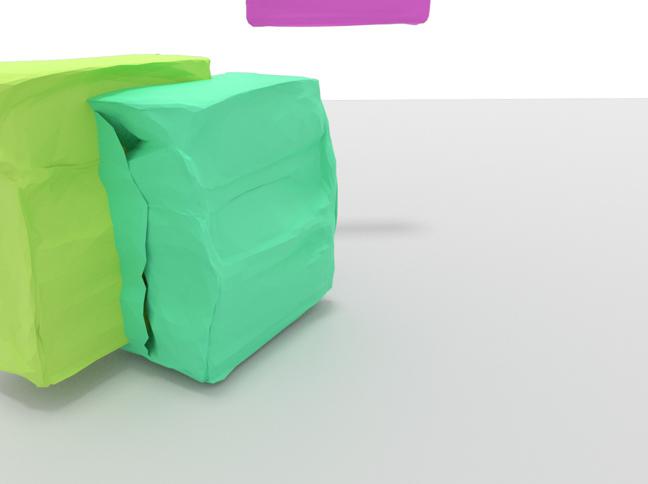}%
		\hfill
		\includegraphics[width=0.49\textwidth]
		{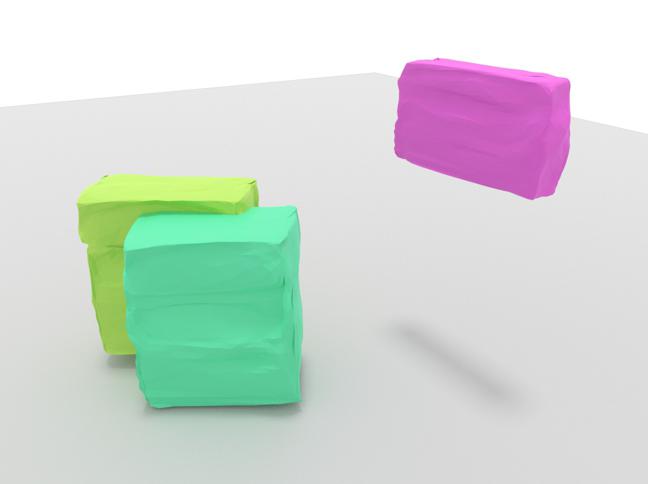}
		\includegraphics[width=0.49\textwidth]
		{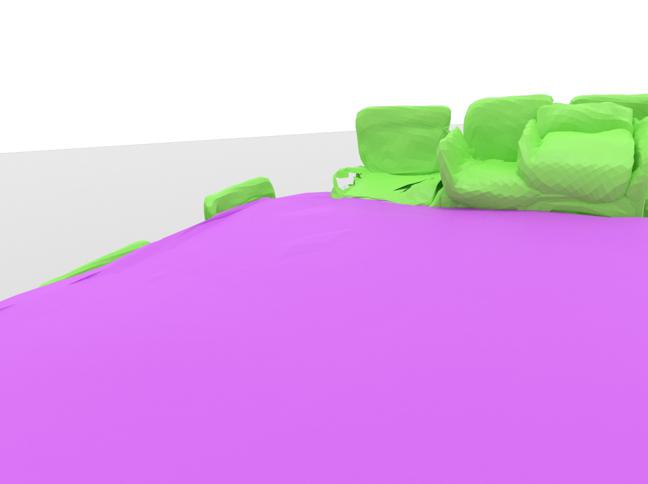}%
		\hfill
		\includegraphics[width=0.49\textwidth]
		{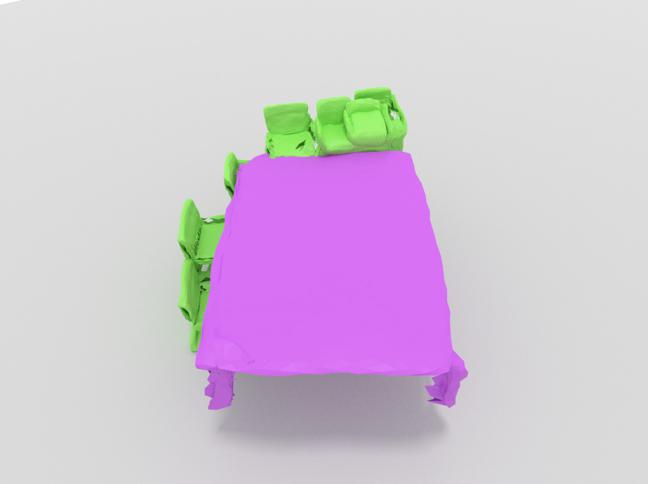}
		\includegraphics[width=0.49\textwidth]
		{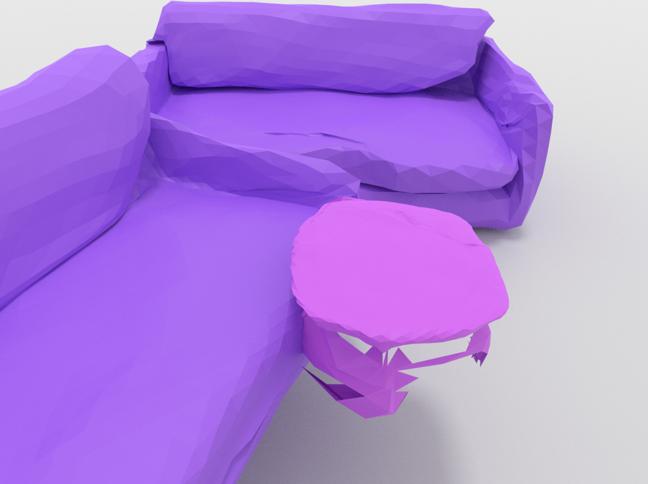}%
		\hfill
		\includegraphics[width=0.49\textwidth]
		{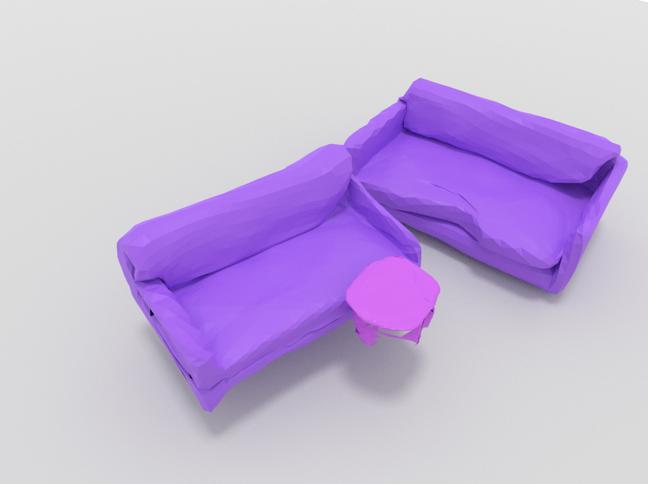}
		\caption{Total3D~\cite{nie2020total3dunderstanding}}
	\end{subfigure}
	\rulesep
	\begin{subfigure}[t]{0.213\textwidth}
		\includegraphics[width=0.49\textwidth]
		{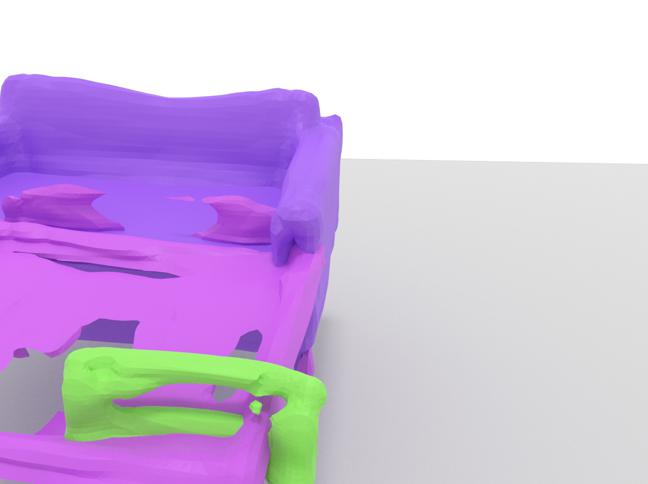}%
		\hfill
		\includegraphics[width=0.49\textwidth]
		{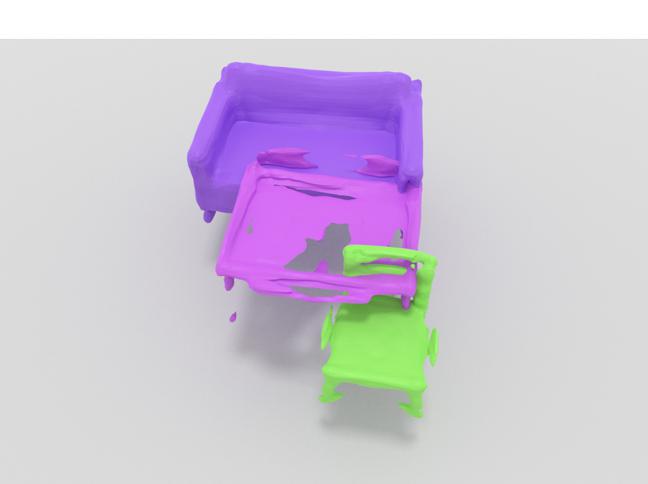}
		\includegraphics[width=0.49\textwidth]
		{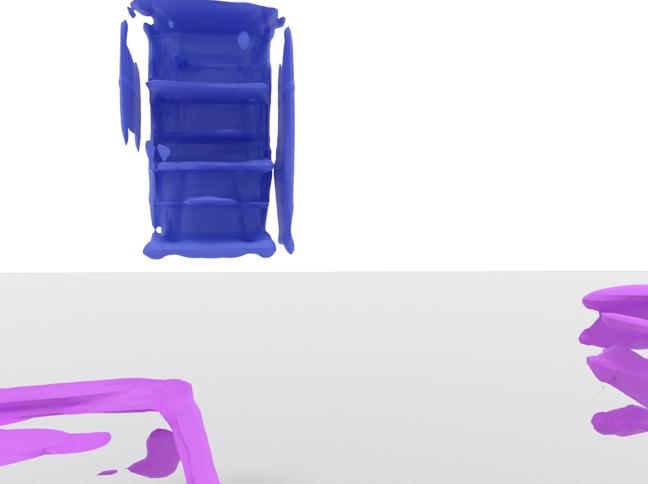}%
		\hfill
		\includegraphics[width=0.49\textwidth]
		{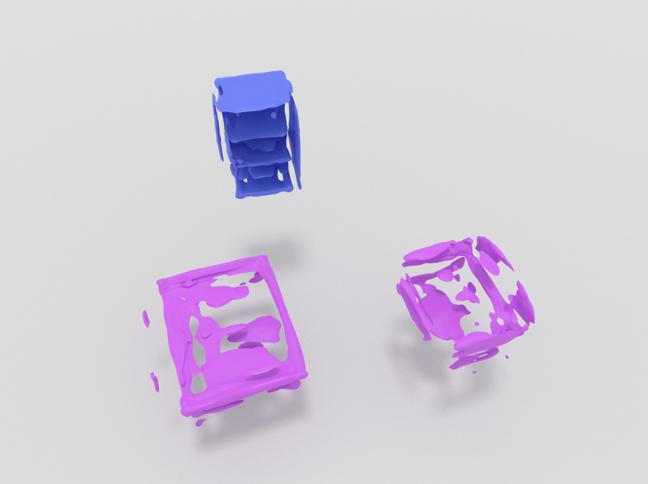}
		\includegraphics[width=0.49\textwidth]
		{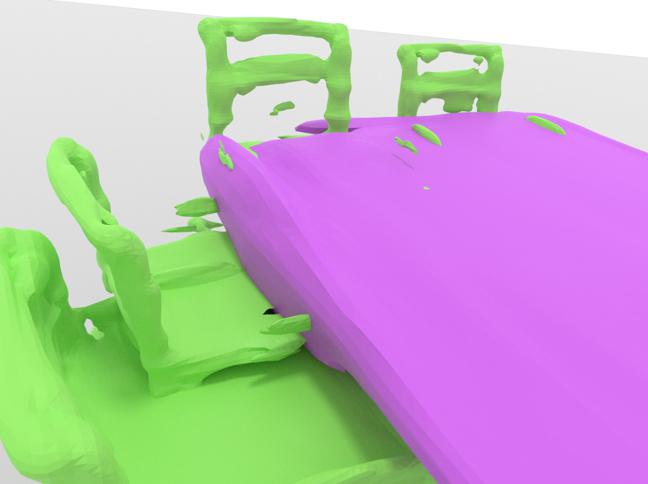}%
		\hfill
		\includegraphics[width=0.49\textwidth]
		{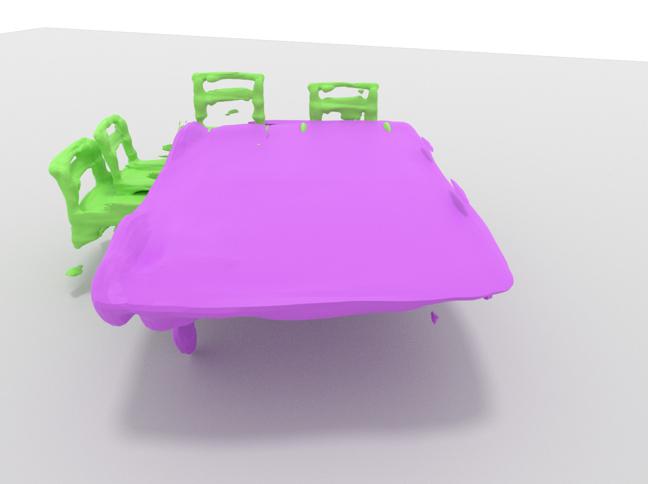}
		\includegraphics[width=0.49\textwidth]
		{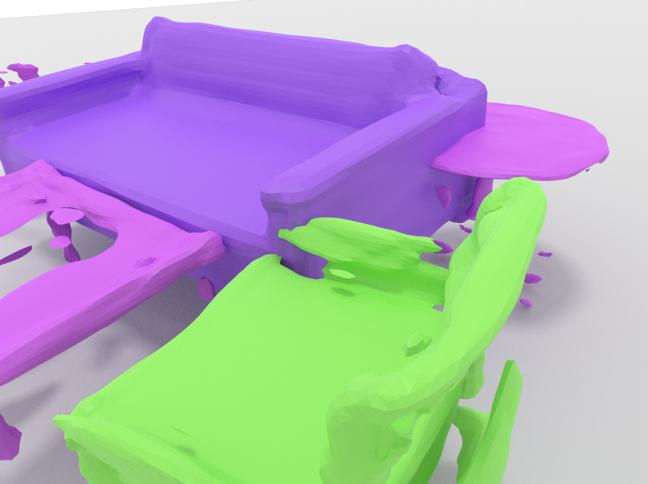}%
		\hfill
		\includegraphics[width=0.49\textwidth]
		{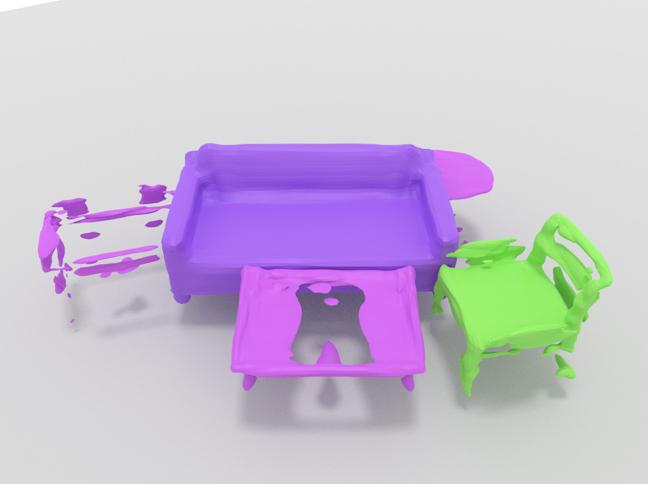}
		\includegraphics[width=0.49\textwidth]
		{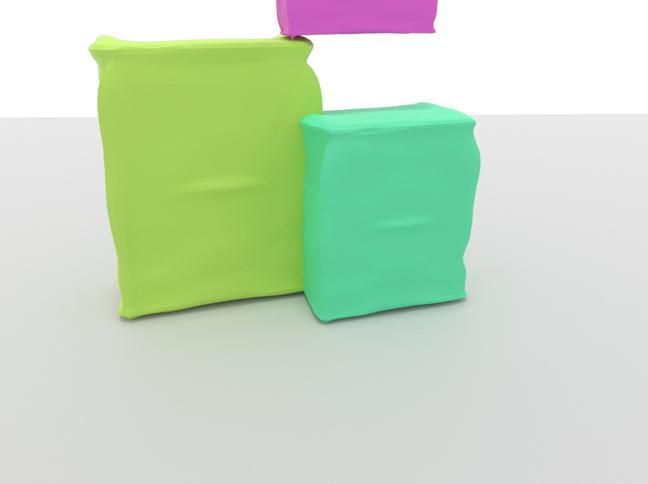}%
		\hfill
		\includegraphics[width=0.49\textwidth]
		{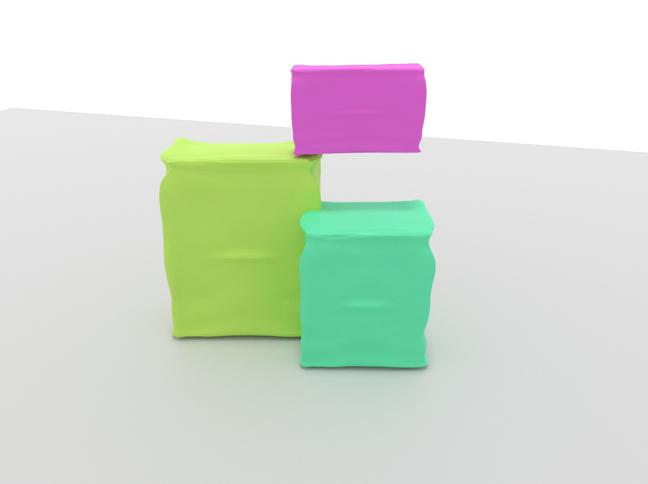}
		\includegraphics[width=0.49\textwidth]
		{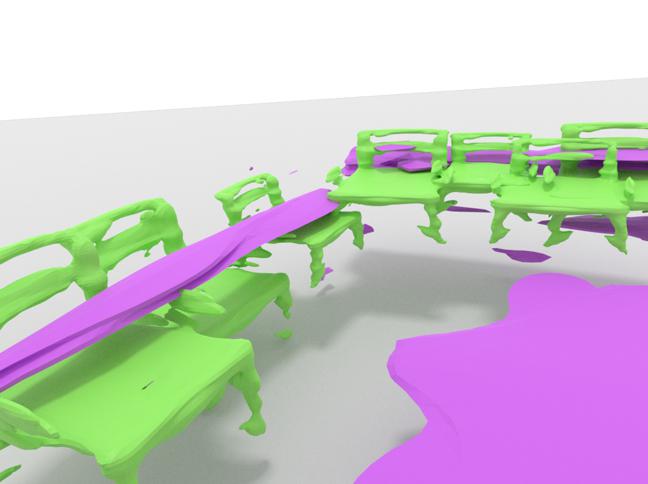}%
		\hfill
		\includegraphics[width=0.49\textwidth]
		{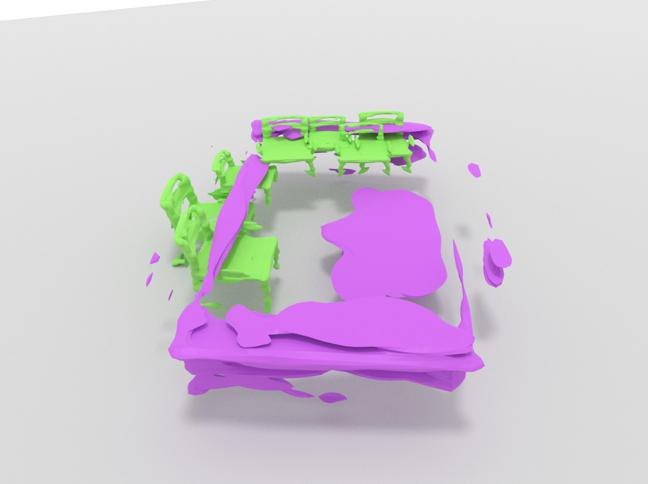}
		\includegraphics[width=0.49\textwidth]
		{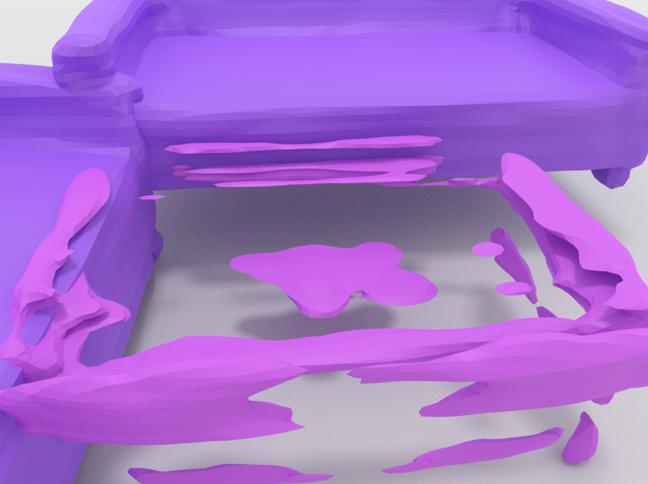}%
		\hfill
		\includegraphics[width=0.49\textwidth]
		{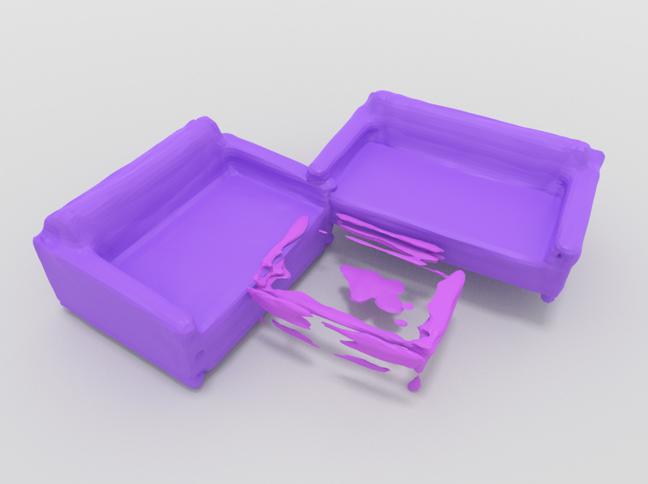}
		\caption{Im3D~\cite{zhang2021holistic}}
	\end{subfigure}
	\rulesep
	\begin{subfigure}[t]{0.32\textwidth}
		\includegraphics[width=0.325\textwidth]
		{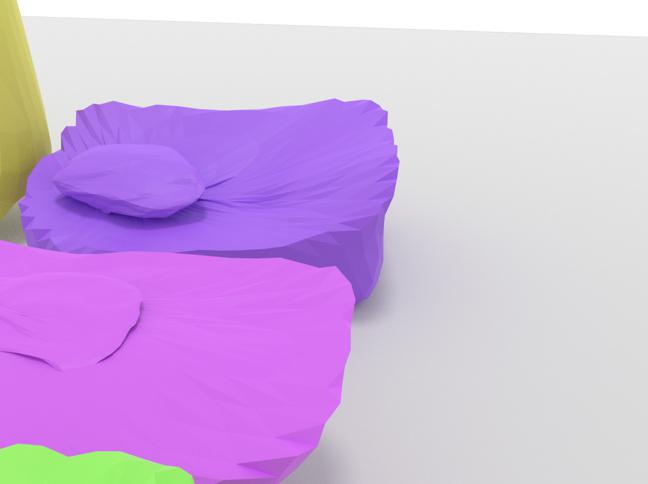}%
		\hfill
		\includegraphics[width=0.325\textwidth]
		{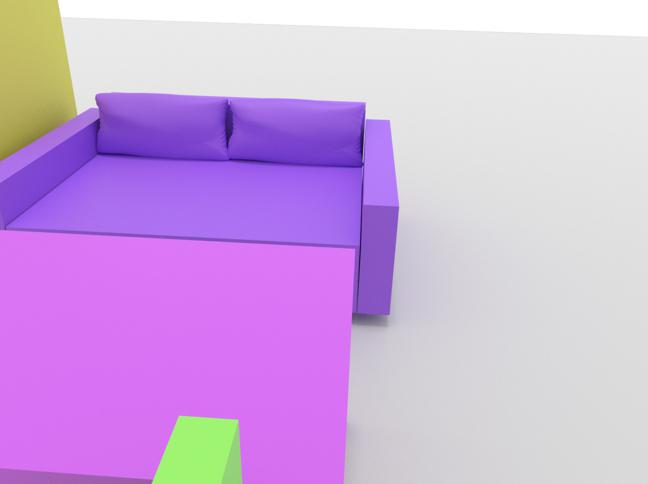}%
		\hfill
		\includegraphics[width=0.325\textwidth]
		{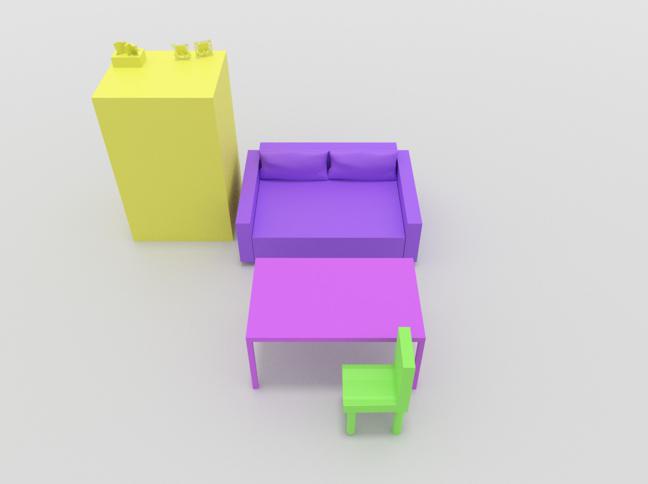}
		\includegraphics[width=0.325\textwidth]
		{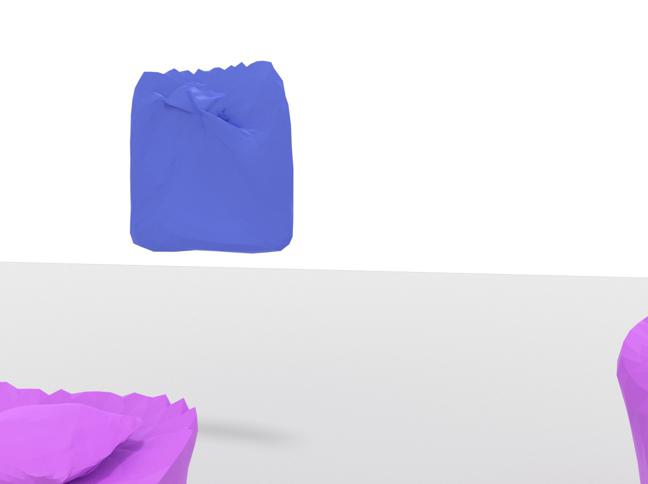}%
		\hfill
		\includegraphics[width=0.325\textwidth]
		{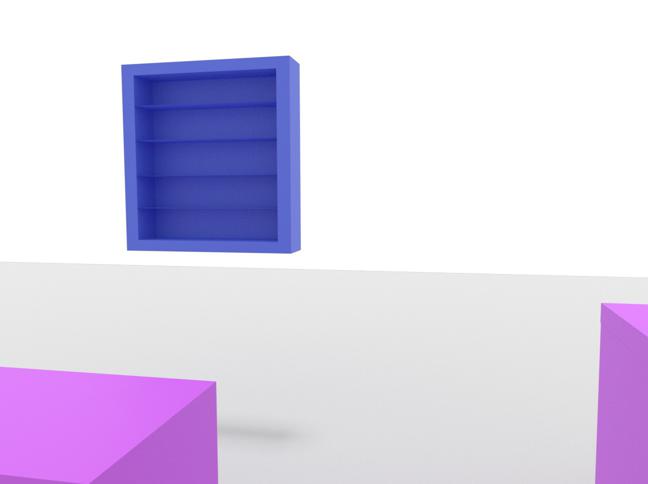}%
		\hfill
		\includegraphics[width=0.325\textwidth]
		{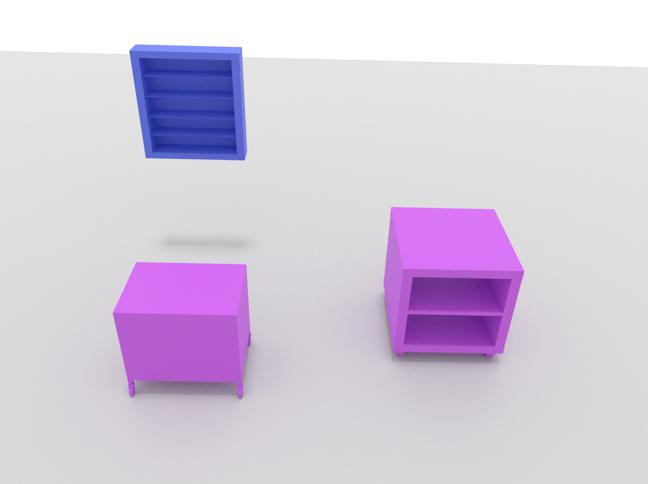}
		\includegraphics[width=0.325\textwidth]
		{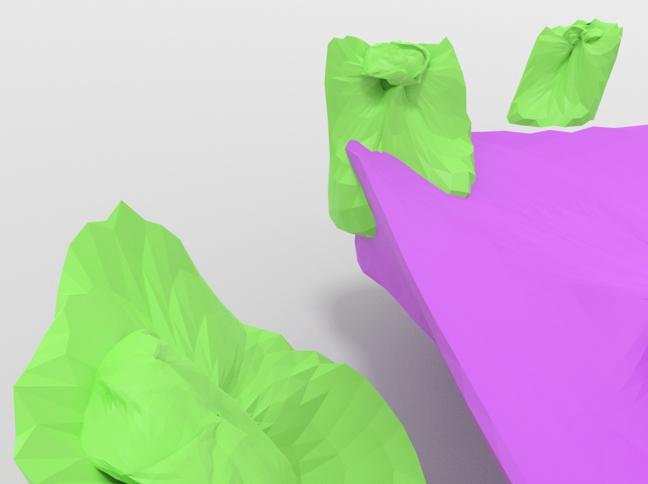}%
		\hfill
		\includegraphics[width=0.325\textwidth]
		{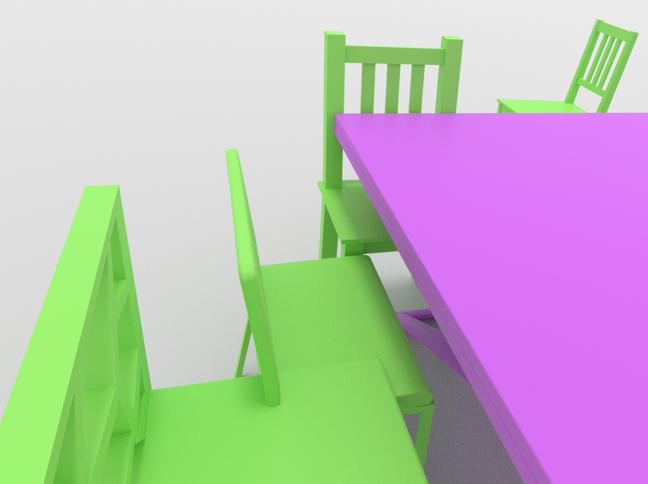}%
		\hfill
		\includegraphics[width=0.325\textwidth]
		{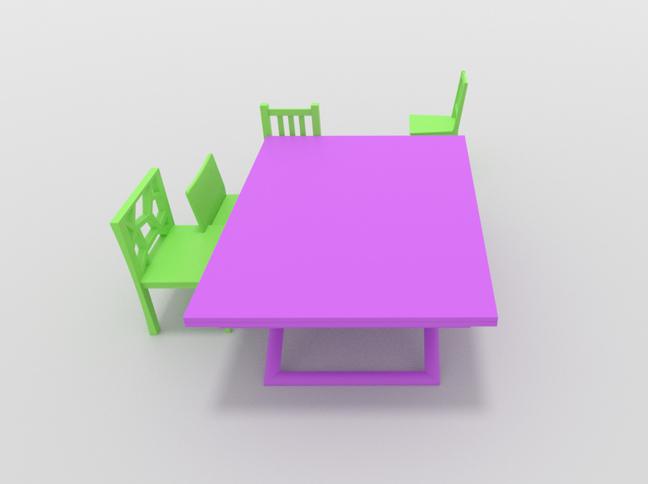}
		\includegraphics[width=0.325\textwidth]
		{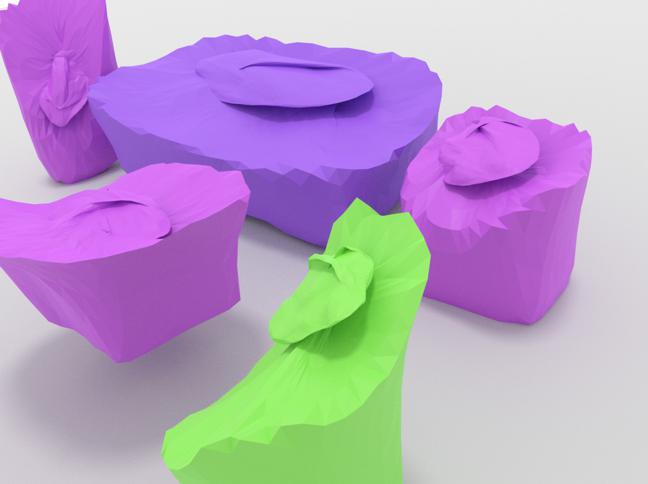}%
		\hfill
		\includegraphics[width=0.325\textwidth]
		{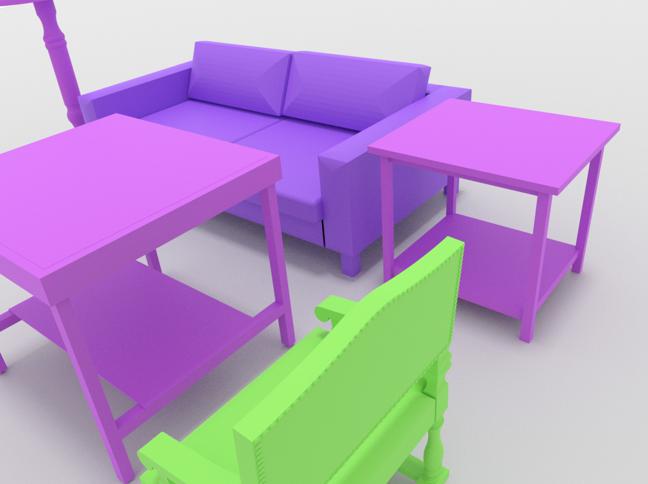}%
		\hfill
		\includegraphics[width=0.325\textwidth]
		{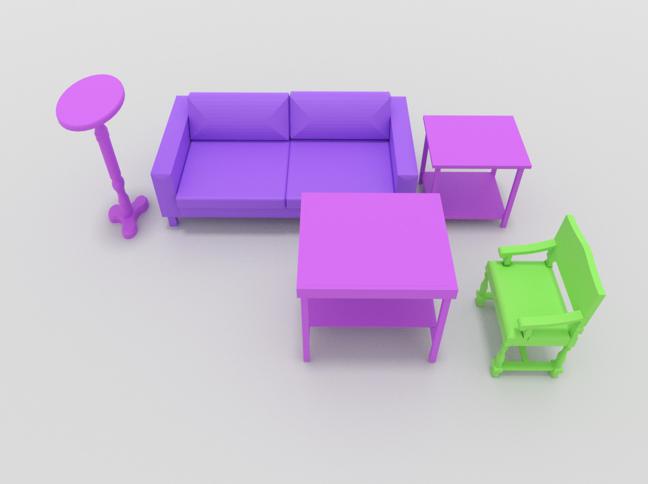}
		\includegraphics[width=0.325\textwidth]
		{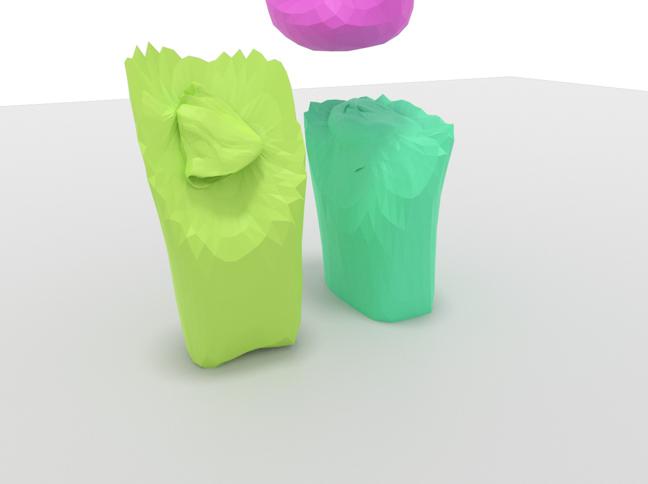}%
		\hfill
		\includegraphics[width=0.325\textwidth]
		{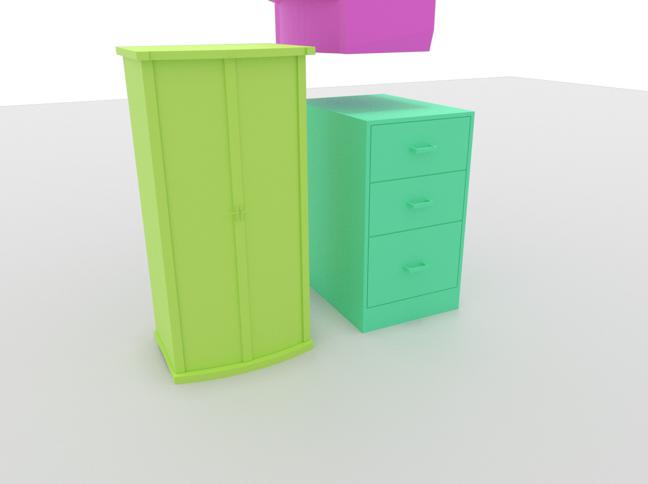}%
		\hfill
		\includegraphics[width=0.325\textwidth]
		{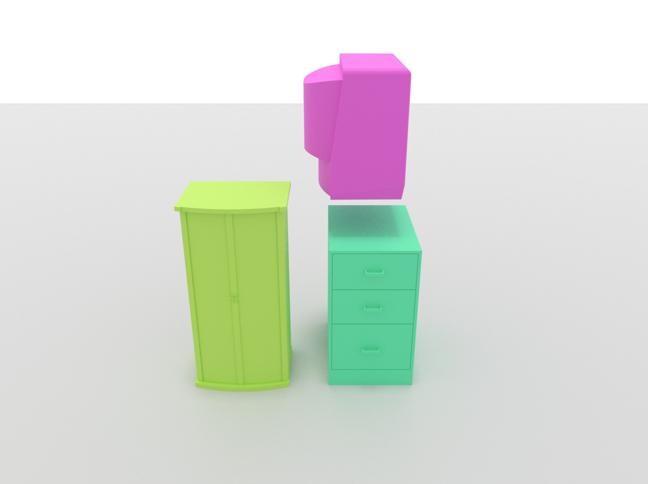}
		\includegraphics[width=0.325\textwidth]
		{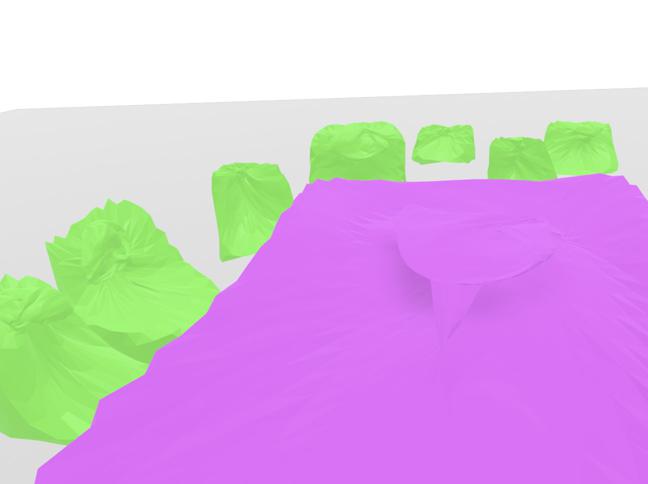}%
		\hfill
		\includegraphics[width=0.325\textwidth]
		{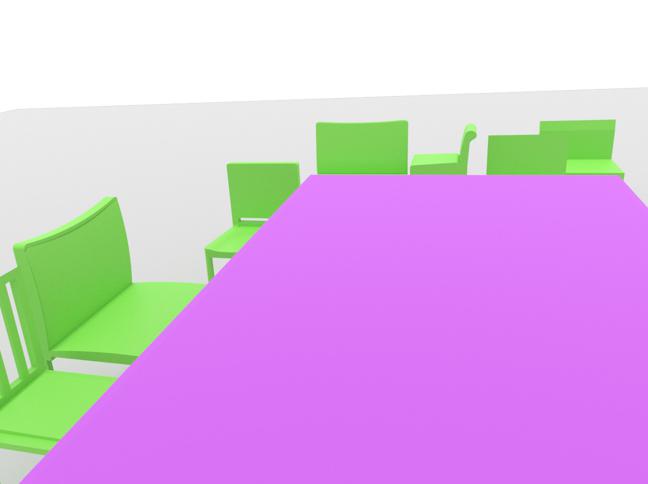}%
		\hfill
		\includegraphics[width=0.325\textwidth]
		{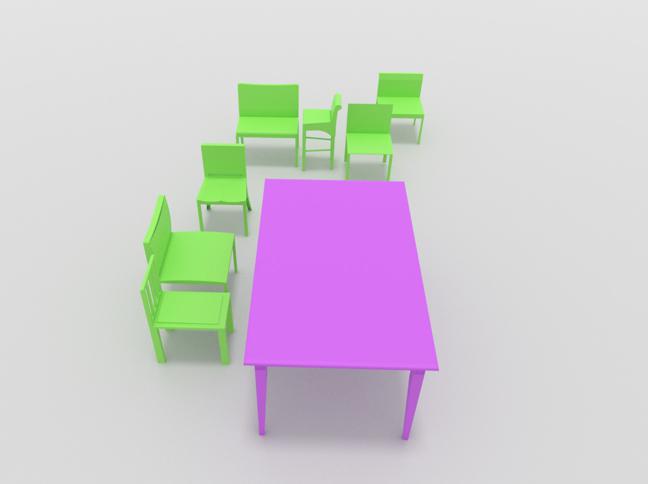}
		\includegraphics[width=0.325\textwidth]
		{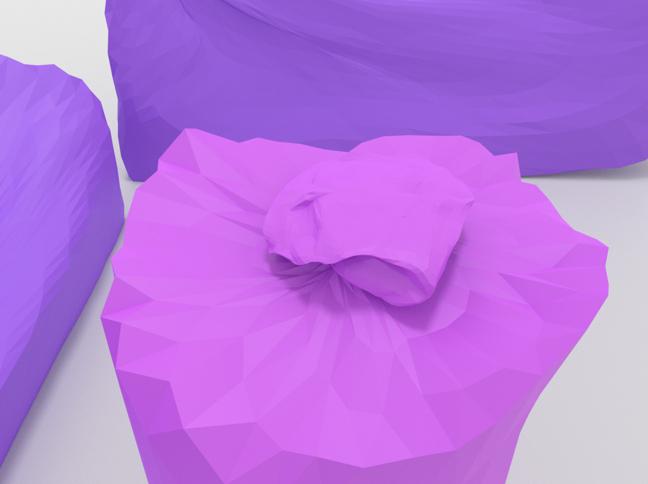}%
		\hfill
		\includegraphics[width=0.325\textwidth]
		{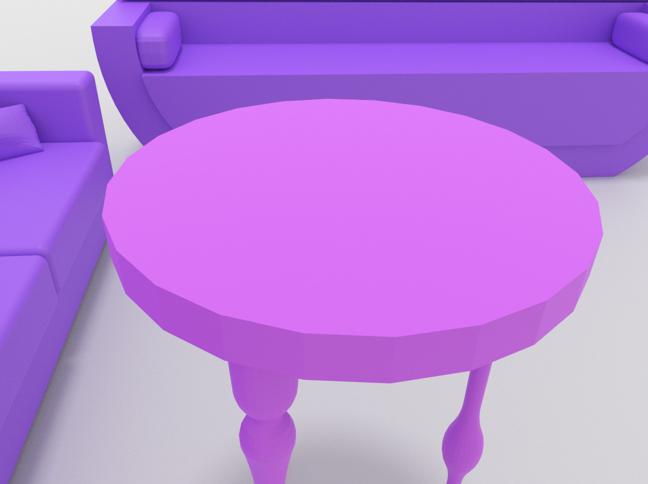}%
		\hfill
		\includegraphics[width=0.325\textwidth]
		{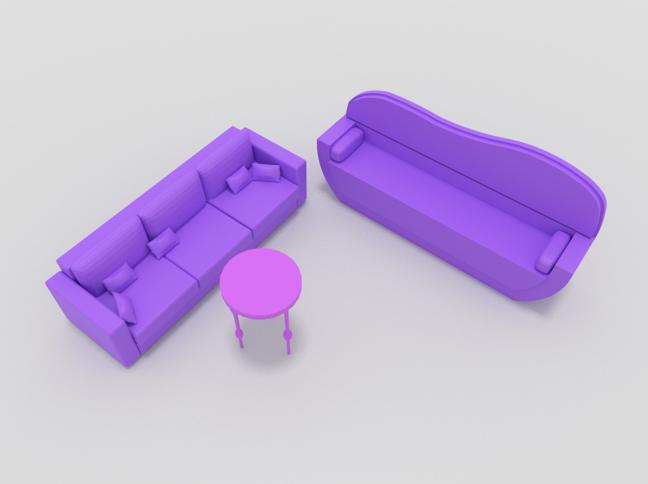}
		\caption{Ours}
	\end{subfigure}
	\caption{Qualitative comparisons with the state-of-the-art on single view reconstruction. Note that both Total3D~\cite{paschalidou2021atiss} and Im3D~\cite{yang2021scene} use 3D data for training while our method learns from only 2D supervisions. For each method we present the reconstruction result from both the image view and a global view. For our method, we also provide the reconstruction result without shape retrieval on the left.}
	\label{fig:qualitative_scene_recon}
	\vspace{-1.5em}
\end{figure*}

\subsection{Single-view Reconstruction}
\paragraph{Baselines} We evaluate our method on single-view instance reconstruction from real images, and compare with the prior arts which use 3D supervision: Total3D~\cite{nie2020total3dunderstanding} and Im3D~\cite{zhang2021holistic}. Both of them learn 3D scenes with a fully supervised manner, i.e., they require ground-truth object meshes, 2D detections, 3D posed object bounding boxes, and camera poses for training, while our method learns from multi-view 2D instance masks with camera poses.
We also compare with USL~\cite{gkioxari2022learning} which is an unsupervised method without using 3D supervision. Similar with ours, they require multi-view tracked instance masks for learning.
We train and test all methods using the official split on ScanNet.
For Total3D~\cite{nie2020total3dunderstanding} and Im3D~\cite{zhang2021holistic}, we use their pretrained model and finetune their 3D detector on ScanNet.
For finetuning, we uniformly sample up to 20 views to cover each scene, with each view contains at least three objects, which produces 11,395 valid views.
For USL~\cite{gkioxari2022learning}, we reproduce their network due to the unavailability of an official implementation, and train it on the same data with ours.
For fair comparison, we use ground-truth 2D object detections and camera poses as the input for training and testing all baselines.

\vspace{-1em}
\paragraph{Qualitative Comparisons}
Fig.~\ref{fig:qualitative_scene_recon} presents the qualitative comparisons with Total3D~\cite{nie2020total3dunderstanding} and Im3D~\cite{zhang2021holistic}, which are the state-of-the-art single-view scene reconstruction methods. Both Total3D~\cite{nie2020total3dunderstanding} and Im3D~\cite{zhang2021holistic} pretrain their shape generator on a 3D shape dataset~\cite{pix3d}. To produce better object geometry, our method keeps the shape retrieval in Sec.~\ref{sec:exp_scene_gen} while using the ShapeNet~\cite{chang2015shapenet} dataset to retrieve CAD models given object positions and meshes. We observe that Total3D presents better mesh quality than Im3D. However, since objects in ScanNet images are usually occluded due to the high scene complexity, it struggles to estimate 3D object poses from 2D detections and results in erroneous object placement (row 1,3,5,6). Im3D addresses this by involving a refinement stage to improve object layout. However, for those severely occluded or partially visible objects, it is vulnerable to object penetration issues (row 1,3,7). Instead of detecting 3D objects from 2D detections (as in Total3D and Im3D), our method focuses on the prior learning of object arrangements, i.e., to understand what a reasonable object layout looks like in multiple views, which helps to produce better scene configurations.

\begin{table}[!t]
	\centering
	\resizebox{1\columnwidth}{!}{
	\begin{tabular}{|l|c|c c|}
		\hline
		& Supervision & 3D Box IoU ($\uparrow$) & CD ($\downarrow$) \\
		\hline
		\hline
		Total3D~\cite{nie2020total3dunderstanding} & 3D & 0.15 & 1.36 \\
		Im3D~\cite{zhang2021holistic} & 3D & 0.23 & 0.84 \\
		\hline
		\hline
		USL*~\cite{gkioxari2022learning} & 2D & 0.10 & 1.73 \\
		Ours (w/o retrieval) & 2D & 0.25 & 1.01 \\
		Ours & 2D & \textbf{0.27} & \textbf{0.79} \\
		\hline
	\end{tabular}}
	\caption{Quantitative comparisons on single-view reconstruction. Note that USL* is reproduced from \cite{gkioxari2022learning} due to the unavailability of the official implementation.}
	\label{tab:quant_scene_recon}
	\vspace{-1em}
\end{table}

\vspace{-1em}
\paragraph{Quantitative Comparisons}
Tab.~\ref{tab:quant_scene_recon} presents the quantitative comparisons, where we observe that Total3D~\cite{nie2020total3dunderstanding} and USL~\cite{gkioxari2022learning} struggles to detect 3D object bounding boxes from 2D detections (lower 3D Box IoU) because both of them predict object centers by depth estimation, which is still a challenging problem due to the ambiguity.
Our method attempts to reconstruct a plausible scene from the learned scene prior conditioned on the input instance masks, which presents more reasonable 3D object bounding boxes. Since training and testing our method requires instance masks for supervision, the mask quality has a direct influence to our reconstruction results. However, mask artifacts are inevitable in real images (row 1-5 in Fig.~\ref{fig:qualitative_scene_recon}~(a)). By learning shapes from multi-view observations with a naive shape retrieval post-processing, our method presents convincing object layouts and geometries.

\subsection{Discussion}
\begin{figure}[!t]
	\centering
	\begin{subfigure}[t]{0.09\textwidth}
		\includegraphics[width=\textwidth]
		{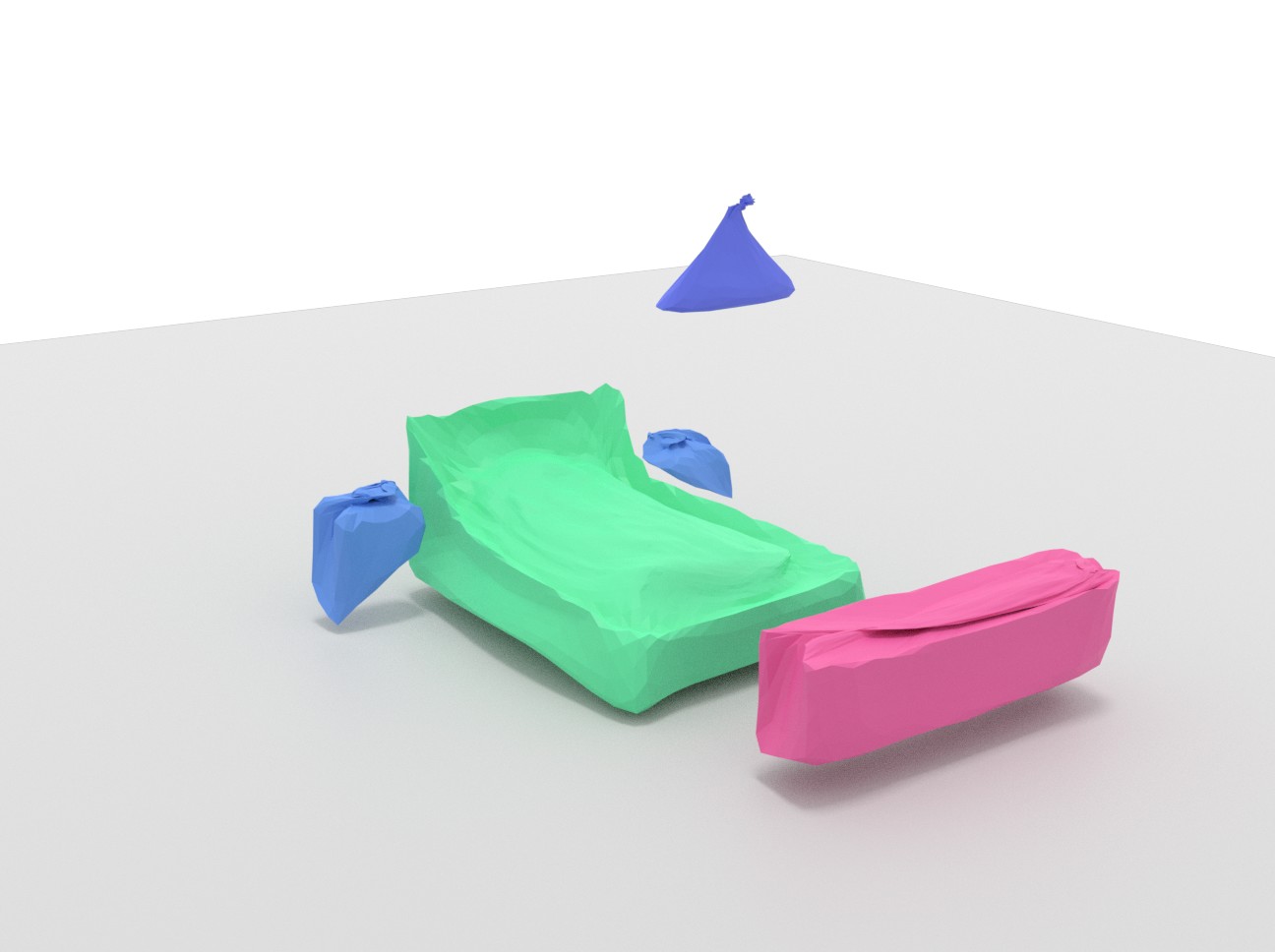}
		\includegraphics[width=\textwidth]
		{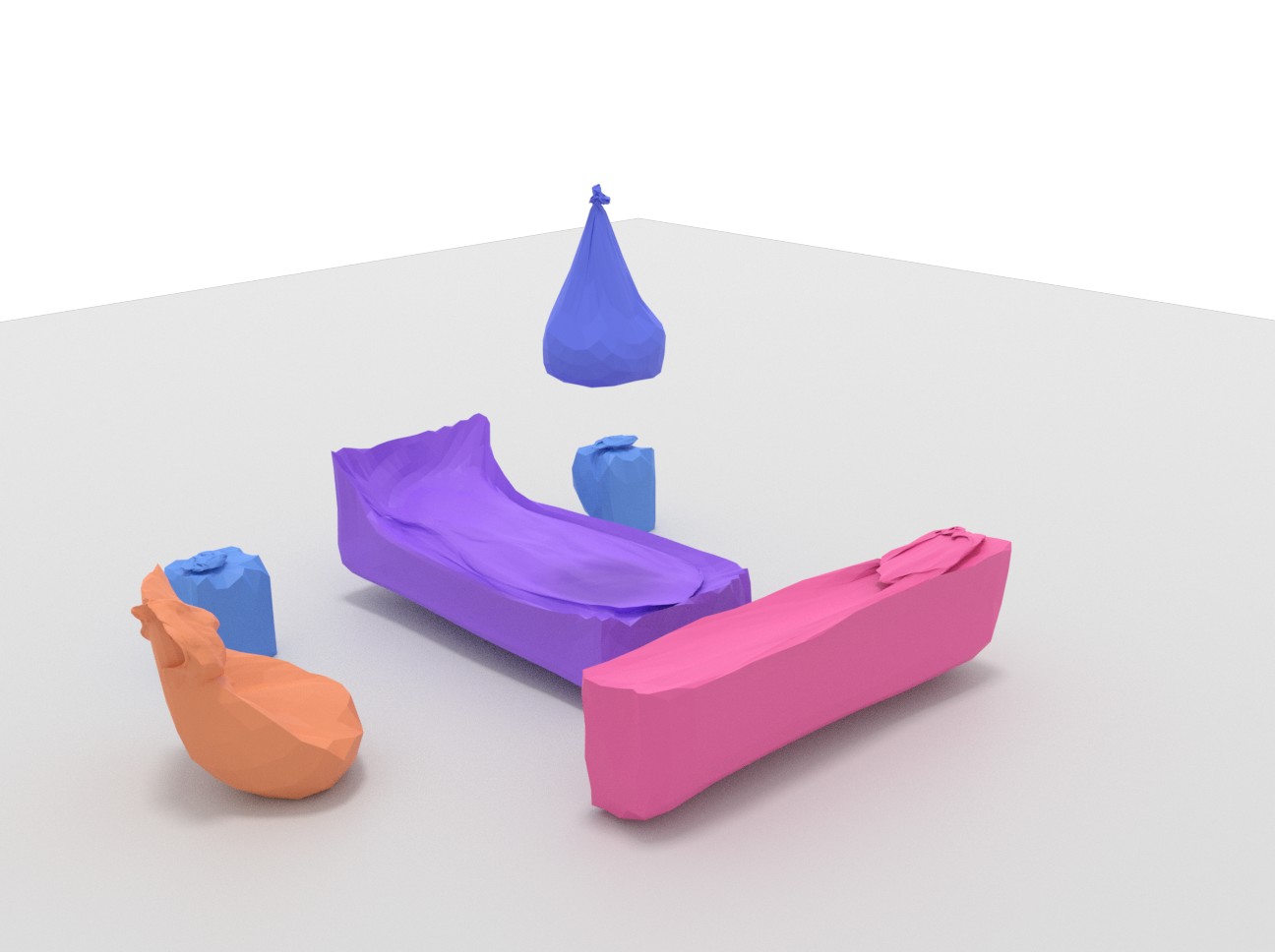}
	\end{subfigure}
	\begin{subfigure}[t]{0.282\textwidth}
		\includegraphics[width=0.32\textwidth]
		{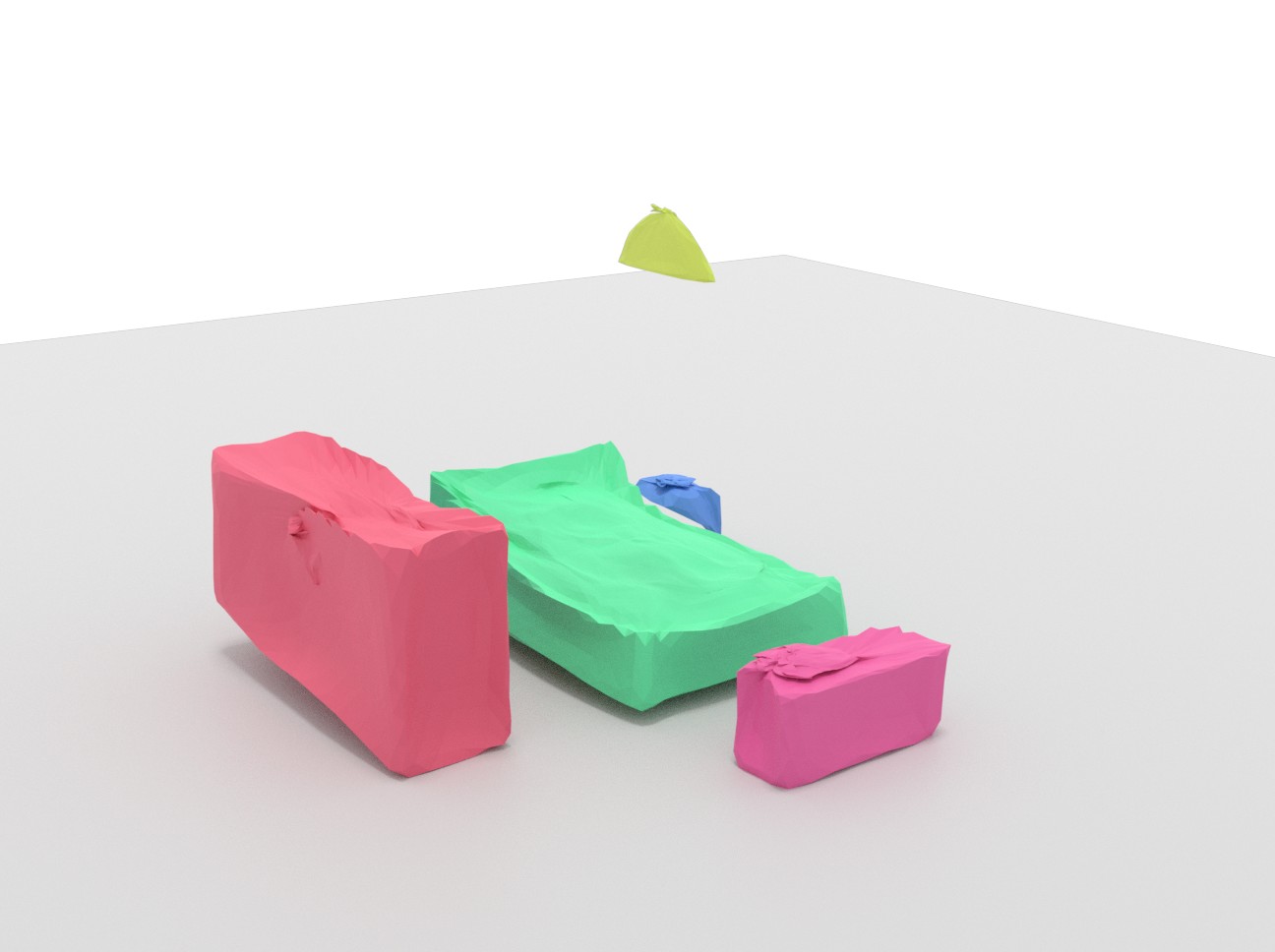}%
		\hfill
		\includegraphics[width=0.32\textwidth]
		{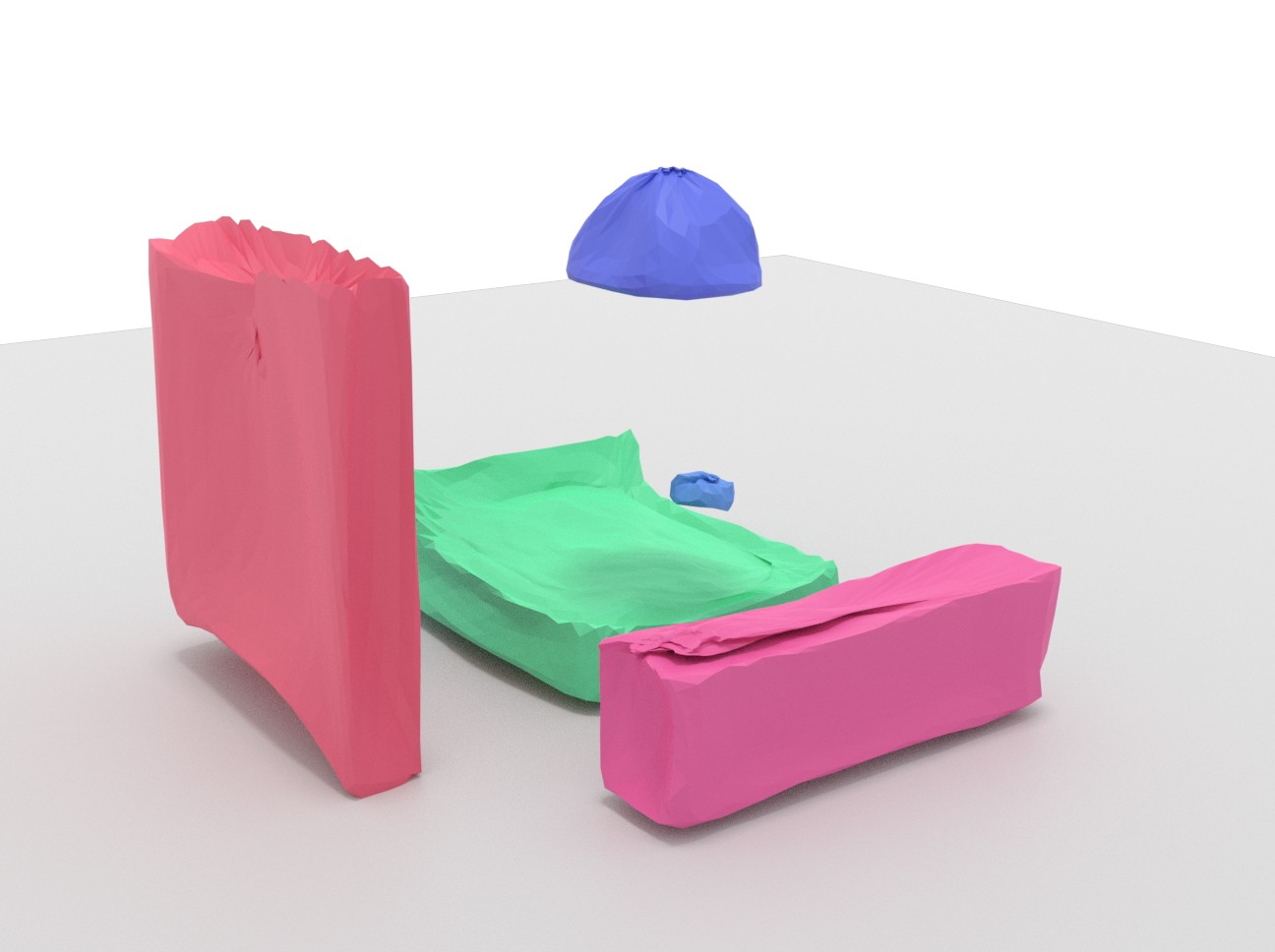}%
		\hfill
		\includegraphics[width=0.32\textwidth]
		{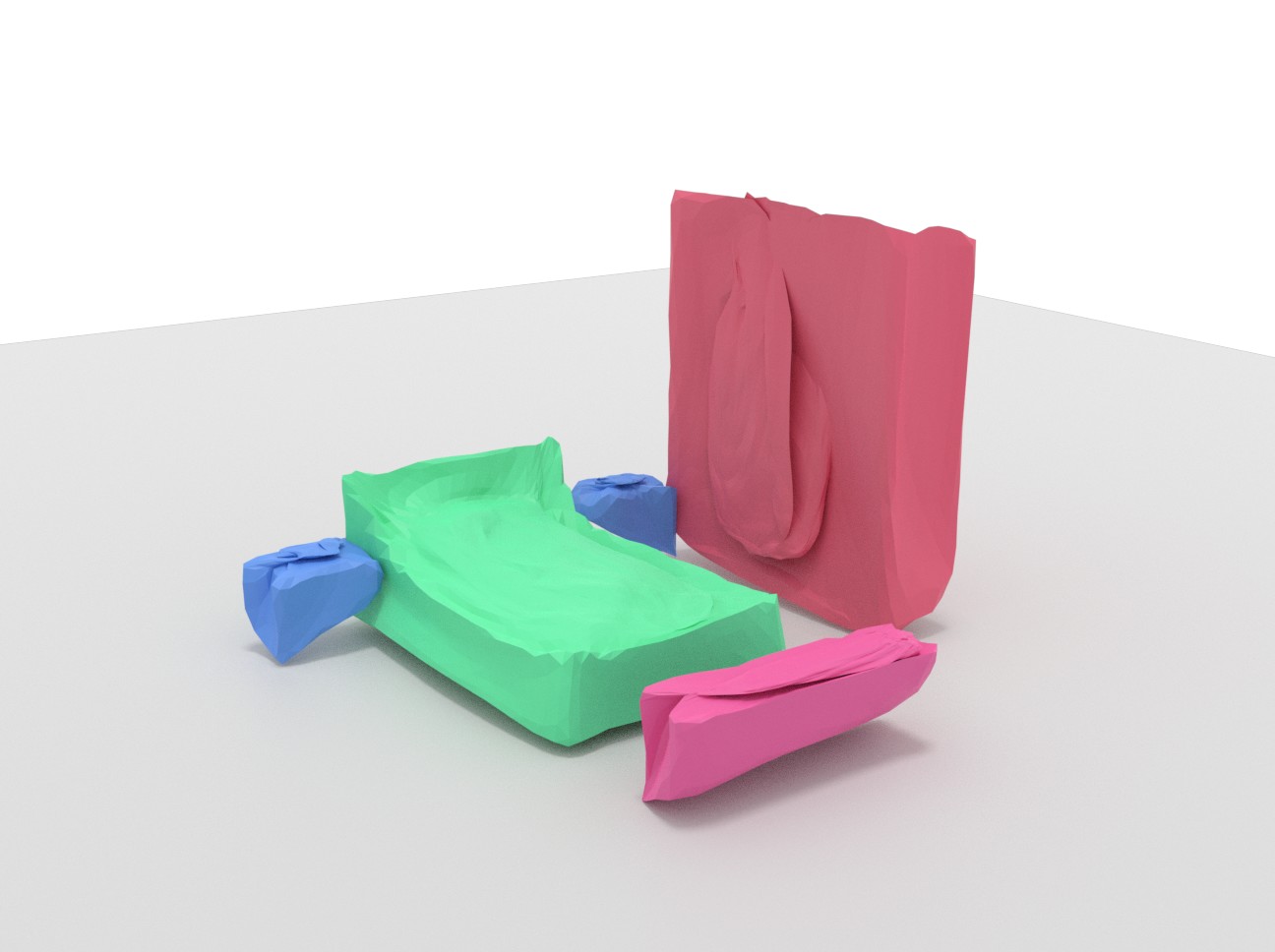}
		\includegraphics[width=0.32\textwidth]
		{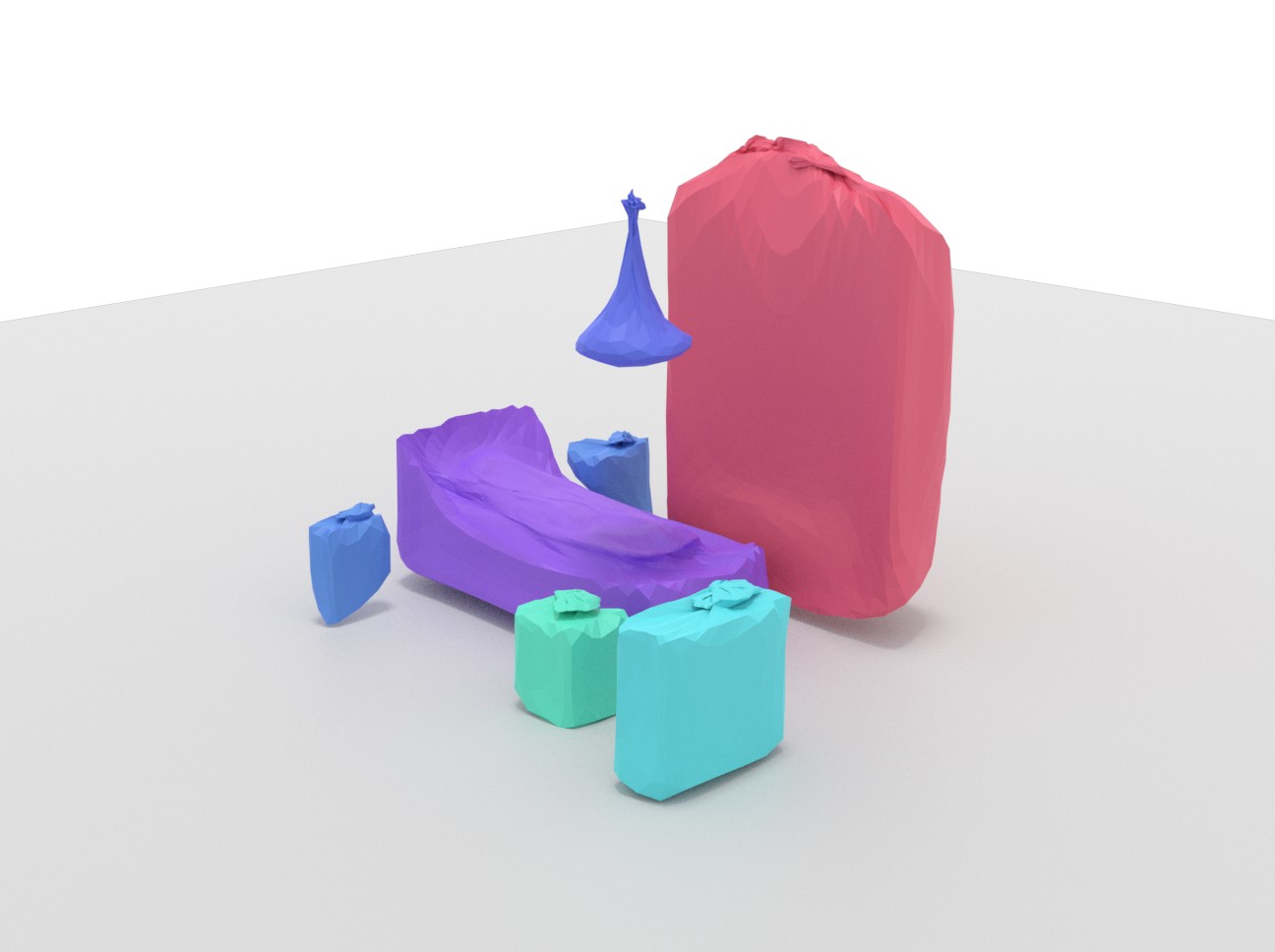}%
		\hfill
		\includegraphics[width=0.32\textwidth]
		{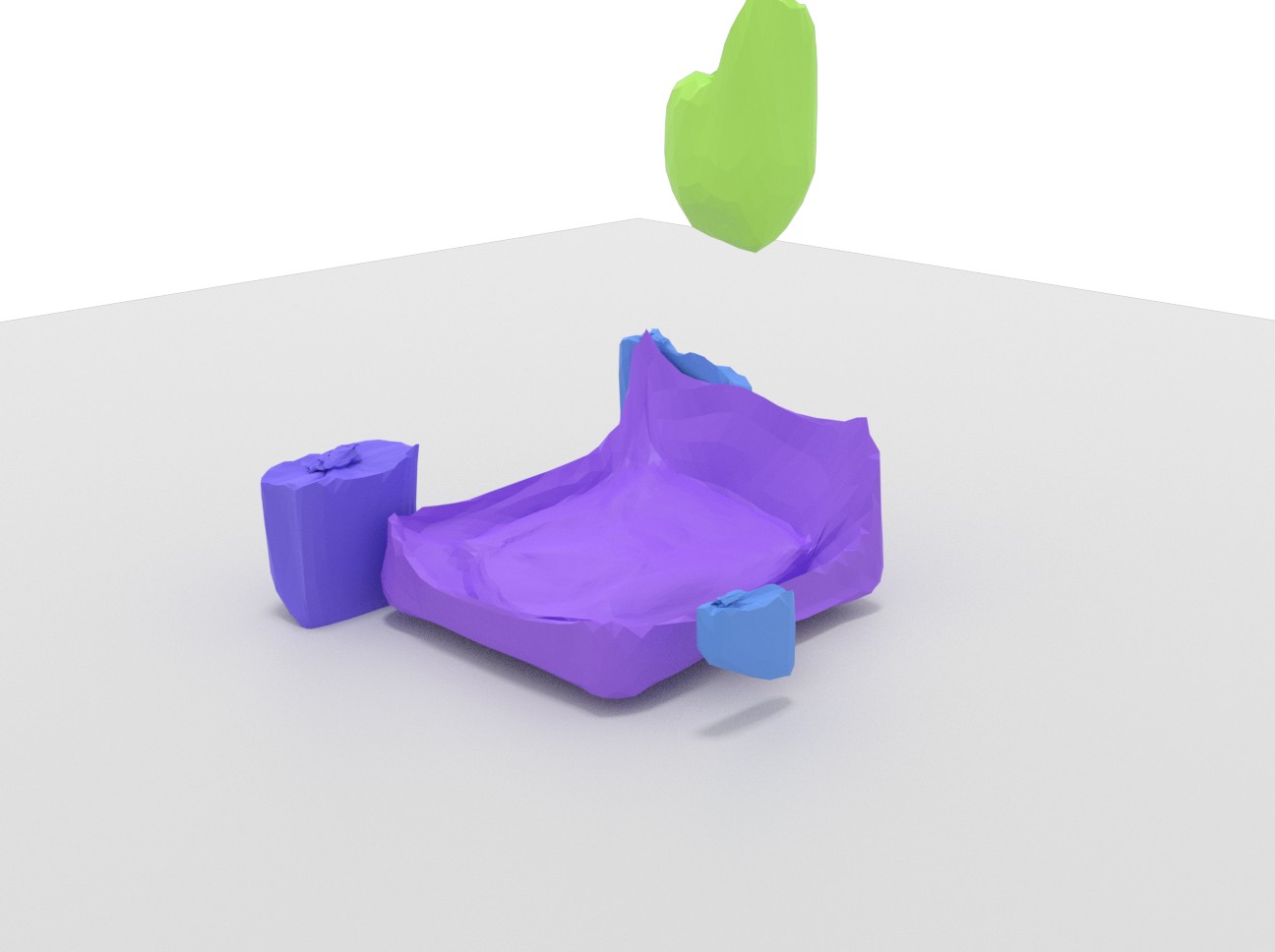}%
		\hfill
		\includegraphics[width=0.32\textwidth]
		{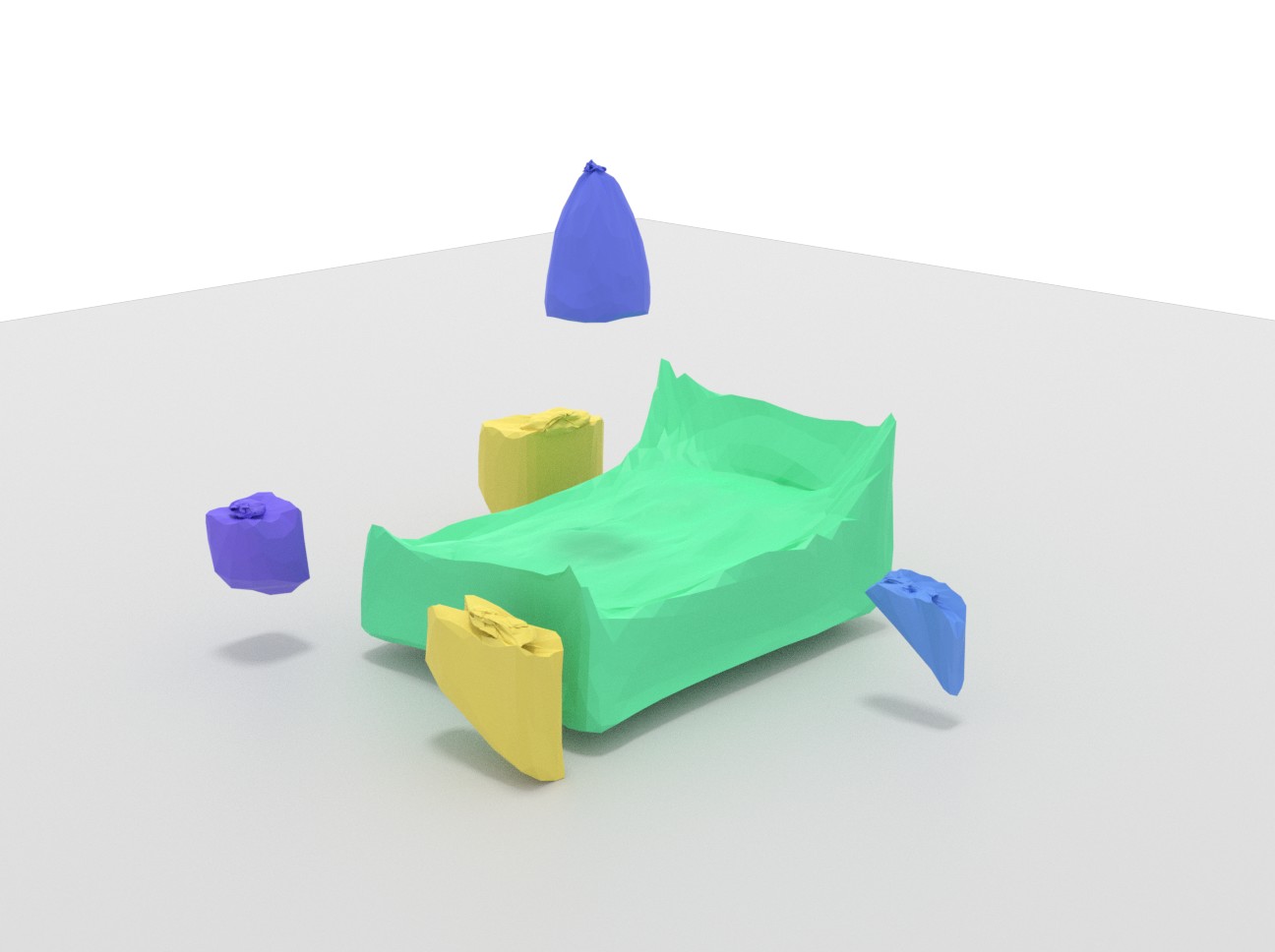}
	\end{subfigure}
	\begin{subfigure}[t]{0.09\textwidth}
		\includegraphics[width=\textwidth]
		{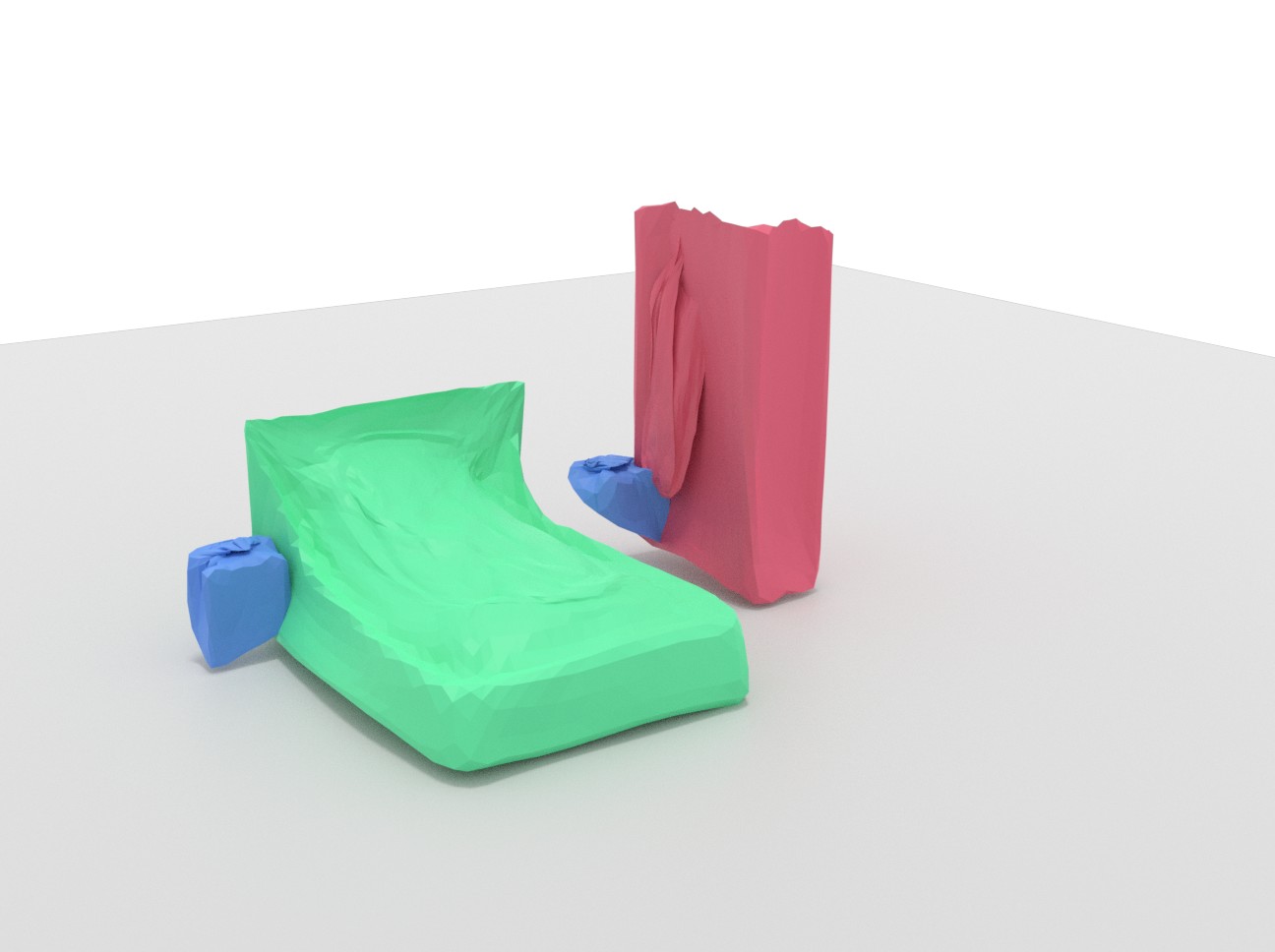}
		\includegraphics[width=\textwidth]
		{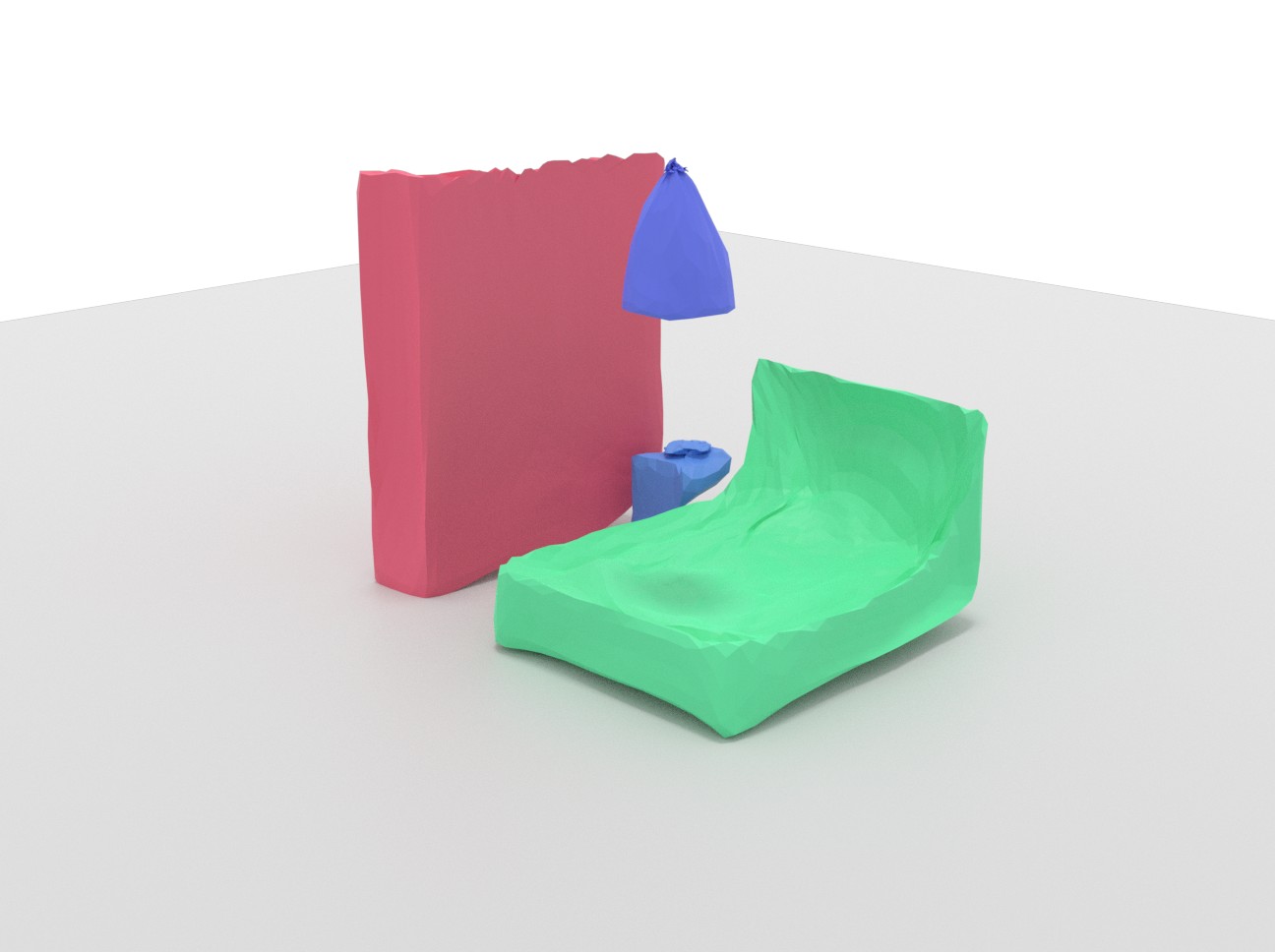}
	\end{subfigure}
	\caption{Scene interpolation between two samples (the leftmost and the rightmost).}
	\label{fig:qualitative_scene_interp}
	\vspace{-1em}
\end{figure}

\paragraph{Scene Interpolation} Fig.~\ref{fig:qualitative_scene_interp} presents the interpolation results between two scenes. Given the latent vectors of two scenes, we interpolate intermediate vectors by sampling from the geodesic curve between them on the hypersphere surface (see Fig.~\ref{fig:intro_figure}). By decoding them, we can obtain the interpolated scenes. The transition between intermediates results show that our method interpolate not only the object pose and geometry but also the semantics and scene context, where the interpolated scenes are constrained with a reasonable scene layout.

\vspace{-1em}
\paragraph{Ablations} In Tab.~\ref{tab:ablation} we investigate the effects of each submodule in scene synthesis, i.e., hyper-spherical latent space, permutation-invariant transformer, and stage-wise layout and shape learning, by replacing them with uniform latent space parameterization (\textbf{c}\textsubscript{1}), LSTM (\textbf{c}\textsubscript{2}), and one-stage layout and shape training (\textbf{c}\textsubscript{3}), respectively. From the results, we observe that our transformer plays the most important role. LSTM encodes object order information. However, objects in a scene is an unordered set, and our transformer learns the permutation-invariant context to predict the next object recursively, which largely improves our performance. The hypersphere parameterization and stage-wise training further increases the output plausibility and stabilizes the learning process, and combining them delivers the best performance.

\vspace{-1em}
\paragraph{Limitations} Since training our network requires 2D posed images with instance masks, erroneous camera poses or mask labels would mislead the learning process. We believe an interesting direction for future work is to incorporate an uncertainty model to learn stable scene priors from noisy masks and camera poses.

\begin{table}[!t]
	\centering
	\resizebox{0.9\columnwidth}{!}{
	\begin{tabular}{|l|c c c|c c c|}
		\hline
		&  \multicolumn{3}{c|}{Bedroom} & \multicolumn{3}{c|}{Living room}\\
		& FID ($\downarrow$) & SCA & KL ($\downarrow$) & FID ($\downarrow$) & SCA & KL ($\downarrow$)  \\
		\hline
		\hline
		\textbf{c}\textsubscript{1} & 30.61
		& 0.96 & 0.24 & 95.51 & 1.00 & 0.38\\
		\textbf{c}\textsubscript{2} & 67.97 & 1.00 & 0.30 & 86.77 & 1.00 & 0.26 \\
		\textbf{c}\textsubscript{3} & 24.24 & 0.90 & 0.04 & 43.43 & 0.98 & 0.03 \\
		Ours & \textbf{21.59} & \textbf{0.85} & \textbf{0.03} & \textbf{40.47} & \textbf{0.96} & \textbf{0.02} \\
		\hline
	\end{tabular}}
	\caption{Ablation analysis on different configurations.}
	\label{tab:ablation}
	\vspace{-1em}
\end{table}

\section{Conclusion}
We have presented a novel approach to learn the prior distribution of object layouts and shapes in 3D scenes with only 2D supervision. Our proposed method represents a 3D scene as a latent vector. By decoding it autoregressively using a permutation-invariant transformer, we recover a sequence of object instances characterized by their class categories, 3D bounding boxes and meshes. Since collecting 3D data at scale is expensive and intractable, we design a 2D view loss to train object generation using only 2D supervision. We demonstrate that such prior learning approach enables various downstream tasks, i.e., scene synthesis, interpolation and single-view reconstruction, and produces convincing scene arrangements. We hope that this establishes a step towards a universal self-supervised scene estimator that mimics human vision to understand, emulate and create 3D surroundings from only image observations.

\vspace{-1em}
\paragraph{Acknowledgments}
This work was supported by the ERC Starting Grant Scan2CAD (804724), the German Research Foundation (DFG) Grant ``Making Machine Learning on Static and Dynamic 3D Data Practical'', the German Research Foundation (DFG) Research Unit ``Learning and Simulation in Visual Computing'', NSFC-62172348, and the Bavarian State Ministry of Science and the Arts coordinated by the Bavarian Research Institute for Digital Transformation (bidt).

\clearpage
{\small
	\bibliographystyle{ieee_fullname}
	\bibliography{egbib}

\begin{thebibliography}{10}\itemsep=-1pt

\bibitem{fasttransformers}
Fast transformers.
\newblock \url{https://fast-transformers.github.io/}.
\newblock Accessed: 2022-11-12.

\bibitem{lstm_tutorial}
Sequence models and long short-term memory networks.
\newblock
  \url{https://pytorch.org/tutorials/beginner/nlp/sequence_models_tutorial.html}.
\newblock Accessed: 2022-11-14.

\bibitem{avetisyan2019scan2cad}
Armen Avetisyan, Manuel Dahnert, Angela Dai, Manolis Savva, Angel~X Chang, and
  Matthias Nie{\ss}ner.
\newblock Scan2cad: Learning cad model alignment in rgb-d scans.
\newblock In {\em Proceedings of the IEEE/CVF Conference on computer vision and
  pattern recognition}, pages 2614--2623, 2019.

\bibitem{carion2020end}
Nicolas Carion, Francisco Massa, Gabriel Synnaeve, Nicolas Usunier, Alexander
  Kirillov, and Sergey Zagoruyko.
\newblock End-to-end object detection with transformers.
\newblock In {\em European conference on computer vision}, pages 213--229.
  Springer, 2020.

\bibitem{chang2014learning}
Angel Chang, Manolis Savva, and Christopher~D Manning.
\newblock Learning spatial knowledge for text to 3d scene generation.
\newblock In {\em Proceedings of the 2014 conference on empirical methods in
  natural language processing (EMNLP)}, pages 2028--2038, 2014.

\bibitem{chang2017sceneseer}
Angel~X Chang, Mihail Eric, Manolis Savva, and Christopher~D Manning.
\newblock Sceneseer: 3d scene design with natural language.
\newblock {\em arXiv preprint arXiv:1703.00050}, 2017.

\bibitem{chang2015shapenet}
Angel~X Chang, Thomas Funkhouser, Leonidas Guibas, Pat Hanrahan, Qixing Huang,
  Zimo Li, Silvio Savarese, Manolis Savva, Shuran Song, Hao Su, et~al.
\newblock Shapenet: An information-rich 3d model repository.
\newblock {\em arXiv preprint arXiv:1512.03012}, 2015.

\bibitem{dahnert2021panoptic}
Manuel Dahnert, Ji Hou, Matthias Nie{\ss}ner, and Angela Dai.
\newblock Panoptic 3d scene reconstruction from a single rgb image.
\newblock {\em Advances in Neural Information Processing Systems},
  34:8282--8293, 2021.

\bibitem{dai2017scannet}
Angela Dai, Angel~X Chang, Manolis Savva, Maciej Halber, Thomas Funkhouser, and
  Matthias Nie{\ss}ner.
\newblock Scannet: Richly-annotated 3d reconstructions of indoor scenes.
\newblock In {\em Proceedings of the IEEE conference on computer vision and
  pattern recognition}, pages 5828--5839, 2017.

\bibitem{davydov2022adversarial}
Andrey Davydov, Anastasia Remizova, Victor Constantin, Sina Honari, Mathieu
  Salzmann, and Pascal Fua.
\newblock Adversarial parametric pose prior.
\newblock In {\em Proceedings of the IEEE/CVF Conference on Computer Vision and
  Pattern Recognition}, pages 10997--11005, 2022.

\bibitem{denninger2019blenderproc}
Maximilian Denninger, Martin Sundermeyer, Dominik Winkelbauer, Youssef Zidan,
  Dmitry Olefir, Mohamad Elbadrawy, Ahsan Lodhi, and Harinandan Katam.
\newblock Blenderproc.
\newblock {\em arXiv preprint arXiv:1911.01911}, 2019.

\bibitem{fisher2010context}
Matthew Fisher and Pat Hanrahan.
\newblock Context-based search for 3d models.
\newblock In {\em ACM SIGGRAPH Asia 2010 papers}, pages 1--10. 2010.

\bibitem{fisher2012example}
Matthew Fisher, Daniel Ritchie, Manolis Savva, Thomas Funkhouser, and Pat
  Hanrahan.
\newblock Example-based synthesis of 3d object arrangements.
\newblock {\em ACM Transactions on Graphics (TOG)}, 31(6):1--11, 2012.

\bibitem{fisher2015activity}
Matthew Fisher, Manolis Savva, Yangyan Li, Pat Hanrahan, and Matthias
  Nie{\ss}ner.
\newblock Activity-centric scene synthesis for functional 3d scene modeling.
\newblock {\em ACM Transactions on Graphics (TOG)}, 34(6):1--13, 2015.

\bibitem{fu20213d}
Huan Fu, Bowen Cai, Lin Gao, Ling-Xiao Zhang, Jiaming Wang, Cao Li, Qixun Zeng,
  Chengyue Sun, Rongfei Jia, Binqiang Zhao, et~al.
\newblock 3d-front: 3d furnished rooms with layouts and semantics.
\newblock In {\em Proceedings of the IEEE/CVF International Conference on
  Computer Vision}, pages 10933--10942, 2021.

\bibitem{fu2017adaptive}
Qiang Fu, Xiaowu Chen, Xiaotian Wang, Sijia Wen, Bin Zhou, and Hongbo Fu.
\newblock Adaptive synthesis of indoor scenes via activity-associated object
  relation graphs.
\newblock {\em ACM Transactions on Graphics (TOG)}, 36(6):1--13, 2017.

\bibitem{Gkioxari_2019_ICCV}
Georgia Gkioxari, Jitendra Malik, and Justin Johnson.
\newblock Mesh r-cnn.
\newblock In {\em Proceedings of the IEEE/CVF International Conference on
  Computer Vision (ICCV)}, October 2019.

\bibitem{gkioxari2022learning}
Georgia Gkioxari, Nikhila Ravi, and Justin Johnson.
\newblock Learning 3d object shape and layout without 3d supervision.
\newblock In {\em Proceedings of the IEEE/CVF Conference on Computer Vision and
  Pattern Recognition}, pages 1695--1704, 2022.

\bibitem{groueix2018papier}
Thibault Groueix, Matthew Fisher, Vladimir~G Kim, Bryan~C Russell, and Mathieu
  Aubry.
\newblock A papier-m{\^a}ch{\'e} approach to learning 3d surface generation.
\newblock In {\em Proceedings of the IEEE conference on computer vision and
  pattern recognition}, pages 216--224, 2018.

\bibitem{gumeli2022roca}
Can G{\"u}meli, Angela Dai, and Matthias Nie{\ss}ner.
\newblock Roca: Robust cad model retrieval and alignment from a single image.
\newblock In {\em Proceedings of the IEEE/CVF Conference on Computer Vision and
  Pattern Recognition}, pages 4022--4031, 2022.

\bibitem{hou2020revealnet}
Ji Hou, Angela Dai, and Matthias Nie{\ss}ner.
\newblock Revealnet: Seeing behind objects in rgb-d scans.
\newblock In {\em Proceedings of the IEEE/CVF Conference on Computer Vision and
  Pattern Recognition}, pages 2098--2107, 2020.

\bibitem{huang2018holistic}
Siyuan Huang, Siyuan Qi, Yixin Zhu, Yinxue Xiao, Yuanlu Xu, and Song-Chun Zhu.
\newblock Holistic 3d scene parsing and reconstruction from a single rgb image.
\newblock In {\em Proceedings of the European conference on computer vision
  (ECCV)}, pages 187--203, 2018.

\bibitem{izadinia2017im2cad}
Hamid Izadinia, Qi Shan, and Steven~M Seitz.
\newblock Im2cad.
\newblock In {\em Proceedings of the IEEE conference on computer vision and
  pattern recognition}, pages 5134--5143, 2017.

\bibitem{jiang2012learning}
Yun Jiang, Marcus Lim, and Ashutosh Saxena.
\newblock Learning object arrangements in 3d scenes using human context.
\newblock {\em arXiv preprint arXiv:1206.6462}, 2012.

\bibitem{katharopoulos_et_al_2020}
A. Katharopoulos, A. Vyas, N. Pappas, and F. Fleuret.
\newblock Transformers are rnns: Fast autoregressive transformers with linear
  attention.
\newblock In {\em Proceedings of the International Conference on Machine
  Learning (ICML)}, 2020.

\bibitem{keshavarzi2020scenegen}
Mohammad Keshavarzi, Aakash Parikh, Xiyu Zhai, Melody Mao, Luisa Caldas, and
  Allen~Y Yang.
\newblock Scenegen: Generative contextual scene augmentation using scene graph
  priors.
\newblock {\em arXiv preprint arXiv:2009.12395}, 2020.

\bibitem{kundu20183d}
Abhijit Kundu, Yin Li, and James~M Rehg.
\newblock 3d-rcnn: Instance-level 3d object reconstruction via
  render-and-compare.
\newblock In {\em Proceedings of the IEEE conference on computer vision and
  pattern recognition}, pages 3559--3568, 2018.

\bibitem{kuo2020mask2cad}
Weicheng Kuo, Anelia Angelova, Tsung-Yi Lin, and Angela Dai.
\newblock Mask2cad: 3d shape prediction by learning to segment and retrieve.
\newblock In {\em European Conference on Computer Vision}, pages 260--277.
  Springer, 2020.

\bibitem{kuo2021patch2cad}
Weicheng Kuo, Anelia Angelova, Tsung-Yi Lin, and Angela Dai.
\newblock Patch2cad: Patchwise embedding learning for in-the-wild shape
  retrieval from a single image.
\newblock In {\em Proceedings of the IEEE/CVF International Conference on
  Computer Vision}, pages 12589--12599, 2021.

\bibitem{leimer2022atek}
Kurt Leimer, Paul Guerrero, Tomer Weiss, and Przemyslaw Musialski.
\newblock Atek: Augmenting transformers with expert knowledge for indoor layout
  synthesis.
\newblock {\em arXiv preprint arXiv:2202.00185}, 2022.

\bibitem{li2019grains}
Manyi Li, Akshay~Gadi Patil, Kai Xu, Siddhartha Chaudhuri, Owais Khan, Ariel
  Shamir, Changhe Tu, Baoquan Chen, Daniel Cohen-Or, and Hao Zhang.
\newblock Grains: Generative recursive autoencoders for indoor scenes.
\newblock {\em ACM Transactions on Graphics (TOG)}, 38(2):1--16, 2019.

\bibitem{liu2022towards}
Haolin Liu, Yujian Zheng, Guanying Chen, Shuguang Cui, and Xiaoguang Han.
\newblock Towards high-fidelity single-view holistic reconstruction of indoor
  scenes.
\newblock In {\em European Conference on Computer Vision}, pages 429--446.
  Springer, 2022.

\bibitem{ma2016action}
Rui Ma, Honghua Li, Changqing Zou, Zicheng Liao, Xin Tong, and Hao Zhang.
\newblock Action-driven 3d indoor scene evolution.
\newblock {\em ACM Trans. Graph.}, 35(6):173--1, 2016.

\bibitem{merrell2011interactive}
Paul Merrell, Eric Schkufza, Zeyang Li, Maneesh Agrawala, and Vladlen Koltun.
\newblock Interactive furniture layout using interior design guidelines.
\newblock {\em ACM transactions on graphics (TOG)}, 30(4):1--10, 2011.

\bibitem{nie2022pose2room}
Yinyu Nie, Angela Dai, Xiaoguang Han, and Matthias Nie{\ss}ner.
\newblock Pose2room: Understanding 3d scenes from human activities.
\newblock In {\em European Conference on Computer Vision}, pages 425--443.
  Springer, 2022.

\bibitem{nie2020total3dunderstanding}
Yinyu Nie, Xiaoguang Han, Shihui Guo, Yujian Zheng, Jian Chang, and Jian~Jun
  Zhang.
\newblock Total3dunderstanding: Joint layout, object pose and mesh
  reconstruction for indoor scenes from a single image.
\newblock In {\em Proceedings of the IEEE/CVF Conference on Computer Vision and
  Pattern Recognition}, pages 55--64, 2020.

\bibitem{Nie_2021_CVPR}
Yinyu Nie, Ji Hou, Xiaoguang Han, and Matthias Niessner.
\newblock Rfd-net: Point scene understanding by semantic instance
  reconstruction.
\newblock In {\em Proceedings of the IEEE/CVF Conference on Computer Vision and
  Pattern Recognition (CVPR)}, pages 4608--4618, June 2021.

\bibitem{Paschalidou2021NEURIPS}
Despoina Paschalidou, Amlan Kar, Maria Shugrina, Karsten Kreis, Andreas Geiger,
  and Sanja Fidler.
\newblock Atiss: Autoregressive transformers for indoor scene synthesis.
\newblock In {\em Advances in Neural Information Processing Systems (NeurIPS)},
  2021.

\bibitem{paschalidou2021atiss}
Despoina Paschalidou, Amlan Kar, Maria Shugrina, Karsten Kreis, Andreas Geiger,
  and Sanja Fidler.
\newblock Atiss: Autoregressive transformers for indoor scene synthesis.
\newblock {\em Advances in Neural Information Processing Systems},
  34:12013--12026, 2021.

\bibitem{popov2020corenet}
Stefan Popov, Pablo Bauszat, and Vittorio Ferrari.
\newblock Corenet: Coherent 3d scene reconstruction from a single rgb image.
\newblock In {\em European Conference on Computer Vision}, pages 366--383.
  Springer, 2020.

\bibitem{purkait2020sg}
Pulak Purkait, Christopher Zach, and Ian Reid.
\newblock Sg-vae: Scene grammar variational autoencoder to generate new indoor
  scenes.
\newblock In {\em European Conference on Computer Vision}, pages 155--171.
  Springer, 2020.

\bibitem{qi2018human}
Siyuan Qi, Yixin Zhu, Siyuan Huang, Chenfanfu Jiang, and Song-Chun Zhu.
\newblock Human-centric indoor scene synthesis using stochastic grammar.
\newblock In {\em Proceedings of the IEEE Conference on Computer Vision and
  Pattern Recognition}, pages 5899--5908, 2018.

\bibitem{ravi2020accelerating}
Nikhila Ravi, Jeremy Reizenstein, David Novotny, Taylor Gordon, Wan-Yen Lo,
  Justin Johnson, and Georgia Gkioxari.
\newblock Accelerating 3d deep learning with pytorch3d.
\newblock {\em arXiv preprint arXiv:2007.08501}, 2020.

\bibitem{ritchie2019fast}
Daniel Ritchie, Kai Wang, and Yu-an Lin.
\newblock Fast and flexible indoor scene synthesis via deep convolutional
  generative models.
\newblock In {\em Proceedings of the IEEE/CVF Conference on Computer Vision and
  Pattern Recognition}, pages 6182--6190, 2019.

\bibitem{savva2017scenesuggest}
Manolis Savva, Angel~X Chang, and Maneesh Agrawala.
\newblock Scenesuggest: Context-driven 3d scene design.
\newblock {\em arXiv preprint arXiv:1703.00061}, 2017.

\bibitem{stewart2016end}
Russell Stewart, Mykhaylo Andriluka, and Andrew~Y Ng.
\newblock End-to-end people detection in crowded scenes.
\newblock In {\em Proceedings of the IEEE conference on computer vision and
  pattern recognition}, pages 2325--2333, 2016.

\bibitem{pix3d}
Xingyuan Sun, Jiajun Wu, Xiuming Zhang, Zhoutong Zhang, Chengkai Zhang, Tianfan
  Xue, Joshua~B Tenenbaum, and William~T Freeman.
\newblock Pix3d: Dataset and methods for single-image 3d shape modeling.
\newblock In {\em IEEE Conference on Computer Vision and Pattern Recognition
  (CVPR)}, 2018.

\bibitem{tulsiani2018factoring}
Shubham Tulsiani, Saurabh Gupta, David~F Fouhey, Alexei~A Efros, and Jitendra
  Malik.
\newblock Factoring shape, pose, and layout from the 2d image of a 3d scene.
\newblock In {\em Proceedings of the IEEE Conference on Computer Vision and
  Pattern Recognition}, pages 302--310, 2018.

\bibitem{vaswani2017attention}
Ashish Vaswani, Noam Shazeer, Niki Parmar, Jakob Uszkoreit, Llion Jones,
  Aidan~N Gomez, {\L}ukasz Kaiser, and Illia Polosukhin.
\newblock Attention is all you need.
\newblock {\em Advances in neural information processing systems}, 30, 2017.

\bibitem{wang2019planit}
Kai Wang, Yu-An Lin, Ben Weissmann, Manolis Savva, Angel~X Chang, and Daniel
  Ritchie.
\newblock Planit: Planning and instantiating indoor scenes with relation graph
  and spatial prior networks.
\newblock {\em ACM Transactions on Graphics (TOG)}, 38(4):1--15, 2019.

\bibitem{wang2018deep}
Kai Wang, Manolis Savva, Angel~X Chang, and Daniel Ritchie.
\newblock Deep convolutional priors for indoor scene synthesis.
\newblock {\em ACM Transactions on Graphics (TOG)}, 37(4):1--14, 2018.

\bibitem{wang2021sceneformer}
Xinpeng Wang, Chandan Yeshwanth, and Matthias Nie{\ss}ner.
\newblock Sceneformer: Indoor scene generation with transformers.
\newblock In {\em 2021 International Conference on 3D Vision (3DV)}, pages
  106--115. IEEE, 2021.

\bibitem{xie2013reshuffle}
Hualiang Xie, Wenzhuo Xu, and Bin Wang.
\newblock Reshuffle-based interior scene synthesis.
\newblock In {\em Proceedings of the 12th ACM SIGGRAPH International Conference
  on Virtual-Reality Continuum and Its Applications in Industry}, pages
  191--198, 2013.

\bibitem{xu2013sketch2scene}
Kun Xu, Kang Chen, Hongbo Fu, Wei-Lun Sun, and Shi-Min Hu.
\newblock Sketch2scene: Sketch-based co-retrieval and co-placement of 3d
  models.
\newblock {\em ACM Transactions on Graphics (TOG)}, 32(4):1--15, 2013.

\bibitem{yang2021scene}
Haitao Yang, Zaiwei Zhang, Siming Yan, Haibin Huang, Chongyang Ma, Yi Zheng,
  Chandrajit Bajaj, and Qixing Huang.
\newblock Scene synthesis via uncertainty-driven attribute synchronization.
\newblock In {\em Proceedings of the IEEE/CVF International Conference on
  Computer Vision}, pages 5630--5640, 2021.

\bibitem{yang2021indoor}
Ming-Jia Yang, Yu-Xiao Guo, Bin Zhou, and Xin Tong.
\newblock Indoor scene generation from a collection of semantic-segmented depth
  images.
\newblock In {\em Proceedings of the IEEE/CVF International Conference on
  Computer Vision}, pages 15203--15212, 2021.

\bibitem{yeh2012synthesizing}
Yi-Ting Yeh, Lingfeng Yang, Matthew Watson, Noah~D Goodman, and Pat Hanrahan.
\newblock Synthesizing open worlds with constraints using locally annealed
  reversible jump mcmc.
\newblock {\em ACM Transactions on Graphics (TOG)}, 31(4):1--11, 2012.

\bibitem{yu2011make}
Lap~Fai Yu, Sai~Kit Yeung, Chi~Keung Tang, Demetri Terzopoulos, Tony~F Chan,
  and Stanley~J Osher.
\newblock Make it home: automatic optimization of furniture arrangement.
\newblock {\em ACM Transactions on Graphics (TOG)-Proceedings of ACM SIGGRAPH
  2011, v. 30,(4), July 2011, article no. 86}, 30(4), 2011.

\bibitem{zhang2021holistic}
Cheng Zhang, Zhaopeng Cui, Yinda Zhang, Bing Zeng, Marc Pollefeys, and
  Shuaicheng Liu.
\newblock Holistic 3d scene understanding from a single image with implicit
  representation.
\newblock In {\em Proceedings of the IEEE/CVF Conference on Computer Vision and
  Pattern Recognition}, pages 8833--8842, 2021.

\bibitem{zhang2019survey}
Song-Hai Zhang, Shao-Kui Zhang, Yuan Liang, and Peter Hall.
\newblock A survey of 3d indoor scene synthesis.
\newblock {\em Journal of Computer Science and Technology}, 34(3):594--608,
  2019.

\bibitem{zhang2020deep}
Zaiwei Zhang, Zhenpei Yang, Chongyang Ma, Linjie Luo, Alexander Huth, Etienne
  Vouga, and Qixing Huang.
\newblock Deep generative modeling for scene synthesis via hybrid
  representations.
\newblock {\em ACM Transactions on Graphics (TOG)}, 39(2):1--21, 2020.

\bibitem{zhu2018modeling}
Ji-Zhao Zhu, Yan-Tao Jia, Jun Xu, Jian-Zhong Qiao, and Xue-Qi Cheng.
\newblock Modeling the correlations of relations for knowledge graph embedding.
\newblock {\em Journal of Computer Science and Technology}, 33(2):323--334,
  2018.

\end{thebibliography}
}
\clearpage

\appendix
In this supplementary material, we provide details of our network architecture in Sec.~\ref{sec:network}, additional details of loss function design in Sec.~\ref{sec:loss}, data collection setup for 3D-Front in Sec.~\ref{sec:dataproc}, latent vector optimization setup for single-view reconstruction in Sec.~\ref{sec:optimsvr}, additional  details of baselines in Sec.~\ref{sec:baselines}, additional shape retrieval details in Sec.~\ref{sec:retrieval}, additional ablation studies in Sec.~\ref{sec:ablation}, and more qualitative results on scene synthesis and single-view reconstruction in Sec.~\ref{sec:qualitative}. \textit{Our code and data will be publicly released.}
\section{Network Specifications}
\label{sec:network}
We detail the full list of layer parameters in this section. We denote the MLP layers in our network uniformly by $\text{MLP}[l_{1},l_{2},...,l_{d}]$, where $l_{i}$ is the number of neurons in the $i$-th layer. Each fully connected layer is followed by a ReLU layer except the final one. We set the latent hypersphere dimension $D_{z}$=512 in Sec.~3.1. Every generated object feature $\bm{x}_{1},...,\bm{x}_{N}$ and the start token $\bm{x}_{0}$ are with the dimension of $D_{z}$=512 as well.

\subsection{Permutation-invariant Transformer}
From the input latent vector $\bm{z}\in\mathbb{R}^{D_{z}}$, we use a transformer to generate $N$ object features autoregressively. $N$ is the maximal number of objects in a scene. For ScanNet scenes~\cite{dai2017scannet}, we use $N=53$. For 3D-Front~\cite{fu20213d} bedrooms and living rooms, we use $N=13, 28$, respectively. We use the transformer library from \cite{fasttransformers} to build the architecture. Since our method does not require 3D supervision, we train the transformer without using teacher-forcing strategy.

\paragraph{Transformer Encoder}
With the previous object features $\bm{x}_{0}, \bm{x}_{1},...,\bm{x}_{k-1}$ as input, we adopt a transformer encoder to produce the scene context feature $\bm{F}_{k}\in\mathbb{R}^{k\times 512}$. The encoder consists of a single layer of multi-head self attention without positional encoding, followed by a feed forward network. We use four heads in our multi-head attention, with the input and output dimension $d_{model}$=512. The feed forward network is a two-layer $\text{MLP}[1024,512]$ but with GeLU activation. For other parameters we keep the default setting as in \cite{fasttransformers}.

\paragraph{Transformer Decoder}
The transformer decoder generates the next object feature $\bm{x}_{k}$ from $\bm{F}_{k}\in\mathbb{R}^{k\times 512}$. It is a single layer of multi-head cross attention without positional encoding, followed by a feed forward network. Similar to the encoder, we use four heads in the cross attention module with $d_{model}$=512. For the cross attention module, we use the scene context $\bm{F}_{k}$ as keys and values, and the latent vector $\bm{z}$ as the query to infer the next object $\bm{x}_{k}$. Similar to the encoder, the feed forward network is a two-layer $\text{MLP}[1024,512]$ with GeLU activation. We keep the default setting for other parameters.

This generates an object feature sequence $\{\bm{x}_{k}\}$, $\bm{x}_{k}\in\mathbb{R}^{512}, k=1,...,N$, from our transformer.

\begin{figure}[!t]
	\centering
	\includegraphics[width=0.5\textwidth]  
	{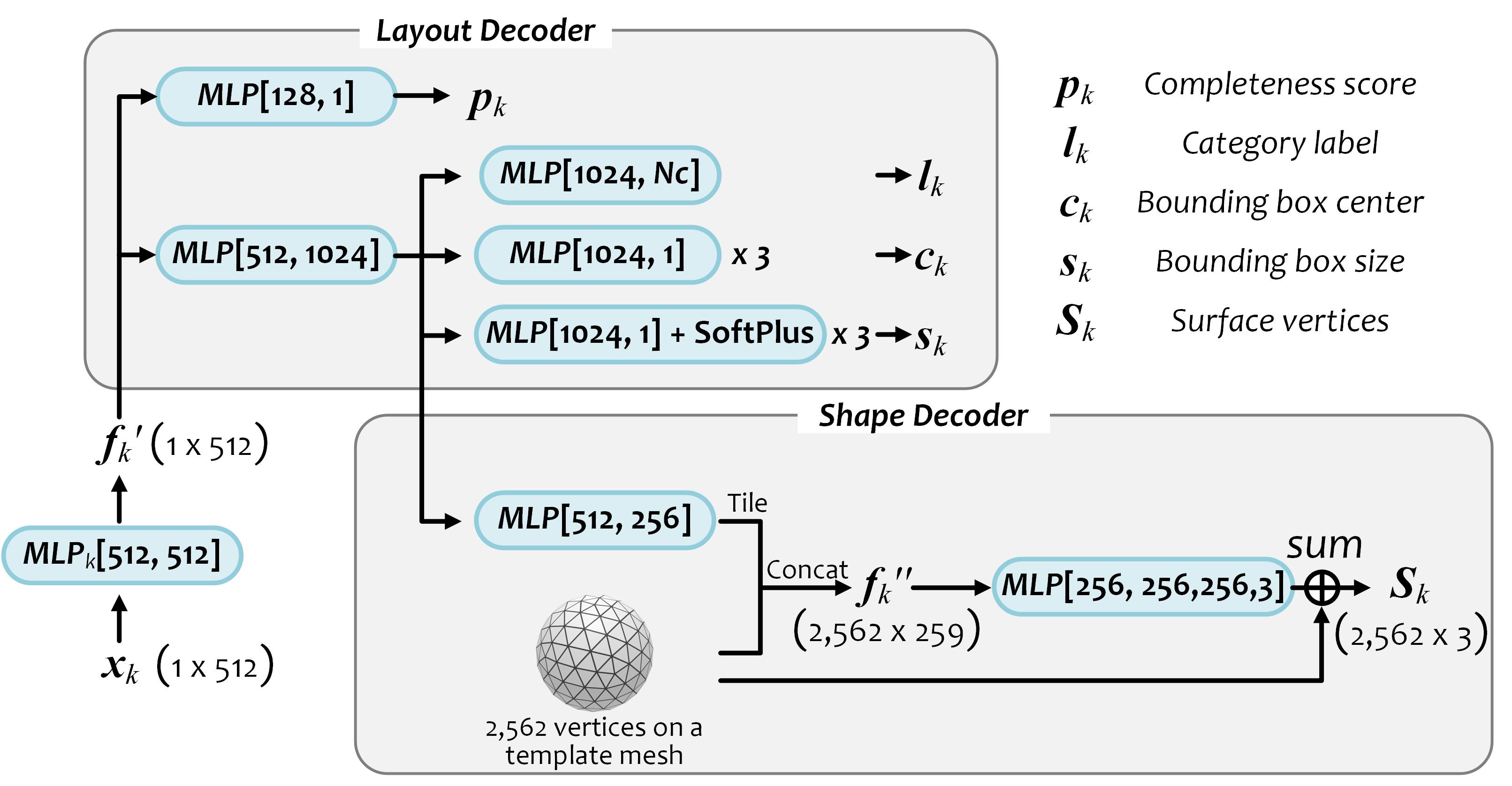}
	\caption{Network architecture details in layout and shape decoders.}
	\label{fig:layout_shape_decoder}
	\vspace{-1em}
\end{figure}

\subsection{Layout and Shape Decoder}
For each object feature $\bm{x}_{k}$, we decode its class label $\bm{l}_{k}\in\mathbb{L}$, axis-aligned 3D bounding box with size $\bm{s}_{k}\in\mathbb{R}^{3}$ and center $\bm{c}_{k}\in\mathbb{R}^{3}$, as well as completeness score $\bm{p}_{k}$, via our layout decoder, and decode its mesh surface $\bm{S}_{k}$ via our shape decoder. We detail the layer specifications with parameters in Fig.~\ref{fig:layout_shape_decoder}. For $\bm{s}_{k}$, we process it with a SoftPlus layer to make sure object sizes are positive values. $N_{c}$ denotes the number of object class categories. For ScanNet, we use the official `nyu40class' object category split and use 19 common indoor furniture categories. For 3D-Front, we follow \cite{Paschalidou2021NEURIPS} and use 22 object categories for bedrooms, and 15 object categories for living rooms. The category names are listed below. All baselines are trained and tested with a unified category split.

\begin{python}
# 19 ScanNet categories
['void', 'bathtub', 'bed', 'bookshelf',
'cabinet', 'chair', 'counter', 'desk',
'dresser', 'lamp', 'night stand',
'refridgerator', 'shelves', 'sink', 'sofa',
'table', 'television', 'toilet', 'whiteboard']

# 22 3D-Front bedroom categories
['void', 'armchair', 'bookshelf', 'cabinet',
'ceiling_lamp', 'chair', 'children_cabinet',
'coffee_table', 'desk', 'double_bed',
'dressing_chair', 'dressing_table', 'kids_bed',
'nightstand', 'pendant_lamp', 'shelf',
'single_bed', 'sofa', 'stool', 'table',
'tv_stand', 'wardrobe']

# 15 3D-Front living room categories
['void', 'armchair', 'bookshelf', 'cabinet',
'ceiling_lamp', 'chair', 'coffee_table', 'desk',
'pendant_lamp', 'shelf', 'sofa', 'stool',
'table', 'tv_stand', 'wardrobe']
\end{python}

In our layout decoder, all objects are constrained to be located above the floor, which can be achieved by 
\begin{equation}
	\bm{c}'_{y} = \bm{c}_{y} + \bm{s}_{y}/2; \ \ \ \ \bm{c}_{y}>0, s_{y}>0,
\end{equation}
where $c_{y}$ is the vertical coordinate of an object center. $c_{y}\geq 0$ as it is outputted after a ReLU layer; $\bm{s}_{y}$ is the height of an object bounding box. In the following calculations, we use $\bm{c}'_{y}$ as the vertical coordinate of the object center.

\subsection{Differentiable Rendering Setup}
Given a scene with generated objects, we render it back to input views with differentiable rendering. Thus in each input view, there is a rendered instance map with object silhouettes.
We use PyTorch3D~\cite{ravi2020accelerating} to implement our rendering process.

During mini-batch training, we randomly select 20 views from all frames in each scene. We render the generated scene back to the 20 views with a lower resolution to the original input image (1/4 of the original\footnote{Images in ScanNet and 3D-Front have resolution  1296$\times$968 and 480$\times$360 respectively.}) due to the high GPU memory consumption of differentiable rendering. We render 50 faces per pixel in mesh rasterization~\cite{ravi2020accelerating}. Blur radius and blend sigma are 1e-4. A soft silhouette shader is applied to render instance silhouettes. Thus, in each view we obtain two maps: one differentiable silhouette map and one instance ID map, from which we can obtain the instance silhouette $\bm{r}_{k}^{p}$ of each object under 20 views.

\section{Additional Loss Details}
\label{sec:loss}
\subsection{Hungarian Matching}
\label{sec:hungarian}
For each object prediction $\bm{o}_{k}$ in a scene ($k$=1,...,$N$), we have its class label $\bm{l}_{k}$ and a set of 2D bounding boxes $\bm{B}_{k}=\{\bm{b}^{1},..., \bm{b}^{T}\}_{k}$ in all input views $T$, where $T$ indicates 20 random views in a mini-batch. Additionally, we also have ground-truth objects $\{\bm{o}^{gt}_{j}\}$ in this scene ($j$=1,...,$n$). For each $\bm{o}^{gt}_{j}$, it has a class label $\bm{l}_{j}^{gt}$ and a set of 2D bounding boxes $\bm{B}^{gt}_{j}=\{\bm{b}^{1},...,\bm{b}^{T_{j}}\}^{gt}_{j}$ in $T_{j}$ views, where $T_{j}$ denotes all the observable views of $\bm{o}^{gt}_{j}$ ($T_{j}\subseteq T$). $N$ is the maximal object number among all scenes. $n$ is the object number of this target scene in a mini-batch. 

For each object prediction $\bm{o}_{k}$, we use the Hungarian algorithm to find its optimal bipartite matching $\bm{o}^{gt}_{\sigma(k)}$ for loss calculation in Sec.~3.4, ${\sigma(k)}=1,...,n$ or $\varnothing$.
We use the implementation of Hungarian algorithm from \cite{carion2020end}.

In our case, a prediction $\bm{o}_{k}$ and a ground-truth object $\bm{o}^{gt}_{\sigma(k)}$ are characterized by $(\bm{B}_{k}, \bm{l}_{k})$ and $(\bm{B}^{gt}_{\sigma(k)}, \bm{l}^{gt}_{\sigma(k)})$, respectively. As in \cite{carion2020end}, our matching cost takes into account the similarity between both class labels \textless$\bm{l}_{k}, \bm{l}^{gt}_{\sigma(k)}$\textgreater and 2D bounding boxes \textless$\bm{B}_{k}, \bm{B}^{gt}_{\sigma(k)}$\textgreater. Thus the matching problem can be formulated as
\begin{equation}
\hat{\sigma}=\underset{\sigma}{\text{argmin}}\sum_{k=1}^{n}\left[\mathcal{L}_{\text{match}}^{l}(\bm{l}_{k}, \bm{l}^{gt}_{\sigma(k)}) +  \lambda_{B}\mathcal{L}_{\text{match}}^{B}(\bm{B}_{k}, \bm{B}^{gt}_{\sigma(k)})\right],
\label{eq:hungarian}
\end{equation}
where $\mathcal{L}_{\text{match}}^{l}$ and $\mathcal{L}_{\text{match}}^{B}$ are pair-wise matching costs. Note that we find the bipartite matching for the first $n$ predictions only, i.e., $\{\bm{o}_{k}\}$, $k$=1, ..., $n$, $n \leq N$, and predict their completeness scores as ones. For the additional predictions $\{\bm{o}_{k}\}$, $k$=$n$+1, ..., $N$, we predict their completeness scores as zeros. The completeness score here indicates whether the generated scene is complete or not. We set $\lambda_{B}$=5 to balance the importance of the two costs.

We keep the definition of $\mathcal{L}_{\text{match}}^{l}$ as in \cite{carion2020end}, and formulate our $\mathcal{L}_{\text{match}}^{B}$ as
\begin{equation}
	\mathcal{L}_{\text{match}}^{B}(\bm{B}_{k}, \bm{B}^{gt}_{\sigma(k)}) = \frac{1}{T_{\sigma{(k)}}} \sum_{p\in T_{\sigma{(k)}}} L_{1}(\bm{b}^{p}_{k}, \bm{b}^{p,gt}_{\sigma{(k)}}),
\end{equation}
where $T_{\sigma{(k)}}$ denotes all visible views of $\bm{o}_{\sigma{(k)}}^{gt}$; $\bm{b}^{p}_{k}$ and $\bm{b}^{p,gt}_{\sigma{(k)}}$ correspond to the 2D bounding box of $\bm{o}_{k}$ and $\bm{o}^{gt}_{\sigma(k)}$ in view $p$, respectively. Each 2D bounding box is parameterized with a vector of $(x_{1},y_{1},x_{2},y_{2})\in[0,1]^{4}$, which contains the 2D coordinates of upper-left and bottom-right box corners relative to the image size.

By solving Eq.~\ref{eq:hungarian}, we can assign a ground-truth object $\bm{o}^{gt}_{\sigma(k)}$ to each prediction $\bm{o}_{k}$, which facilitates our view loss calculation in Sec.3.4. 

\begin{figure}[!t]
	\centering
	\includegraphics[width=0.45\textwidth]  
	{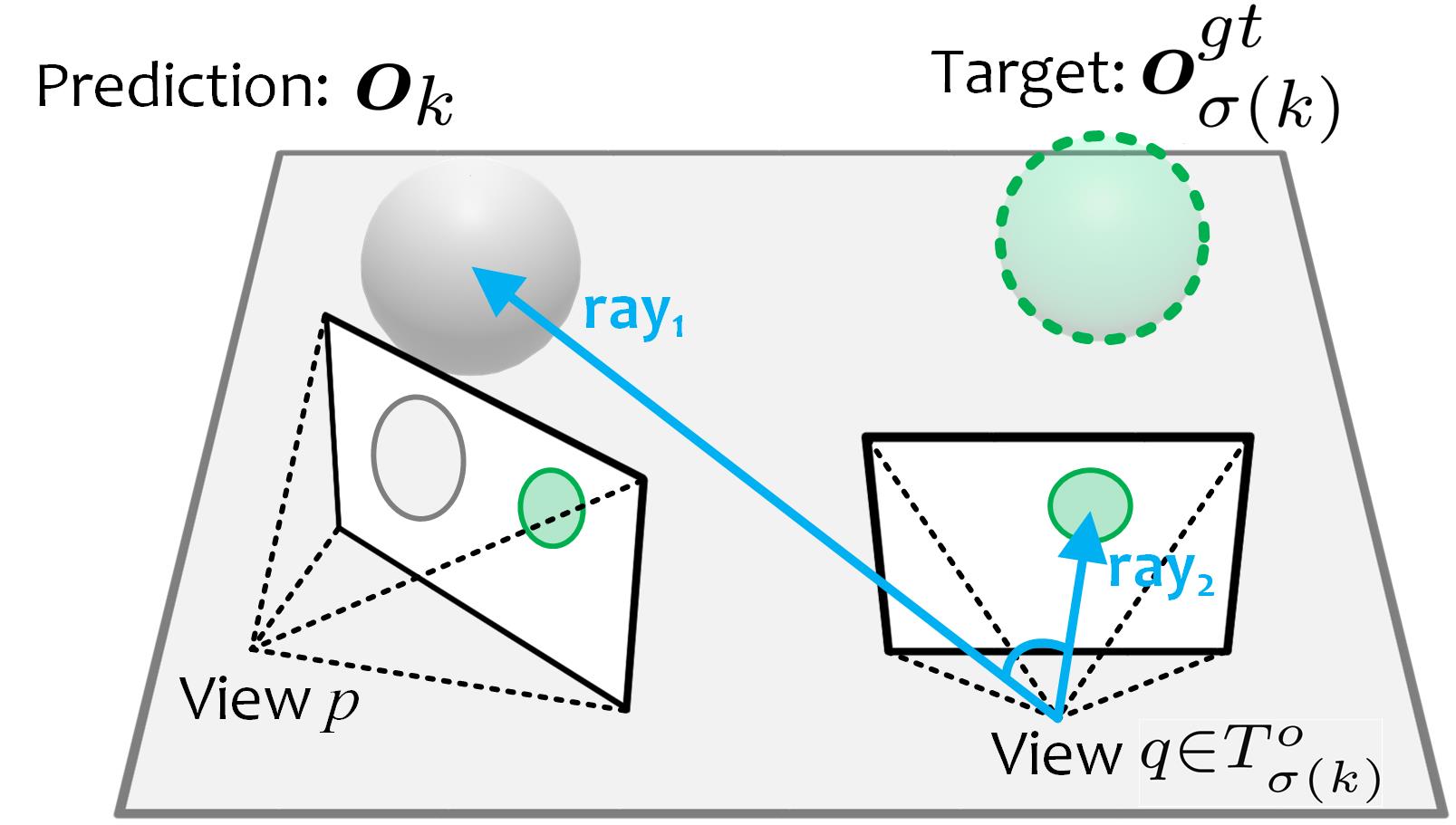}
	\caption{Illustration of frustum loss. $p$ and $q$ are two views where the ground-truth object $\bm{o}^{gt}_{\sigma(k)}$ is visible, while our prediction $\bm{o}_{k}$ is visible in view $p$ only. We design a frustum loss, by maximizing the cosine similarity between ray\textsubscript{1} and ray\textsubscript{2}, to optimize $\bm{o}_{k}$'s location and make it visible in view $q$ as well. }
	\label{fig:frustum_loss}
	\vspace{-1em}
\end{figure}

\subsection{Frustum Loss}
As a component in layout loss $\mathcal{L}_{L}$ (see Sec.3.4), the frustum loss is designed to optimize the 3D center $\bm{c}_{k}$ of a prediction $\bm{o}_{k}$ if it is not located inside a view frustum $q\in T_{\sigma{(k)}}$ from its matched ground-truth object $\bm{o}^{gt}_{\sigma(k)}$. An illustration of our frustum loss is shown in Fig.~\ref{fig:frustum_loss}.

As in our paper, we denote $T_{\sigma{(k)}}$ as all the visible views of the ground-truth object $\bm{o}^{gt}_{\sigma(k)}$. We calculate the average view loss between $\bm{o}_{k}$ and $\bm{o}^{gt}_{\sigma(k)}$ over all views in $T_{\sigma{(k)}}$, where the view loss includes object classification loss $\mathcal{L}_{l}$, 2D bounding box loss $\mathcal{L}_{box}$, completeness loss $\mathcal{L}_{p}$, frustum loss $\mathcal{L}_{f}$, and shape loss $\mathcal{L}_{S}$ (see Sec.3.4). However, when $\bm{o}_{k}$ is out of some view frustum $q\in T_{\sigma{(k)}}$, the calculation of $\mathcal{L}_{box}$ and $\mathcal{L}_{S}$ are meaningless, making convergence difficult. In this case, we design a frustum loss to force $\bm{o}_{k}$ to move towards the view frustum of $q$.

Assume that $\bm{o}_{k}$ is outside of $T^{o}_{\sigma{(k)}}$ views, where $T^{o}_{\sigma{(k)}}\subseteq T_{\sigma{(k)}}$. Then, we formulate the frustum loss $\mathcal{L}_{f}$ of $\bm{o}_{k}$ as
\begin{equation}
	\mathcal{L}_{f} = \frac{1}{|T^{o}_{\sigma{(k)}}|}\sum_{q\in T^{o}_{\sigma{(k)}}} 1-\text{Cosine}(\bm{c}_{k}-\bm{c}_{cam}^{q}, \bm{c}^{q,gt}_{\sigma{(k)}}-\bm{c}_{cam}^{q}),
\end{equation}
where $|T^{o}_{\sigma{(k)}}|$ is the number of views in $T^{o}_{\sigma{(k)}}$; $\bm{c}_{k}$ is the 3D center of $\bm{o}_{k}$; $\bm{c}_{cam}^{q}$ is the 3D camera position at view $q$. $\bm{c}^{q,gt}_{\sigma{(k)}}$ is the center of $\bm{b}_{\sigma(k)}^{q,gt}$ in 3D space, where $\bm{b}^{q,gt}_{\sigma{(k)}}$ is the 2D bounding box of $\bm{o}^{gt}_{\sigma(k)}$ in view $q$. Therefore, $\bm{c}_{k}-\bm{c}_{cam}^{q}$ denotes a ray in world system from the camera center $\bm{c}_{cam}^{q}$ to the 3D object center $\bm{c}_{k}$, and $\bm{c}^{q,gt}_{\sigma{(k)}}-\bm{c}_{cam}^{q}$ is the ray from $\bm{c}_{cam}^{q}$ to the 2D ground-truth center on the image plane.

For an object $\bm{o}_{k}$ invisible to views $T^{o}_{\sigma{(k)}}$, we minimize the frustum loss while switching off the box loss $\mathcal{L}_{box}$ and shape loss $\mathcal{L}_{S}$ under those views. For visible views, we consider all view losses in Sec.~3.4 while switching off the frustum loss $\mathcal{L}_{f}$, because $\bm{o}_{k}$ in visible views have valid 2D bounding boxes and instance masks.

\section{Additional Rendering Details for 3D-Front}
\label{sec:dataproc}
We use BlenderProc~\cite{denninger2019blenderproc} to sample cameras and render 2D images in 3D-Front scenes. Each scene in 3D-Front is an apartment which has several room types (bedroom, living room, library, etc.). In each scene, we uniformly sample at most 100 view points, with each view rendered into a 360$\times$480 image with field of view of 90 degrees. The view number in each room is proportional to its floor area. The average object number captured in each view is 3.89 and 5.79, for bedroom and living room respectively, while the average object number contained in each room is 5.53 and 9.82 respectively. For each view, we export camera intrinsic and extrinsic parameters, instance masks, IDs, and category labels for our training.
\section{Optimization Setup for Single View Reconstruction}
\label{sec:optimsvr}
In single-view scene reconstruction, we have an image with instance masks as the input. Our network and latent vectors are trained under multiple views. In this task, we freeze our pretrained network while only optimizing the latent vector for each single image, where instance masks are used for supervision. We use our view loss for this optimization but do not consider completeness loss in our final loss, because a single image is only a partial observation of an entire scene and we do not know how many objects this scene contains. We use RMSProp from PyTorch as the optimizer and train 1000 epochs with the initial learning rate at 0.01, which drops by 0.1x after 500 epochs.

After training, we input the optimized latent vector to our network and generate a set of objects. We output the first $n$ objects for our qualitative and quantitative evaluation, where $n$ is the object number in each input image.

\begin{figure*}[!ht]
	\centering
	\begin{subfigure}[t]{0.16\textwidth}
		\includegraphics[width=\textwidth]
		{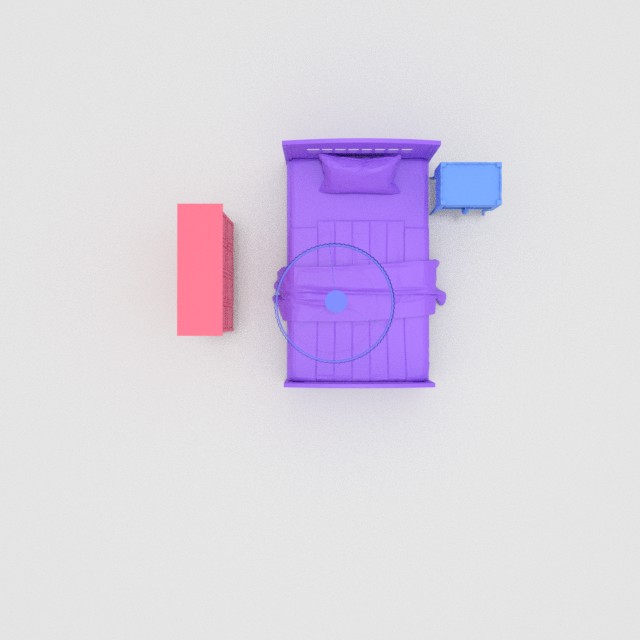}
		\includegraphics[width=\textwidth]
		{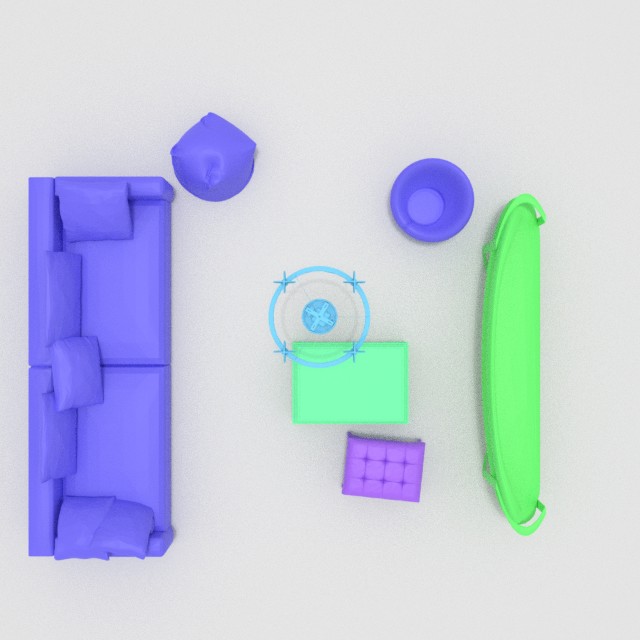}
		\includegraphics[width=\textwidth]
		{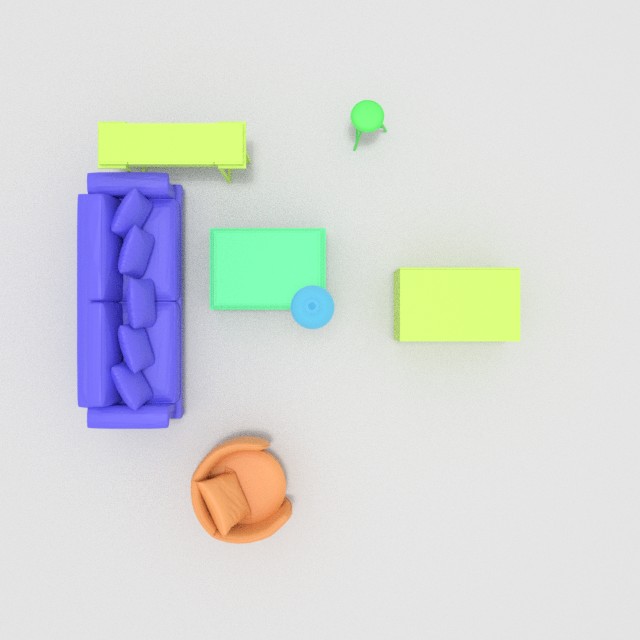}
		\includegraphics[width=\textwidth]
		{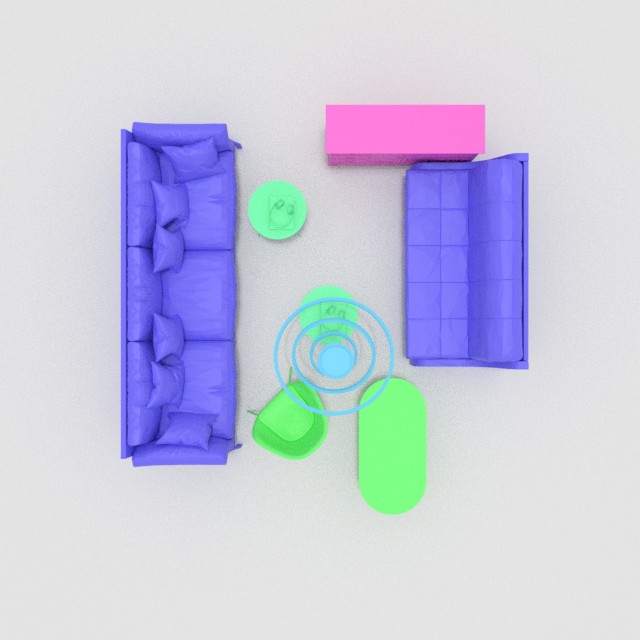}
	\end{subfigure}
	\begin{subfigure}[t]{0.16\textwidth}
		\includegraphics[width=\textwidth]
		{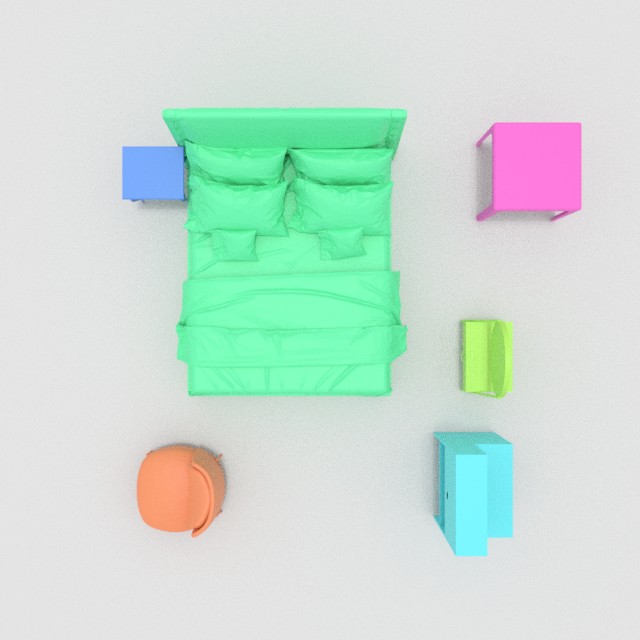}
		\includegraphics[width=\textwidth]
		{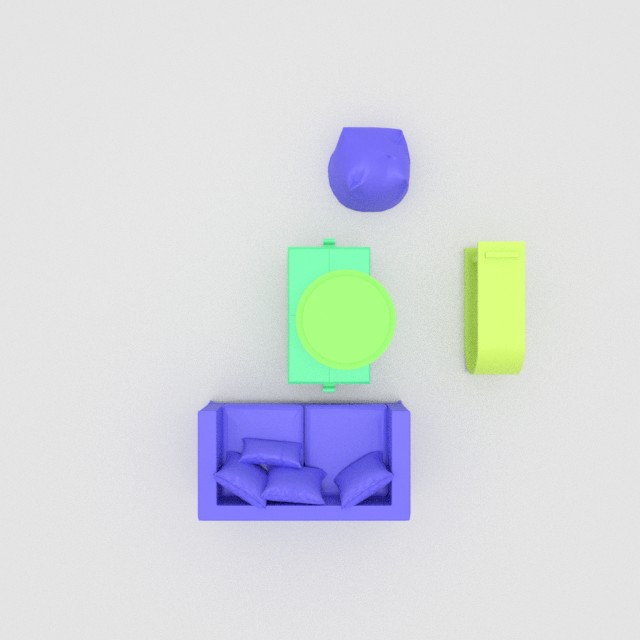}
		\includegraphics[width=\textwidth]
		{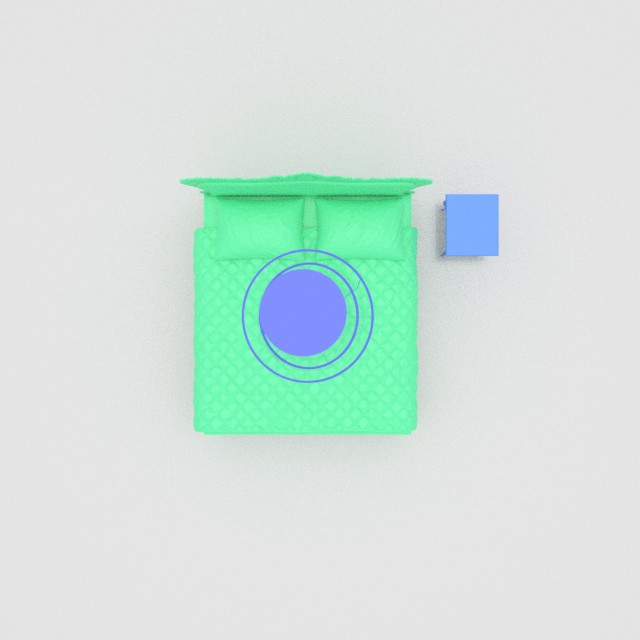}
		\includegraphics[width=\textwidth]
		{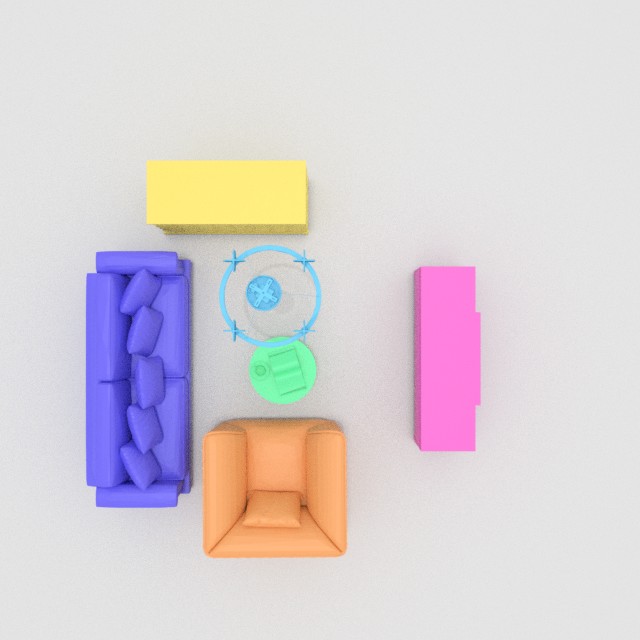}
	\end{subfigure}
	\begin{subfigure}[t]{0.16\textwidth}
		\includegraphics[width=\textwidth]
		{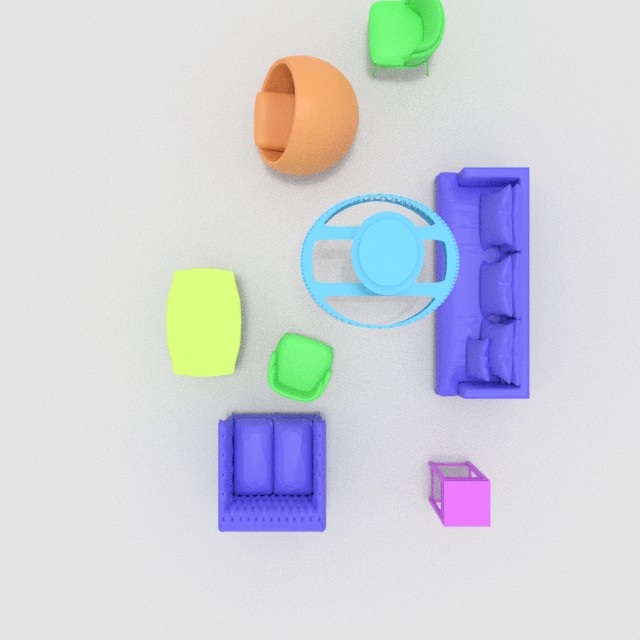}
		\includegraphics[width=\textwidth]
		{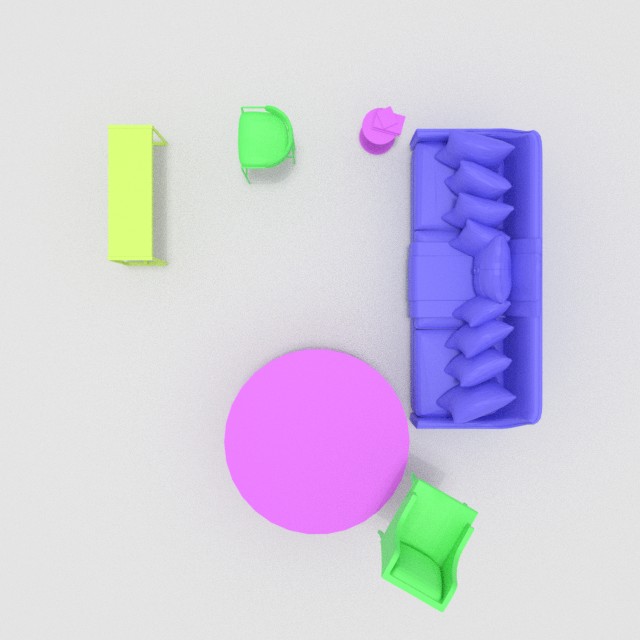}
		\includegraphics[width=\textwidth]
		{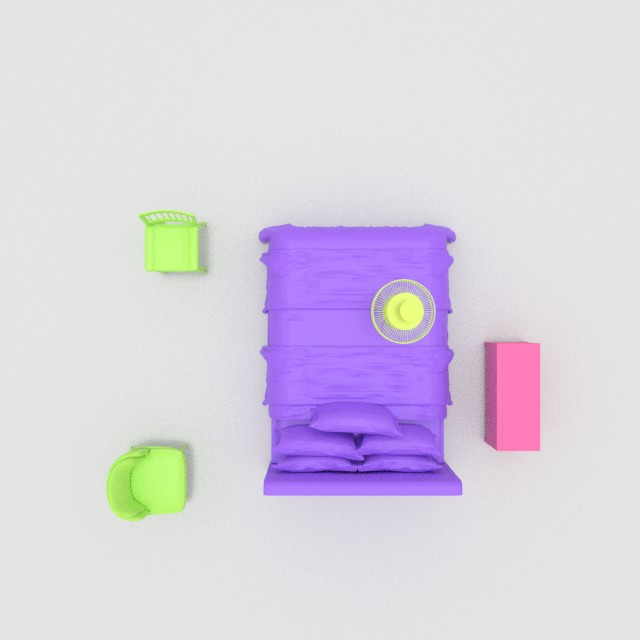}
		\includegraphics[width=\textwidth]
		{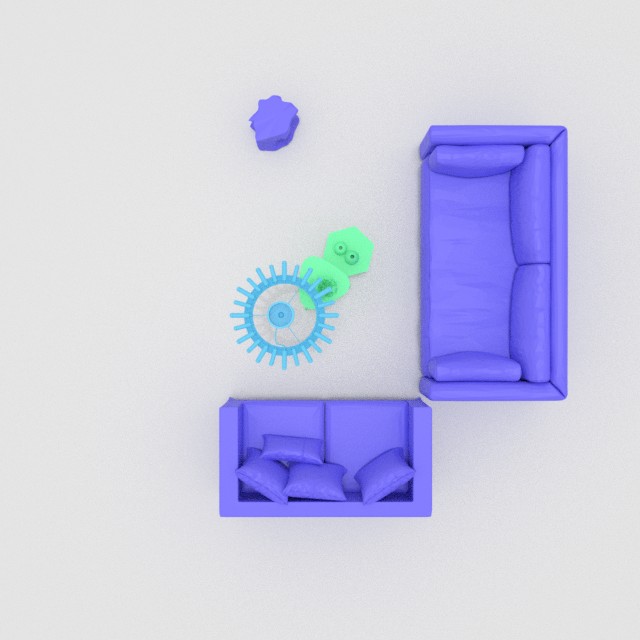}
	\end{subfigure}
	\begin{subfigure}[t]{0.16\textwidth}
		\includegraphics[width=\textwidth]
		{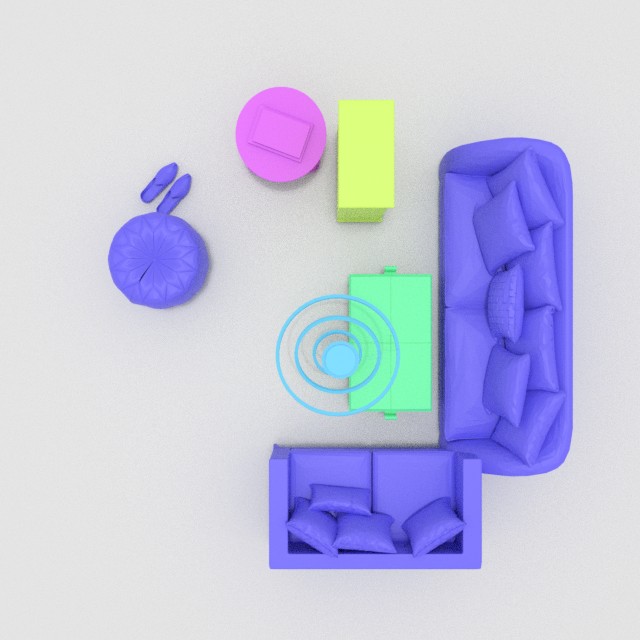}
		\includegraphics[width=\textwidth]
		{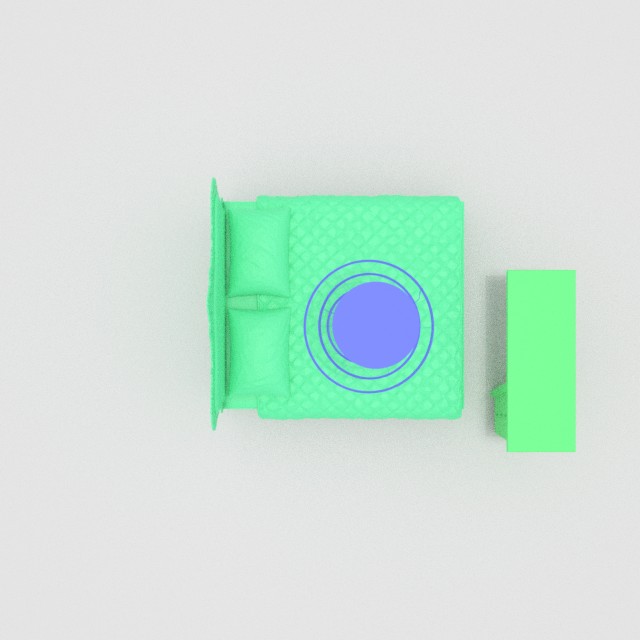}
		\includegraphics[width=\textwidth]
		{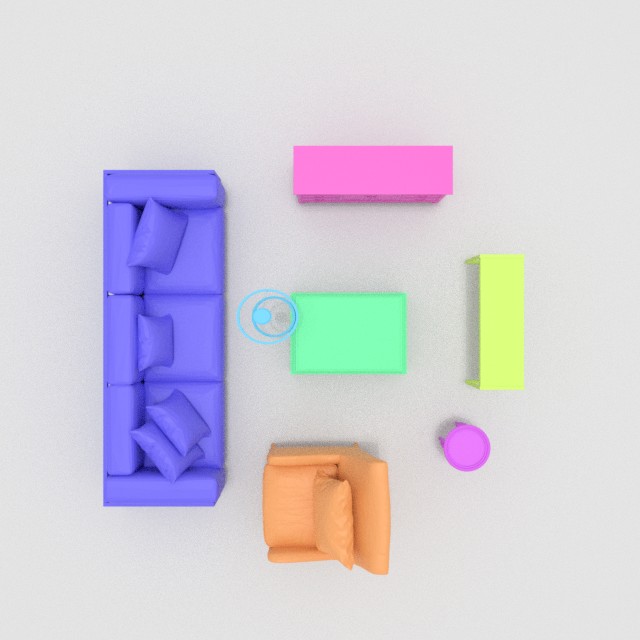}
		\includegraphics[width=\textwidth]
		{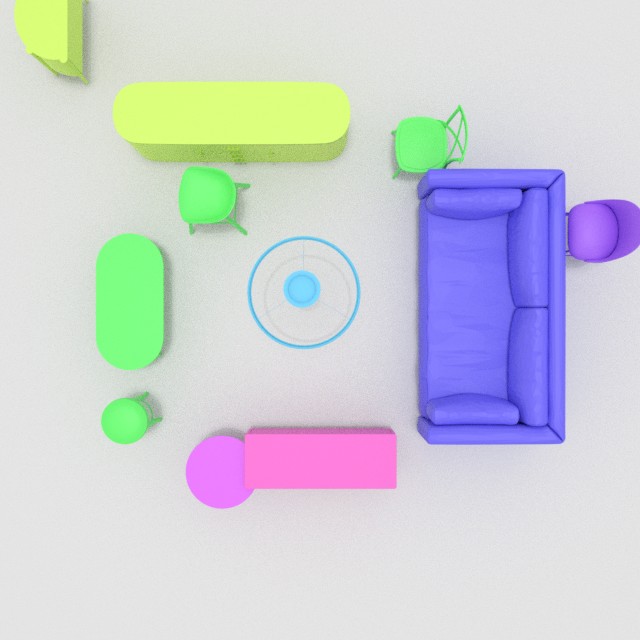}
	\end{subfigure}
	\begin{subfigure}[t]{0.16\textwidth}
		\includegraphics[width=\textwidth]
		{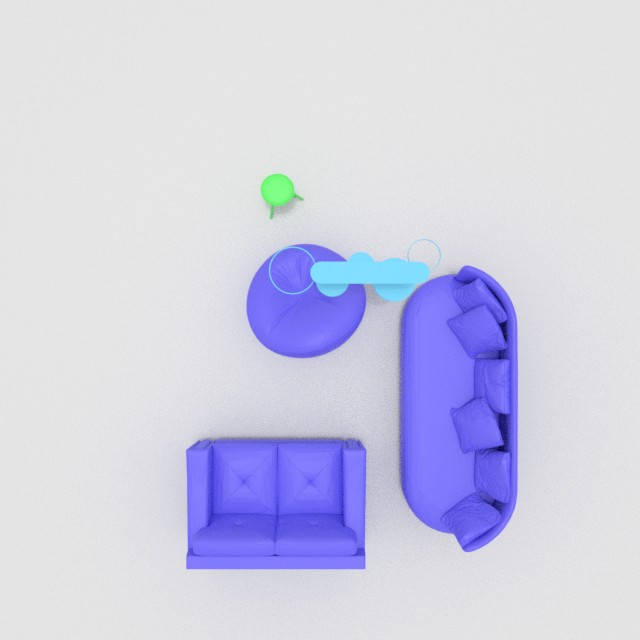}
		\includegraphics[width=\textwidth]
		{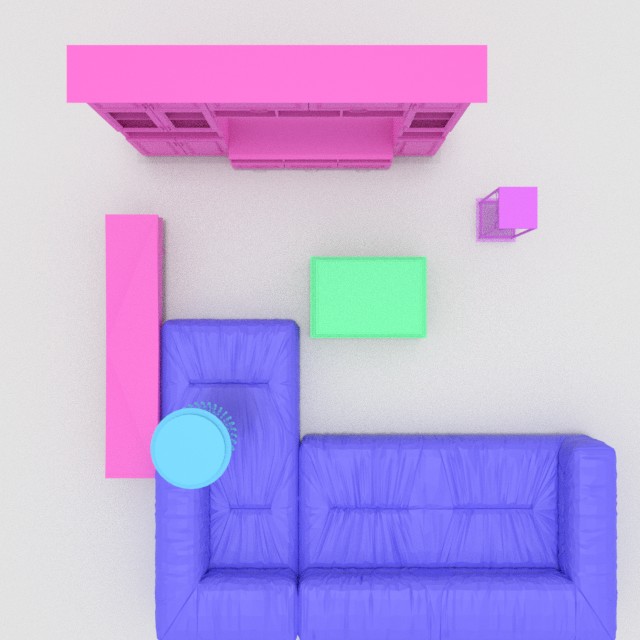}
		\includegraphics[width=\textwidth]
		{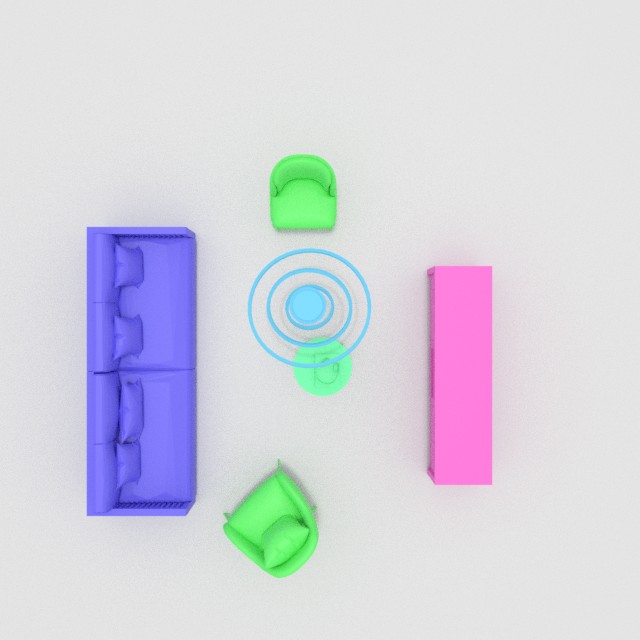}
		\includegraphics[width=\textwidth]
		{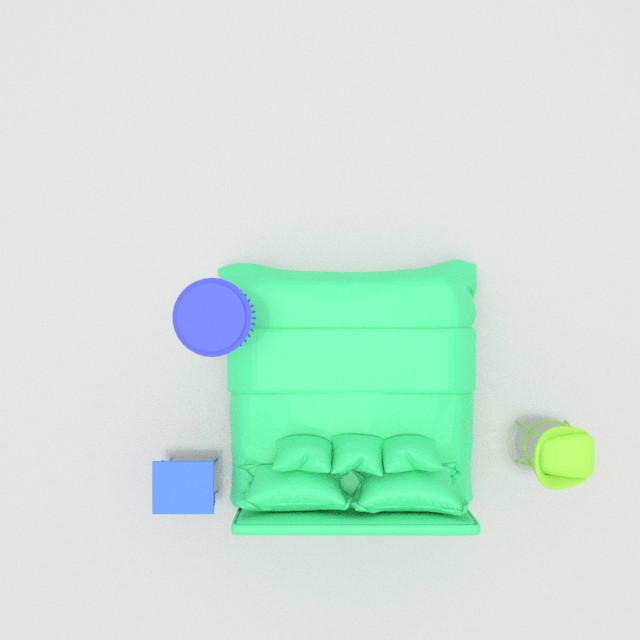}
	\end{subfigure}
	\begin{subfigure}[t]{0.16\textwidth}
		\includegraphics[width=\textwidth]
		{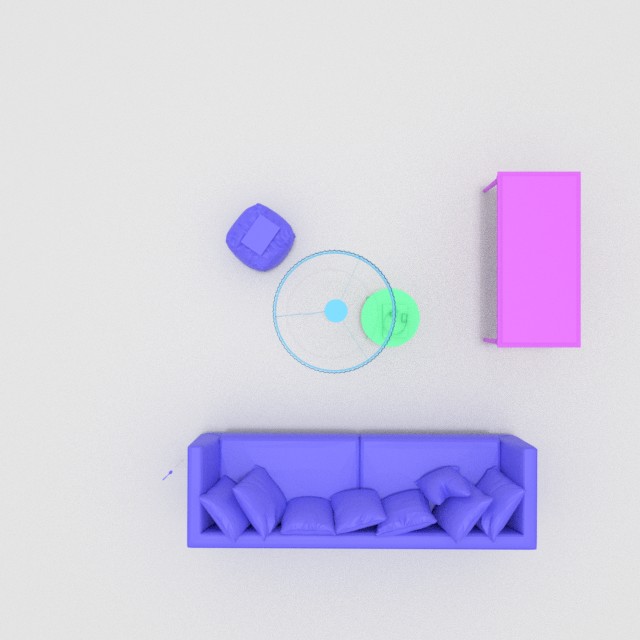}
		\includegraphics[width=\textwidth]
		{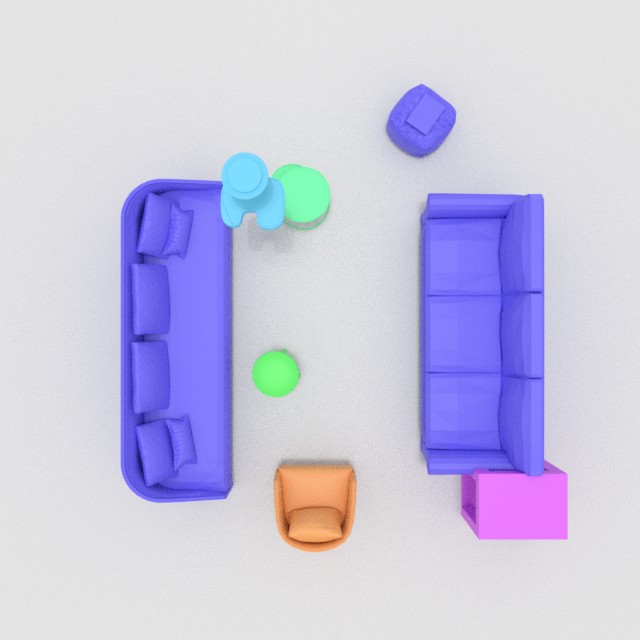}
		\includegraphics[width=\textwidth]
		{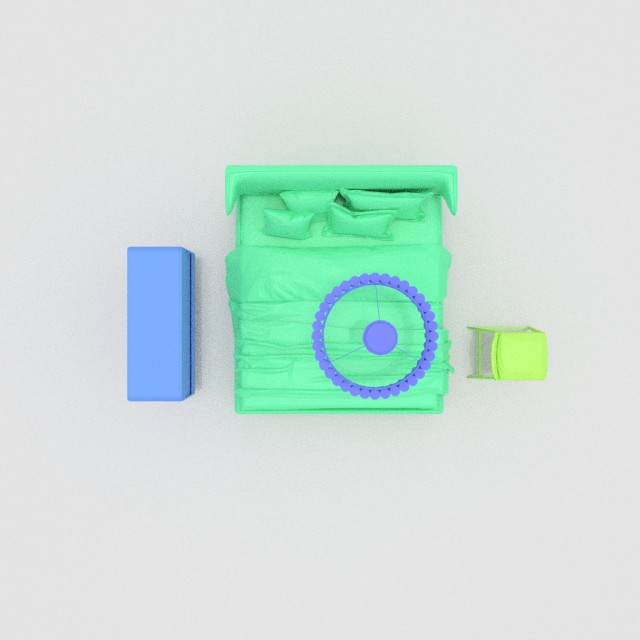}
		\includegraphics[width=\textwidth]
		{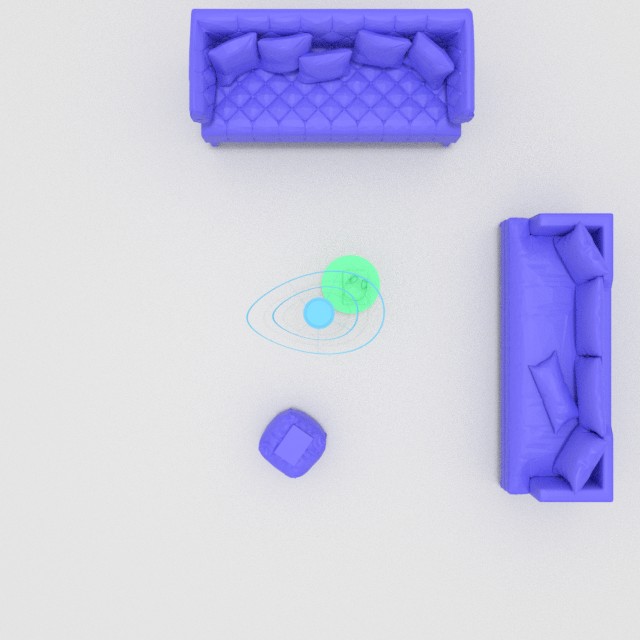}
	\end{subfigure}
	\caption{Additional qualitative results on 3D scene synthesis.}
	\label{fig:add_qualitative_scene_gen}
	\vspace{-1em}
\end{figure*}

\section{Baselines}
\label{sec:baselines}
\paragraph{ATISS-2D} We replace our transformer with the transformer encoder from ATISS~\cite{paschalidou2021atiss} to generate object features, while keeping other settings unchanged. Note that the transformer encoder in ATISS is also permutation-invariant. We use the latent vector (sampled from the hypersphere surface) as the empty embedding in ATISS transformer to autoregressively generate a set of object features, which are input to our layout decoder and shape decoder for scene generation.

\paragraph{GAN} We use our transformer as the generator to generate scenes and use the discriminator from \cite{yang2021indoor} to distinguish the generated scene to real scenes. We apply a discriminator loss here to replace our view loss. The generator predicts a scene from a random vector sampled from our latent space. Then we render the semantic maps of this scene back to input views, and use the discriminator from \cite{yang2021indoor} to classify if they are real or fake. Note that we do not consider the depth channel in the discriminator loss calculation, i.e., we only use the 2D semantic maps.

\paragraph{LSTM} We replace our transformer with a traditional LSTM RNN~\cite{lstm_tutorial} and keep other settings unchanged. It takes a latent vector $z$ as input and recursively generates a sequence of features.

\section{Additional Details of Shape Retrieval}
In Sec.~5, we apply a shape retrieval post-processing to search for a CAD model for our reconstructed mesh. Since we already predict an object mesh, shape retrieval from this prediction is much easier. For each predicted object $\bm{o}$, we have its category label $\bm{l}$, 3D bounding box $\bm{b}$ and mesh $\bm{m}$. For those objects whose center height $\leq 1$ meter, we extrude the bottom face of their 3D bounding box onto the floor. We search through the CAD models under the same category $\bm{l}$, and transform (move and scale) them to the 3D bounding box $\bm{b}$. All CAD models are augmented by rotating with 0, 90, 180, 270 degrees in the bounding box, then we calculate the Chamfer distance between the surface points from all augmented CAD models and the surface points from our mesh $\bm{m}$. The CAD model with the lowest Chamfer distance is retrieved. We repeat this process for all objects to model a 3D CAD scene.
\label{sec:retrieval}
\begin{table}[!h]
	\centering
	\resizebox{1\columnwidth}{!}{
		\begin{tabular}{|l|c c c|c c c|}
			\hline
			& \multicolumn{3}{c|}{Bedroom} & \multicolumn{3}{c|}{Living room}\\
			& FID ($\downarrow$) & SCA & KL ($\downarrow$) & FID ($\downarrow$) & SCA & KL ($\downarrow$)  \\
			\hline
			\hline
			25\% & 29.48 & 0.96 & 0.06 & 105.33 & 0.98 & 0.35\\
			50\% & 24.91 & 0.91 & 0.04
			& 44.85 & 0.98 & 0.04 \\
			Full & \textbf{21.59} & \textbf{0.85} & \textbf{0.03} & \textbf{40.47} & \textbf{0.96} & \textbf{0.02} \\
			\hline
	\end{tabular}}
	\caption{Ablation analysis on the number of views for training, evaluating scene generation.}
	\label{tab:ab_view_num}
	\vspace{-1em}
\end{table}

\begin{figure*}[!ht]
	\centering
	\begin{subfigure}[t]{0.155\textwidth}
		\includegraphics[width=\textwidth]
		{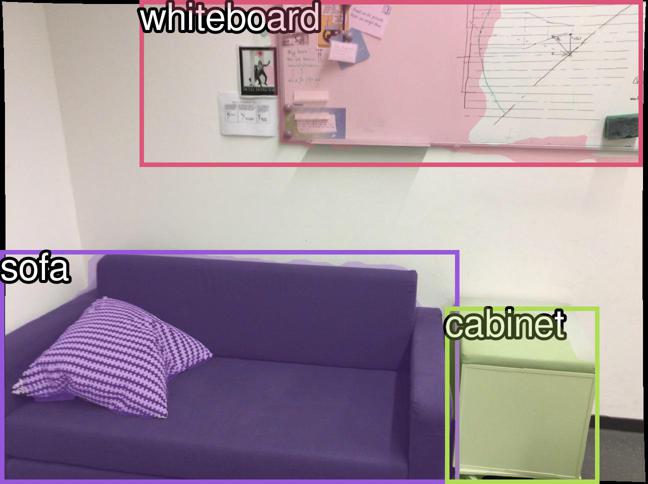}
		\includegraphics[width=\textwidth]
		{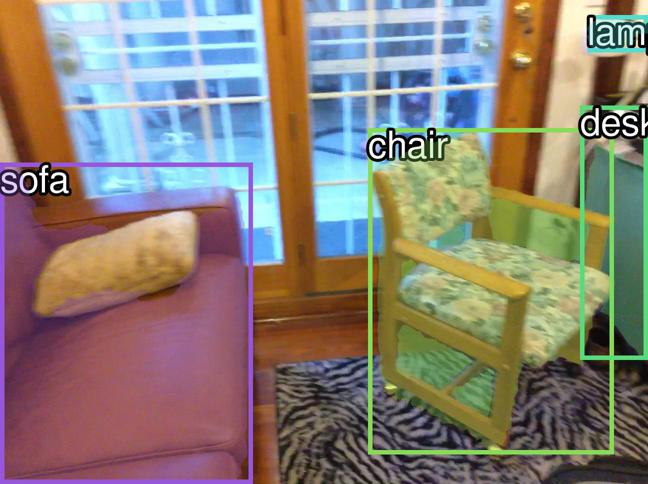}
		\includegraphics[width=\textwidth]
		{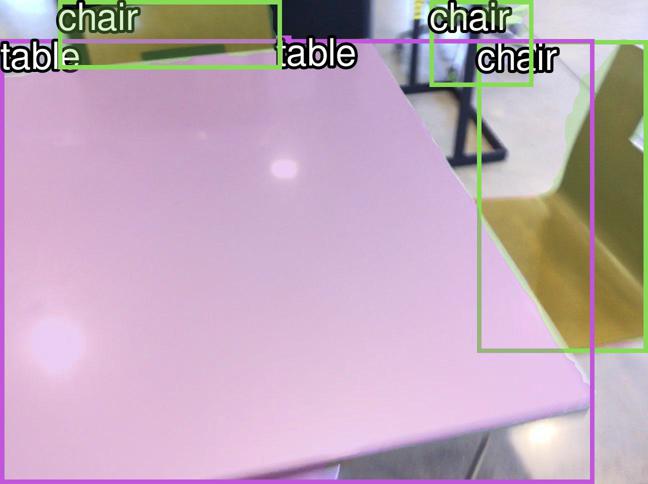}
		\includegraphics[width=\textwidth]
		{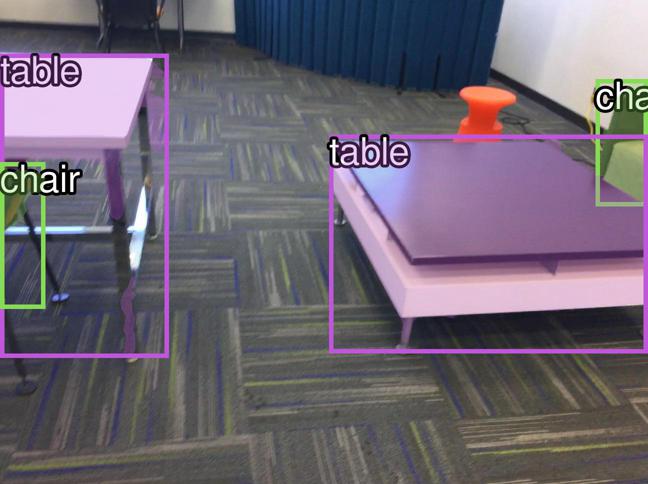}
		\includegraphics[width=\textwidth]
		{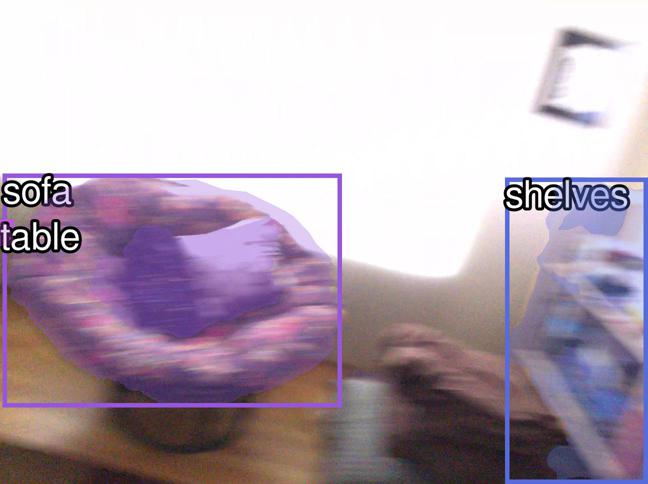}
		\includegraphics[width=\textwidth]
		{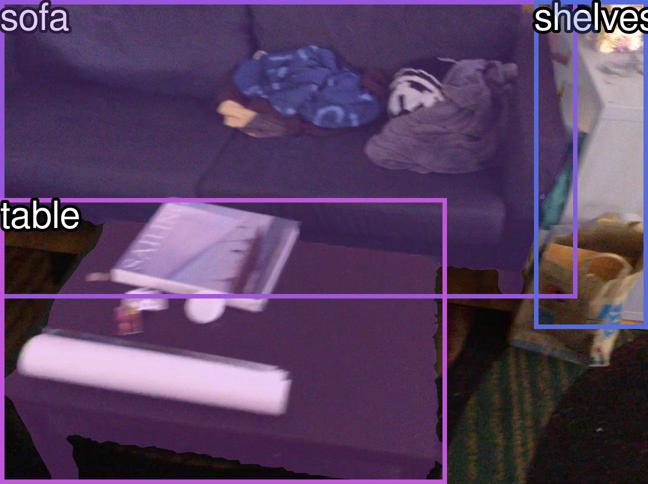}
		\includegraphics[width=\textwidth]
		{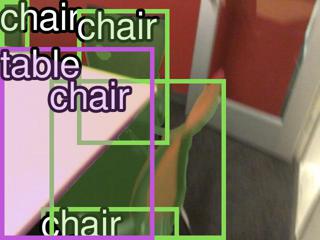}
	\end{subfigure}
	\begin{subfigure}[t]{0.155\textwidth}
		\includegraphics[width=\textwidth]
		{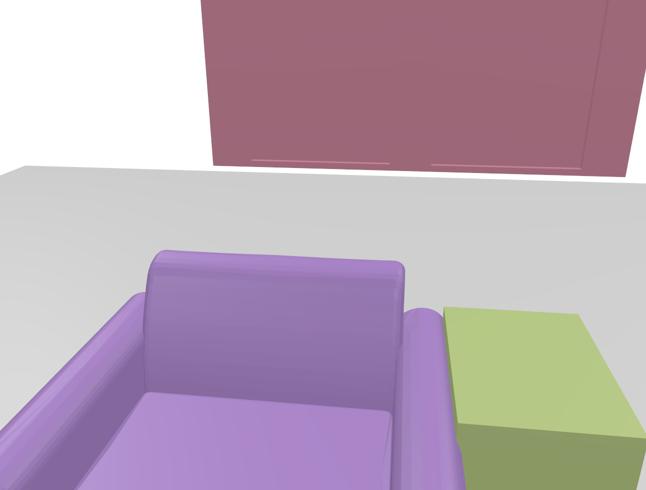}
		\includegraphics[width=\textwidth]
		{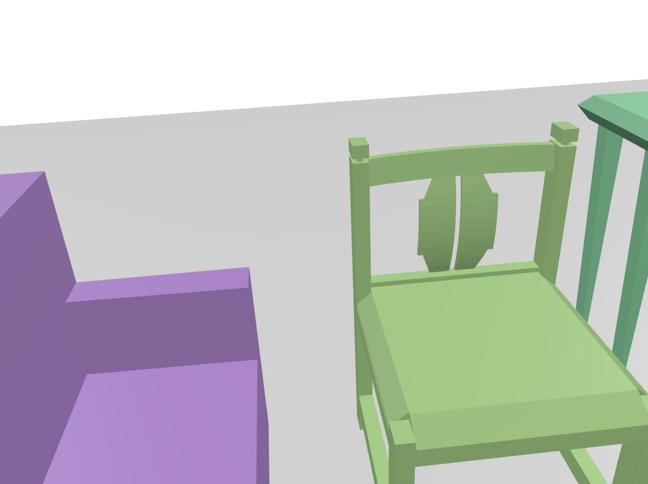}
		\includegraphics[width=\textwidth]
		{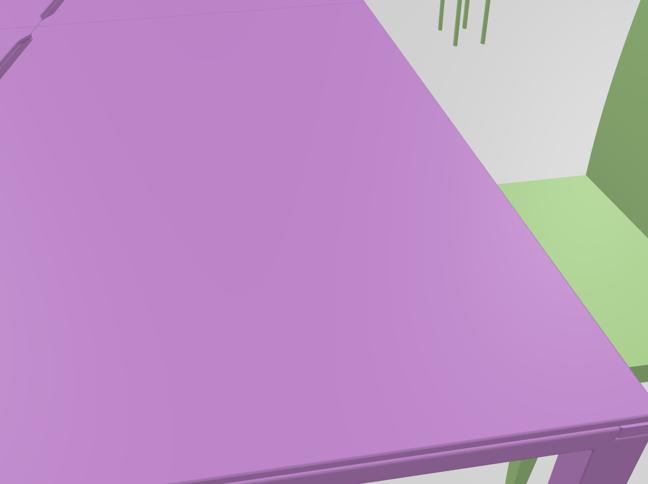}
		\includegraphics[width=\textwidth]
		{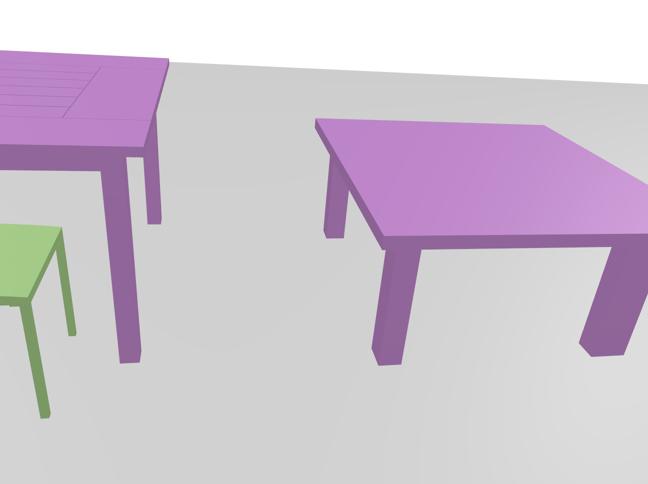}
		\includegraphics[width=\textwidth]
		{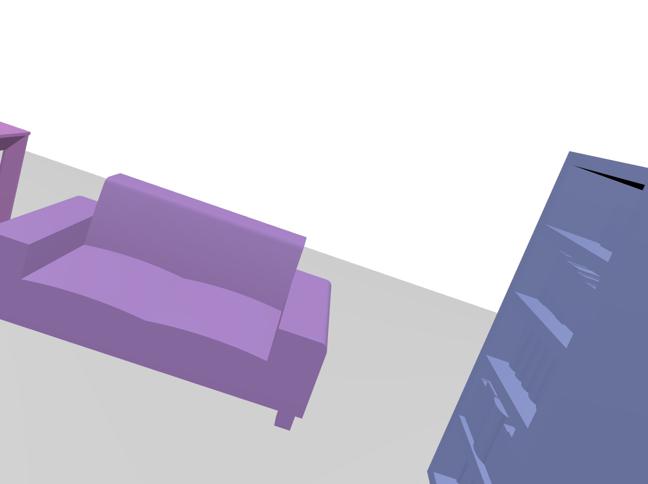}
		\includegraphics[width=\textwidth]
		{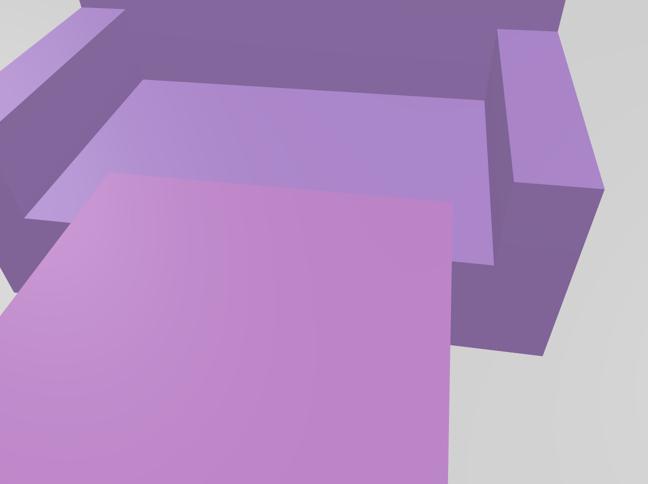}
		\includegraphics[width=\textwidth]
		{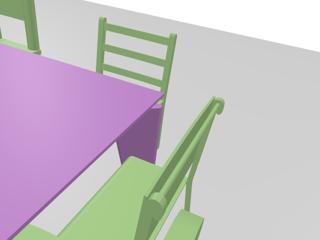}
	\end{subfigure}
	\rulesep
	\begin{subfigure}[t]{0.155\textwidth}
		\includegraphics[width=\textwidth]
		{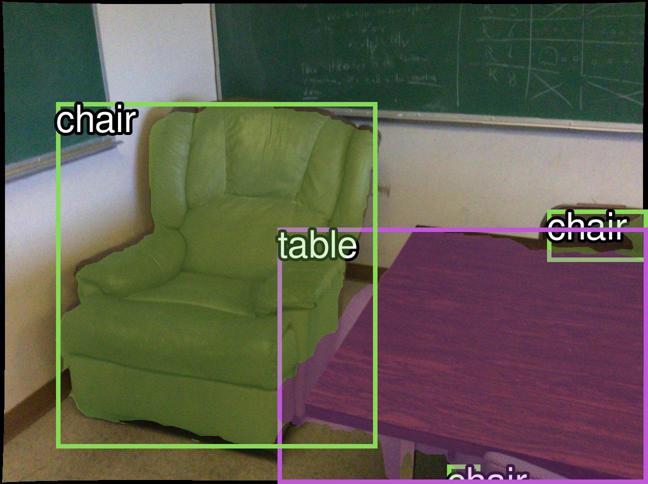}
		\includegraphics[width=\textwidth]
		{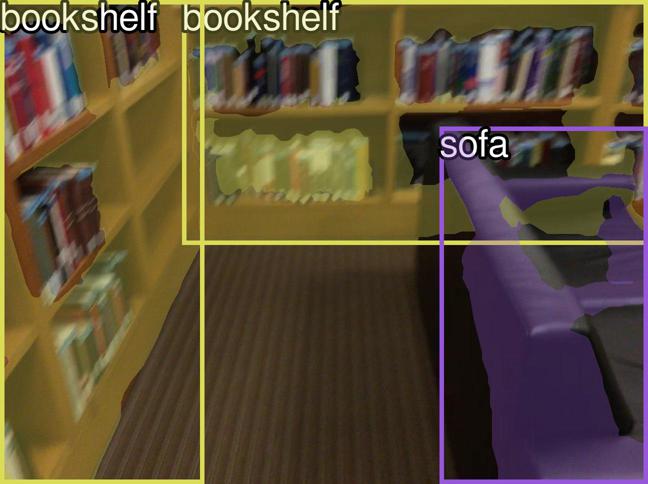}
		\includegraphics[width=\textwidth]
		{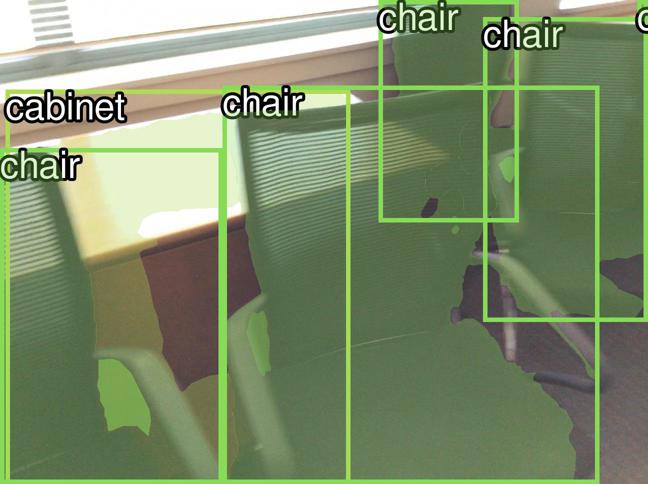}
		\includegraphics[width=\textwidth]
		{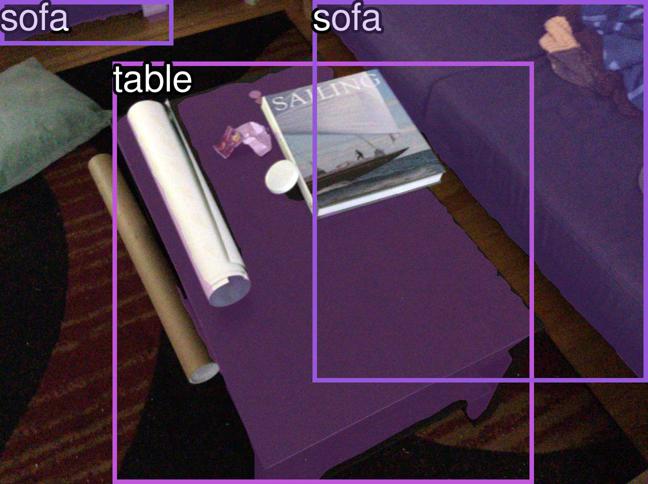}
		\includegraphics[width=\textwidth]
		{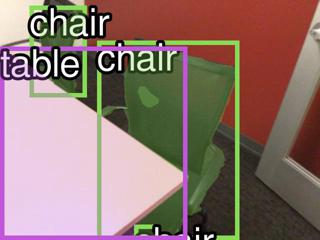}
		\includegraphics[width=\textwidth]
		{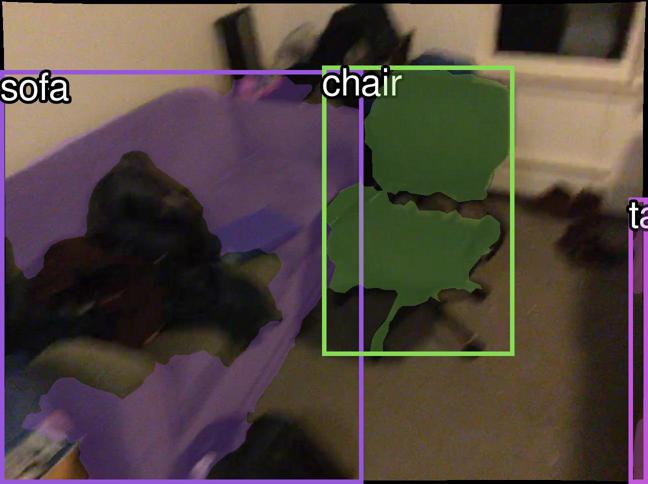}
		\includegraphics[width=\textwidth]
		{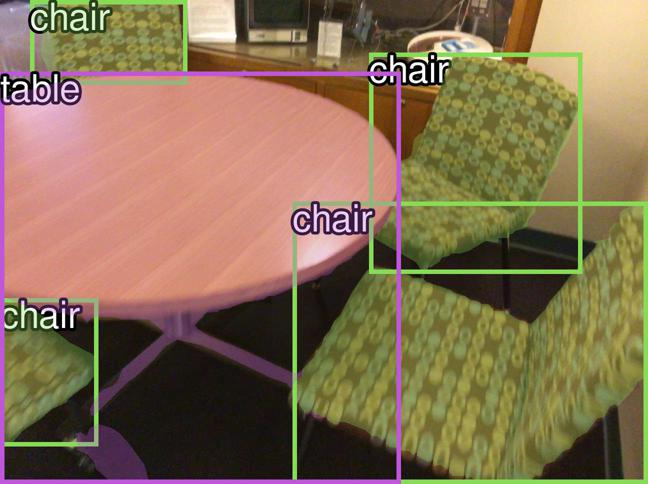}
	\end{subfigure}
	\begin{subfigure}[t]{0.155\textwidth}
		\includegraphics[width=\textwidth]
		{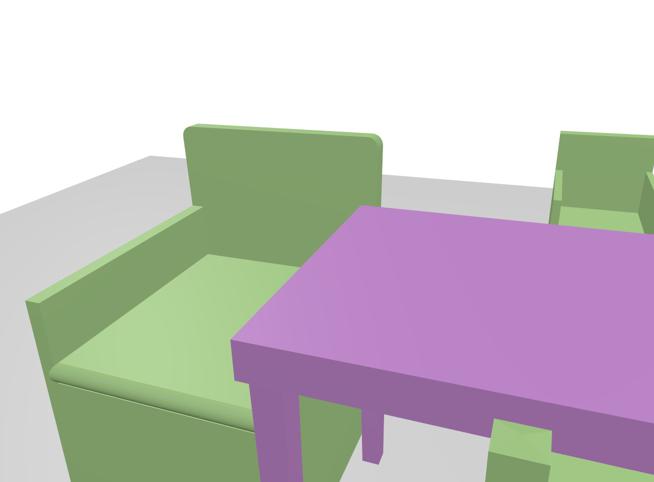}
		\includegraphics[width=\textwidth]
		{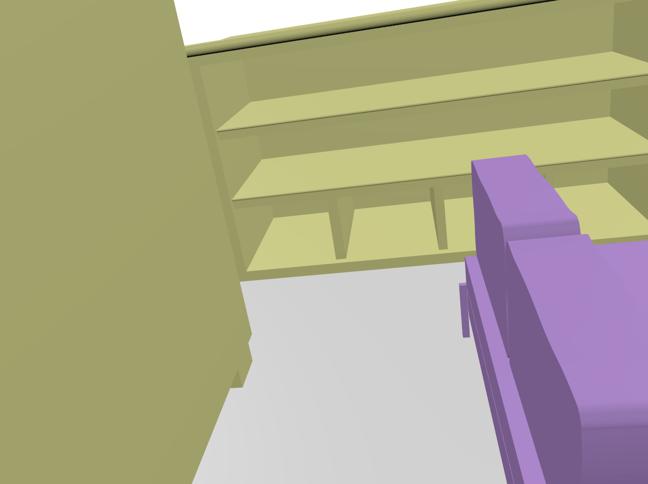}
		\includegraphics[width=\textwidth]
		{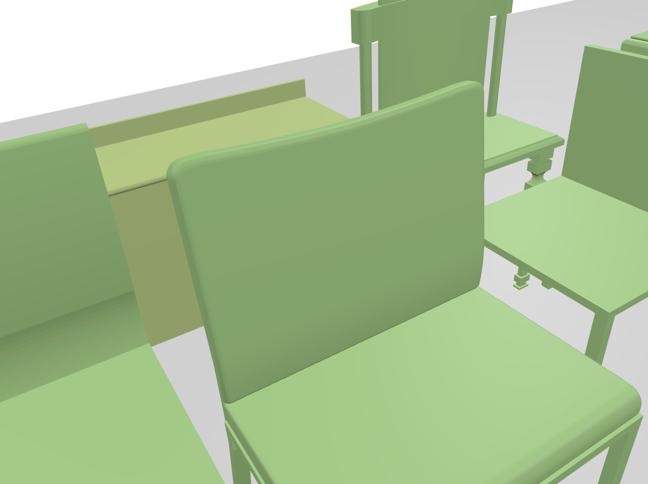}
		\includegraphics[width=\textwidth]
		{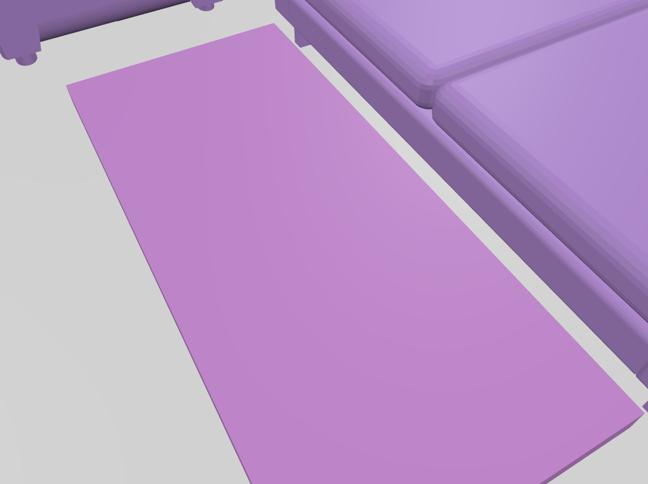}
		\includegraphics[width=\textwidth]
		{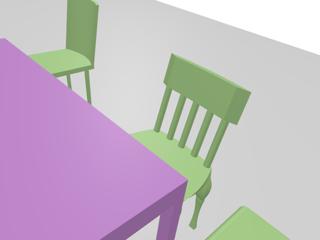}
		\includegraphics[width=\textwidth]
		{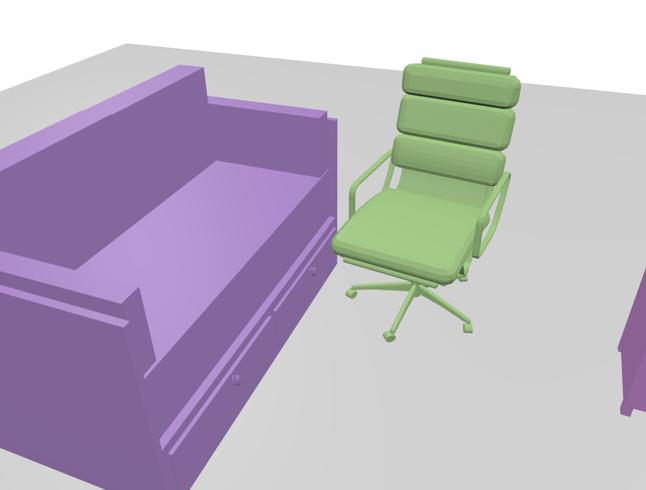}
		\includegraphics[width=\textwidth]
		{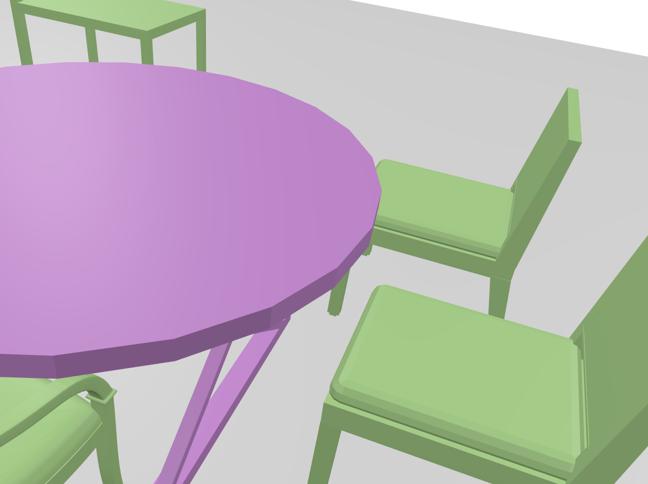}
	\end{subfigure}
	\rulesep
	\begin{subfigure}[t]{0.155\textwidth}
		\includegraphics[width=\textwidth]
		{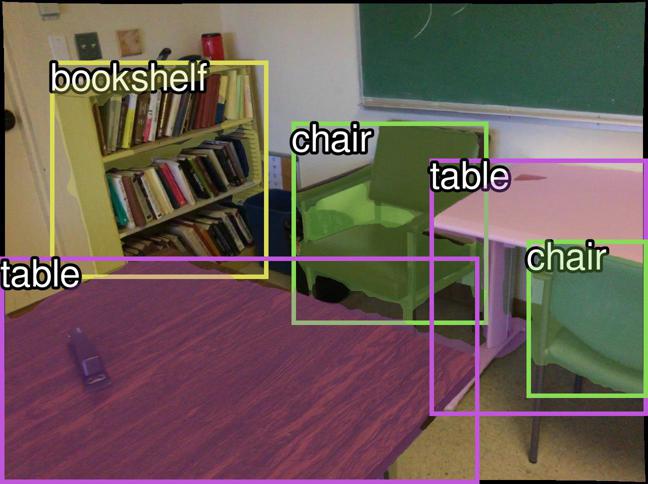}
		\includegraphics[width=\textwidth]
		{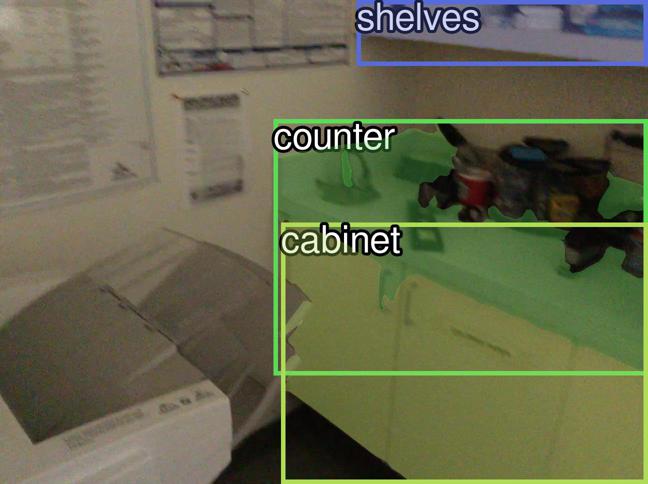}
		\includegraphics[width=\textwidth]
		{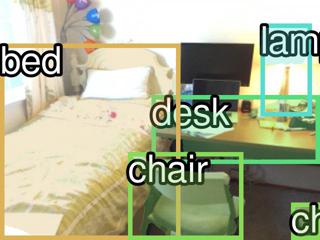}
		\includegraphics[width=\textwidth]
		{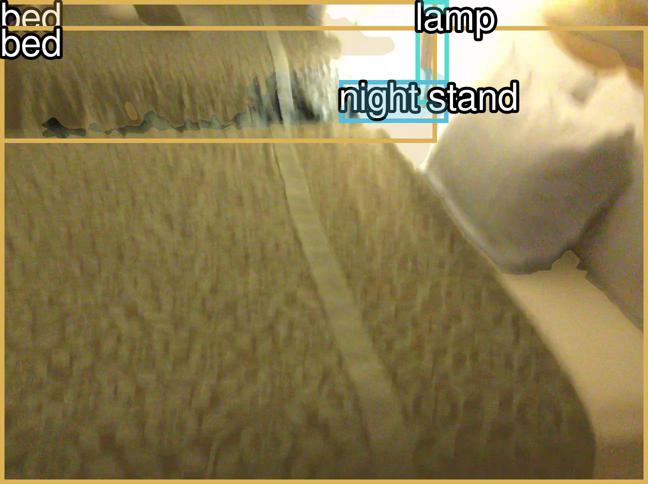}
		\includegraphics[width=\textwidth]
		{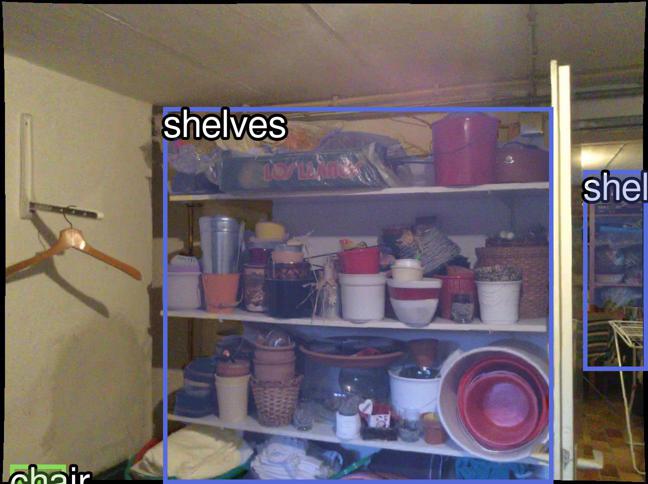}
		\includegraphics[width=\textwidth]
		{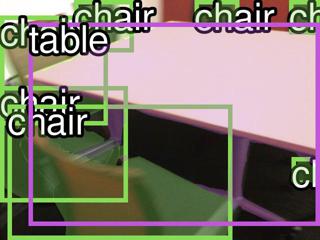}
		\includegraphics[width=\textwidth]
		{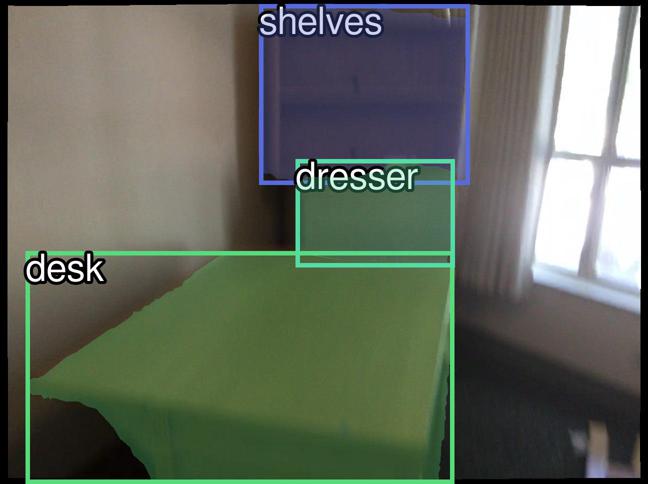}
	\end{subfigure}
	\begin{subfigure}[t]{0.155\textwidth}
		\includegraphics[width=\textwidth]
		{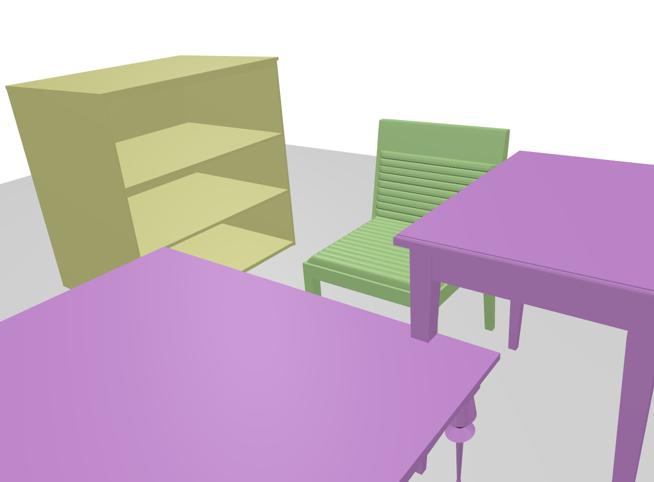}
		\includegraphics[width=\textwidth]
		{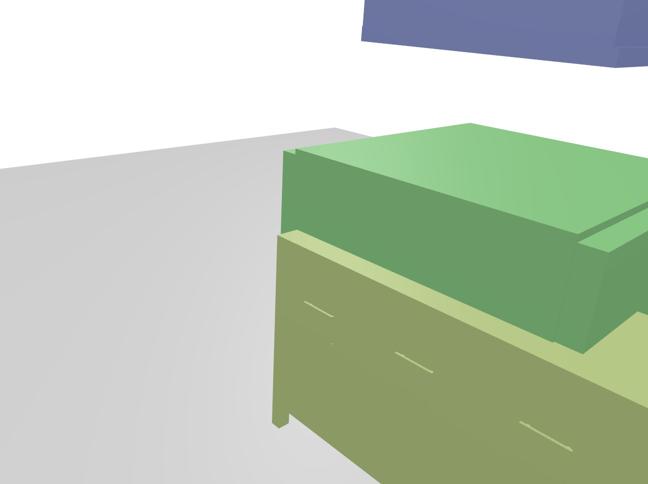}
		\includegraphics[width=\textwidth]
		{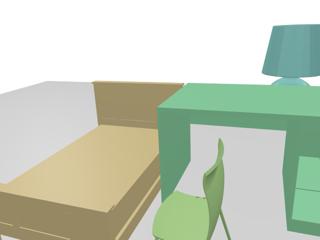}
		\includegraphics[width=\textwidth]
		{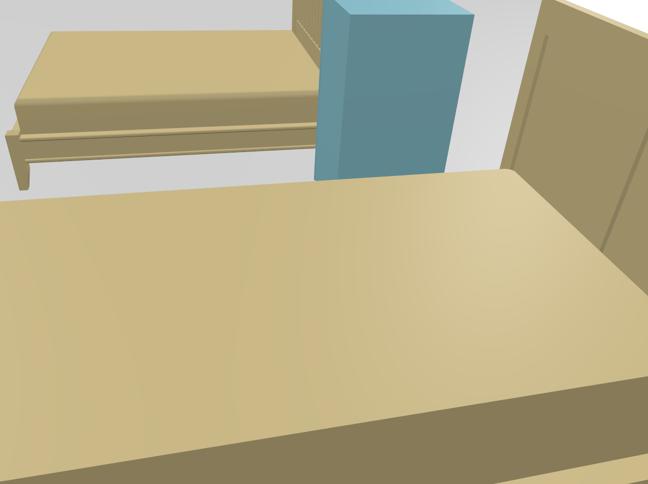}
		\includegraphics[width=\textwidth]
		{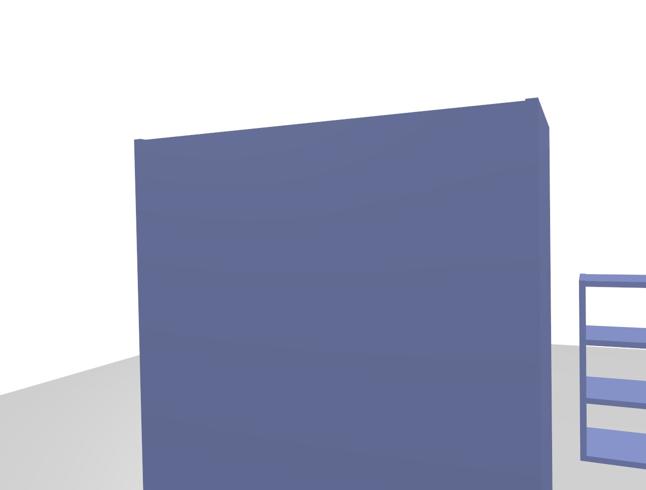}
		\includegraphics[width=\textwidth]
		{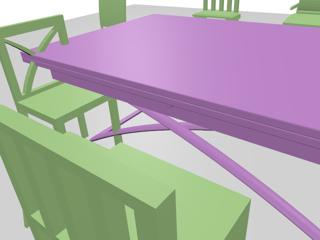}
		\includegraphics[width=\textwidth]
		{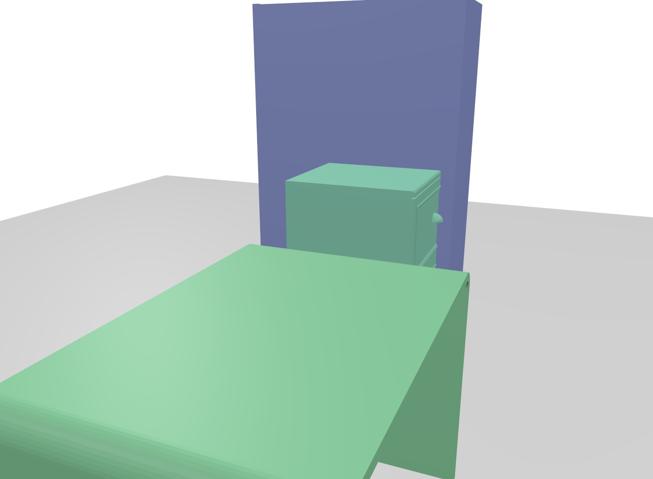}
	\end{subfigure}
	\caption{Additional qualitative results on single-view scene reconstruction.}
	\label{fig:add_qualitative_svr}
	\vspace{-1em}
\end{figure*}

\section{Additional Ablation Experiments}
\label{sec:ablation}
In our training data, each scene contains at most 100 images to train our model (see Sec.~\ref{sec:dataproc}), where each scene has multiple rooms (bedroom, living room, etc.). The view number of each room is proportional to its floor area. More views sampled in a room indicate better coverage to capture more indoor objects.

In Tab.~\ref{tab:ab_view_num}, we investigate the influence of different view numbers on our scene generation performance. We observe that better view coverage brings notable gains to all metrics, which is particularly significant for large-scale rooms (e.g., living rooms).
This indicates that, without enough view coverage, our method cannot observe objects in different views, which would lead to ambiguities in object localization and deformation. However, more camera views means more training time, and extra human labor on data collection. In our experiments, we observe that using 100 views for both 3D-Front and ScanNet scenes presents enough scene coverage to learn general 3D scene priors.

\section{Additional Qualitative Results}
\label{sec:qualitative}
In Fig.~\ref{fig:add_qualitative_scene_gen} and Fig.~\ref{fig:add_qualitative_svr}, we list additional qualitative results on 3D scene synthesis and single-view reconstruction from the test set.

\end{document}